\RequirePackage{fix-cm}
\documentclass[11pt]{article}
\usepackage{arxiv}
\usepackage{needspace}
\usepackage{array}
\usepackage[authoryear,round]{natbib}
\usepackage{authblk}

\date{}

\renewcommand{\headeright}{}
\renewcommand{\undertitle}{}
\renewcommand{\shorttitle}{Tabular Foundation Models for Credit Risk Prediction}
\usepackage[T1]{fontenc}      
\usepackage[utf8]{inputenc}   
\usepackage{newtxtext}        

\usepackage{amsmath,amsfonts,amsthm,bm}
\usepackage{newtxmath}        
\usepackage{bbm}
\usepackage{empheq}

\usepackage{microtype}

\usepackage{setspace}
\usepackage{graphicx}
\usepackage{subcaption}
\usepackage{adjustbox}
\usepackage{rotating}

\usepackage{booktabs}
\usepackage{makecell}
\usepackage{multirow}
\usepackage{arydshln}
\usepackage{siunitx}

\usepackage{tcolorbox}

\usepackage{pifont}
\usepackage{verbatim}
\usepackage{xurl}

\theoremstyle{definition}

\usepackage{hyperref}

\hypersetup{
colorlinks=true,
breaklinks=true,
linkcolor=blue,
citecolor=blue,
urlcolor=blue,
bookmarksopen=false
}

\usepackage[acronym,nowarn]{glossaries}
\glsdisablehyper
\setacronymstyle{long-short}

\newacronym{cd}{CD}{critical difference}
\newacronym{rope}{ROPE}{region of practical equivalence}
\newacronym{ols}{OLS}{ordinary least squares}
\newacronym{mlp}{MLP}{multilayer perceptron}
\newacronym{talent}{TALENT}{Tabular Analytics and Learning Toolbox}
\newacronym{oned}{1D}{one-dimensional}
\newacronym{twod}{2D}{two-dimensional}
\newacronym{gmsc}{GMSC}{Give Me Some Credit}

\newacronym{ai}{AI}{artificial intelligence}
\newacronym{llm}{LLM}{large language model}
\newacronym{nlp}{NLP}{natural language processing}
\newacronym{dl}{DL}{deep learning}
\newacronym{ml}{ML}{machine learning}
\newacronym{gbm}{GBM}{gradient boosting machine}
\newacronym{pfn}{PFN}{prior-data fitted network}
\newacronym{tfm}{TFM}{tabular foundation model}
\newacronym{icl}{ICL}{in-context learning}
\newacronym{ssl}{SSL}{self-supervised learning}
\newacronym{scm}{SCM}{structural causal model}
\newacronym{bnn}{BNN}{Bayesian neural network}
\newacronym{gnn}{GNN}{graph neural network}
\newacronym{iid}{IID}{independent and identically distributed}
\newacronym{pca}{PCA}{principal component analysis}

\newacronym{ppd}{PPD}{Posterior Predictive Distribution}
\newacronym{nll}{NLL}{Negative Log-Likelihood}
\newacronym{kl}{KL}{Kullback-Leibler}
\newacronym{mcmc}{MCMC}{Markov Chain Monte Carlo}
\newacronym{vi}{VI}{Variational Inference}

\newacronym{pd}{PD}{probability of default}
\newacronym{lgd}{LGD}{loss given default}
\newacronym{ead}{EAD}{exposure at default}
\newacronym{sme}{SME}{Small and Medium-sized Enterprises}
\newacronym{fi}{FI}{financial institution}
\newacronym{irb}{IRB}{internal ratings-based}
\newacronym{ifrs 9}{IFRS 9}{International Financial Reporting Standards 9}
\newacronym{gdpr}{GDPR}{General Data Protection Regulation}
\newacronym{rwa}{RWA}{Risk-Weighted Asset}
\newacronym{emp}{EMP}{expected maximum profit}

\newacronym{xai}{XAI}{Explainable Artificial Intelligence}
\newacronym{shap}{SHAP}{SHapley Additive exPlanations}
\newacronym{lime}{LIME}{Local Interpretable Model-agnostic Explanations}

\newacronym{ftt}{FTT}{FT-Transformer}
\newacronym{knn}{KNN}{K-Nearest Neighbors}
\newacronym{svm}{SVM}{Support Vector Machine}
\newacronym{svr}{SVR}{Support Vector Regression}
\newacronym{ncm}{NCM}{Nearest Class Mean}

\newacronym{auc}{AUC}{Area Under the Receiver Operating Characteristic Curve}
\newacronym{roc}{ROC}{Receiver Operating Characteristic}
\newacronym{ap}{AP}{Average Precision}
\newacronym{ks}{KS}{Kolmogorov--Smirnov}
\newacronym{mcc}{MCC}{Matthews Correlation Coefficient}
\newacronym{mse}{MSE}{mean squared error}
\newacronym{rmse}{RMSE}{root mean squared error}
\newacronym{mae}{MAE}{mean absolute error}
\newacronym{medae}{MedAE}{median absolute error}
\newacronym{pama}{PAMA}{probability of achieving maximum accuracy}
\newacronym{hpo}{HPO}{hyperparameter optimization}

\newacronym{rq}{RQ}{research question}
\newacronym{eu}{EU}{European Union}
\newacronym{gpu}{GPU}{graphics processing unit}
\newacronym{vsc}{VSC}{Flemish Supercomputer Center}
\newacronym{uci}{UCI}{University of California, Irvine}
\newacronym{fico}{FICO}{Fair Isaac Corporation}
\newacronym{sas}{SAS}{Statistical Analysis System}
\newacronym{ltfs}{LTFS}{L\&T Financial Services}
\newacronym{hmeq}{HMEQ}{Home Equity}
\newacronym{qassmax}{QASSMax}{Query-Aware Scalable Softmax}
\newacronym{gb}{GB}{gradient boosting}
\newacronym{rf}{RF}{random forest}
\newacronym{dt}{DT}{decision tree}

\hypersetup{
  pdftitle={Tabular Foundation Models for Credit Risk Prediction: A Dedicated Review and Empirical Comparison},
  pdfauthor={Bart Baesens; Andreas Goethals; Simon De Vos; Cristián Bravo; Stefan Lessmann; David Martens; Victor Medina-Olivares; Christophe Mues; María Óskarsdóttir; Seppe vanden Broucke; Tony Van Gestel; Tim Verdonck; Wouter Verbeke},
  pdfsubject={Credit risk prediction with tabular foundation models}
}
\title{Tabular Foundation Models for Credit Risk Prediction: A Dedicated Review and Empirical Comparison}

\author[1]{Bart Baesens}
\author[1]{Andreas Goethals}
\author[1]{Simon De Vos}
\author[2]{Cristián Bravo}
\author[3]{Stefan Lessmann}
\author[4]{David Martens}
\author[5]{Victor Medina-Olivares}
\author[6]{Christophe Mues}
\author[7,8]{María Óskarsdóttir}
\author[9,1]{Seppe vanden Broucke}
\author[10]{Tony Van Gestel}
\author[11,12]{Tim Verdonck}
\author[1]{Wouter Verbeke}

\affil[1]{Faculty of Economics and Business, KU Leuven, Belgium}
\affil[2]{Department of Statistical and Actuarial Sciences, Western University, Canada}
\affil[3]{School of Business and Economics, Humboldt University of Berlin, Germany}
\affil[4]{Department of Engineering Management, University of Antwerp, Belgium}
\affil[5]{Business School, University of Edinburgh, United Kingdom}
\affil[6]{Business School, University of Southampton, United Kingdom}
\affil[7]{School of Mathematical Sciences, University of Southampton, United Kingdom}
\affil[8]{Department of Computer Science, Reykjavik University, Iceland}
\affil[9]{Department of Business Informatics and Operations Management, Ghent University, Belgium}
\affil[10]{Dexia, Belgium}
\affil[11]{Department of Mathematics, University of Antwerp, Belgium}
\affil[12]{Department of Mathematics, KU Leuven, Belgium}

\begin{document}
\maketitle

\begin{abstract}
Predictive models are central to credit risk management, where the accuracy of default-probability and loss estimates affects the profitability of lending and the stability of the financial system. Decades of benchmarking research have consolidated the state of the art, with \gls{gb} models often representing the strongest \gls{ml} benchmark, while linear and logistic models remain important regulatory and industry reference models.
The recent advent of tabular data foundation models (TFMs)\glsunset{tfm} introduces a new paradigm for predictive modeling. Rather than fitting a model to a target dataset, they use \gls{icl} to integrate information from the target dataset with prior knowledge acquired through large-scale pretraining on diverse collections of (synthetic) tabular data. 
We conjecture that access to information beyond the target data is beneficial in small-data settings and may help address longstanding challenges, including low-default portfolios. Whether this promise holds in practice remains an open question.
This paper benchmarks state-of-the-art \glspl{tfm} against a broad set of classical and advanced \gls{ml} competitors on two core tasks: \gls{pd} and \gls{lgd} modeling. We evaluate $33$ classification methods across $14$ \gls{pd} datasets and $27$ regression methods across $7$ \gls{lgd} datasets, spanning a range of performance indicators and experimental conditions. The most recent \glspl{tfm}, in particular TabPFN-3 and TabICLv2, achieve the strongest average results, with the largest gains in small-sample and strongly imbalanced settings. These results are notable because these models are used with their pretrained parameters fixed and without downstream \gls{hpo} or task-specific parameter updates.
\end{abstract}

\noindent\textbf{Keywords:} Analytics \textbullet{} Credit Risk \textbullet{} Tabular Foundation Models \textbullet{} Probability of Default \textbullet{} Loss Given Default

\clearpage
\section{Introduction}
\label{sec:introduction}
\setlength{\parskip}{2pt}
\enlargethispage{\baselineskip}

Foundation models are large-scale \gls{ml} systems trained on vast and heterogeneous corpora, enabling transfer across a wide range of downstream tasks, often after task-specific fine-tuning \citep{schneider2024foundation}. They have shown great promise for \gls{nlp} and computer vision, where systems such as GPT-4, CLIP, and diffusion models advanced text generation, zero-shot classification, and image synthesis \citep{radford2021learning,bubeck2023sparks,awais2025foundation}, marking a shift from task-specific models toward general-purpose systems.

Tabular data, the typical data format in credit risk analytics, differs markedly from unstructured data, due to heterogeneous feature types, disparate scales, and dataset-specific semantics. These characteristics have historically impeded knowledge transfer. Tree-based ensembles, especially \glspl{gbm}, have long served as the strongest benchmarks for many tabular prediction tasks, consistently outperforming many task-specific \gls{dl} alternatives in broad comparative studies \citep{2015.LESSMANN.benchmarkingstateoftheartclassificationalgorithmscreditscoringupdateresearch,2021.GUNNARSSON.deeplearningcreditscoringdont,2022.SHWARTZ-ZIV.tabulardatadeeplearningnotallyouneed,2022.GRINSZTAJN.whytreebasedmodelsstilloutperformdeeplearningtypicaltabulardata}. However, recent progress in \glspl{pfn} challenges the status quo by introducing foundation-model-style prediction for tabular data. \glspl{pfn} pretrain on vast amounts of (synthetic) data and transfer the extracted statistical regularities to downstream tasks through \gls{icl} \citep{2025.HOLLMANN.accuratepredictionssmalldatatabularfoundationmodel,2025.QU.tabicltabularfoundationmodelincontextlearninglargedata,2025.ZHANG.mitramixedsyntheticpriorsenhancingtabularfoundationmodels,2025.MA.tabdptscalingtabularfoundationmodelsrealdata}. At inference, a \gls{pfn} conditions on labeled examples and outputs predictive distributions for the unlabeled instances, with no parameter updates or hyperparameter tuning. Specifically, standard \glspl{pfn} ingest the entire labeled and unlabeled table in a single forward pass, whereas retrieval-based variants condition each prediction on a query-specific set of neighbors \citep{2025.MA.tabdptscalingtabularfoundationmodelsrealdata}.

This paper examines whether this \gls{pfn}-based \gls{tfm} paradigm benefits credit risk modeling. Credit risk management is critically dependent on accurate predictions of risk parameters, notably the \gls{pd} and the \gls{lgd}. More accurate credit decisions directly affect capital requirements, allocation, and portfolio steering under the Basel framework, provisioning under \gls{ifrs 9}, and risk-adjusted pricing. These stakes make \glspl{pfn} particularly relevant for credit risk modeling practice, because their properties address several operational challenges faced in credit risk prediction. First, as zero-shot predictors, they avoid repeated hyperparameter tuning and retraining across tasks and portfolios. This property is attractive for new or developing portfolios with limited data availability, such as low-default portfolios, specialized corporate segments, or new credit products. Second, \glspl{pfn} could enable a consistent modeling approach across risk parameters (\gls{pd}, \gls{lgd}, and potentially \gls{ead}), thereby benefiting governance by simplifying validation, monitoring, and model risk management. From a supervisory perspective, cross-parameter consistency within \glspl{fi}, as well as methodological uniformity across them, may reduce the burden of model review for these institutions. Third, by combining broad synthetic pretraining with \gls{icl} on institutional data, \glspl{pfn} may improve sample efficiency and reduce over-reliance on historically biased local patterns, thereby supporting fairer and more inclusive decision making \citep{oskarsdottir2019value,kozodoi2022fairness,moldovan2023algorithmic}.

Despite this potential, caution is warranted. Credit risk modeling poses specific challenges (small datasets, severe target class imbalance, heterogeneous borrower populations, economic cycles, and regulatory constraints) that current \gls{pfn}-based \glspl{tfm} may not fully address \citep{2016.BAESENS.creditriskanalyticsmeasurementtechniquesapplicationsexamplessas}. Moreover, a systematic evaluation of \glspl{pfn} for credit risk prediction has been lacking: while anecdotal evidence points to strong performance on small datasets, it remains unclear whether \glspl{pfn} can consistently match tuned baselines across diverse credit portfolios and targets.

We address this gap through three \glspl{rq}, each examined through a corresponding empirical analysis. \Gls{rq}~$1$ asks whether \gls{pfn}-based \glspl{tfm} are predictively competitive in \gls{pd} classification and \gls{lgd} regression. To answer it, we benchmark these models against an extensive set of classical and advanced \gls{ml} methods: $33$ classification algorithms across $14$ \gls{pd} datasets and $27$ regression algorithms across $7$ \gls{lgd} datasets that vary in size, dimensionality, and target characteristics. Our evaluation follows the benchmarking tradition in credit scoring, uses cross-validation, and reports a comprehensive set of performance measures. As a by-product, the benchmark provides the first credit-risk assessment, to the best of our knowledge, of several recent non-\gls{pfn} \gls{dl} models for tabular data \citep{ye2024closer}. \Gls{rq}~$2$ asks how the models perform under data scarcity; we address it through an auxiliary learning-curve experiment that progressively varies the available sample size. \Gls{rq}~$3$ asks how robust the models are to class imbalance; we address it through a second auxiliary experiment that progressively reduces the minority-class proportion in the \gls{pd} datasets. Together, these analyses assess both the models' general predictive competitiveness and how their relative performance changes under controlled data scarcity and artificial class imbalance.

The remainder of this paper is organized as follows. Section~\ref{sec:literature_review} reviews credit risk modeling and recent advances in \gls{dl} for tabular data. Section~\ref{sec:prior-data_fitted_networks_pfns} introduces the Bayesian foundations, architectures, and pretraining designs of \glspl{pfn}. Section~\ref{sec:experimental_design} formulates the three \glspl{rq} and describes the datasets, benchmarked methods, and evaluation protocols. Section~\ref{sec:results_and_discussion} presents the empirical findings and answers each \gls{rq}. Section~\ref{sec:implications_and_rd} discusses the implications for credit risk management and research directions, and Section~\ref{sec:conclusions} concludes the paper. Appendices~\ref{supp:stats}--\ref{supp:data} provide, respectively, the statistical analysis, additional performance measures, supporting visualizations, per-dataset prediction distributions, complete rank-based results, and further dataset documentation, including target distributions, eligibility for the auxiliary experiments, and the pretraining-overlap audit.

\setlength{\parskip}{5.5pt}
\clearpage
\section{Literature review}
\label{sec:literature_review}
This paper presents extensive empirical results on the predictive performance of \gls{pfn}-based \glspl{tfm} for \gls{pd} and \gls{lgd} modeling. We build on a large body of prior work on credit risk prediction and extend it by introducing a new family of predictive methods. Beyond credit risk and, more generally, financial applications, the paper also draws on prior work on \gls{dl} for tabular data, which paved the way for the recently introduced \glspl{pfn}. This section reviews prior work in these two fields and elaborates on our contributions.

\Needspace{8\baselineskip}
\subsection{Credit risk modeling}
\label{subsec:credit_risk_modeling}
Credit risk refers to the potential loss arising from a counterparty's failure to meet contractual obligations \citep{2016.BAESENS.creditriskanalyticsmeasurementtechniquesapplicationsexamplessas}. In retail and wholesale banking, risk models quantify key risk parameters used for pricing, capital, provisioning, and portfolio management: the \gls{pd},  which estimates the likelihood of a borrower defaulting over a fixed horizon, and the \gls{lgd}, which measures the proportion of the exposure that is not recovered if default occurs. In both \gls{pd} and \gls{lgd} modeling, predictive accuracy is key to the efficiency and effectiveness of credit risk management \citep{2003.BAESENS.benchmarkingstateoftheartclassificationalgorithmscreditscoring,2012.LOTERMAN.benchmarkingregressionalgorithmslossgivendefaultmodeling,2021.GUNNARSSON.deeplearningcreditscoringdont}. Better predictions directly translate into improved capital efficiency (e.g., when estimating capital requirements under the Basel Accord's \gls{irb} approach), more precise and timely provisioning under \gls{ifrs 9} through improved estimation of lifetime expected credit losses, reduced impairments, and better pricing decisions \citep{bastos2022explainable}.
Regulators likewise emphasize the discriminatory power and calibration of scorecards, for example, through backtesting, monitoring stability, and validation requirements. 

Aiming to unlock financial rewards from more accurate predictions, the community routinely assesses the potential of new prediction methods and proposes methodological adjustments tailored to characteristic modeling challenges. For \gls{pd}, these challenges include, e.g., class imbalance \citep{brown2012experimental,marques2013suitability,engelmann2021conditional,wang2025secure}, reject inference \citep{banasik2007reject,kozodoi2025fighting}, or economic cycle dependency \citep{2012.BELLOTTI.lossgivendefaultmodelsincorporatingmacroeconomicvariablescreditcards,dirick2019macroeconomic,djeundje2025devil,distaso2025business}. Compared with \gls{pd} prediction, fewer studies have considered \gls{lgd} modeling \citep{yao2015support,hurlin2018loss,cheng2025multiview}, which may be due to the limited public availability of \gls{lgd} datasets. Datasets frequently used in \gls{lgd} modeling research include the \textit{LendingClub} and \textit{Freddie Mac} datasets \citep{zhou2018p2p,calabrese2022modelling}, which provide sufficient information to approximate \gls{lgd}. Given the characteristic bimodal shape of real-world \gls{lgd} distributions, arising, for example, from mixed portfolios comprising both secured and unsecured loans, many studies recommend multi-stage models \citep{tomarchio2019modelling,papouskova2019two,starosta2021loss}. Dynamic models that explicitly recognize capital inflows and outflows throughout the workout period \citep{thomas2016modelling} are another popular research topic in the \gls{lgd} literature, often involving survival modeling \citep[e.g.,][]{joubert2021adapting}.

Given the prevalence of \gls{ml} and advanced predictive methods in the credit scoring literature, several benchmarking studies have periodically consolidated the state of the art by applying an arsenal of statistical and \gls{ml} methods across multiple datasets, performance metrics, and experimental conditions. Prominent examples include \citet{2003.BAESENS.benchmarkingstateoftheartclassificationalgorithmscreditscoring,2015.LESSMANN.benchmarkingstateoftheartclassificationalgorithmscreditscoringupdateresearch,2021.GUNNARSSON.deeplearningcreditscoringdont,jiang2023benchmarking} and \citet{2012.LOTERMAN.benchmarkingregressionalgorithmslossgivendefaultmodeling,bellotti2021forecasting}, for \gls{pd} and \gls{lgd} prediction, respectively. Closely connected to comparing the predictive performance of alternative methods is the question of how to properly assess predictions. Several attempts have been made to connect the accuracy of model predictions to profitability and other measures of business impact \citep{2014.VERBRAKEN.developmentapplicationconsumercreditscoringmodelsusingprofitbasedclassificationmeasures,garrido2018robust,martin2025novel}. While quantification of business impact is more developed in \gls{pd} modeling, some studies have pursued similar goals in the context of \gls{lgd} \citep{hurlin2018loss}.

Conditional on the availability of suitable indicators of a risk model's business impact, the predictive models themselves can be optimized for cost-minimization \citep{bahnsen2015example} or profit maximization \citep{finlay2010credit,serranocinca2016use,alfonsosanchez2024optimizing,xu2025profit}. Beyond the actual risk model, other stages in the modeling pipeline can be optimized for cost/profit objectives. For example, \citet{maldonado2017integrated,kozodoi2019multiobjective} introduce techniques for profit-oriented feature selection. Also targeting the input data of predictive models, recent work aims to extract information from borrower networks that standard methods relying on the \gls{iid} assumption would fail to capture. Using methods from geospatial econometrics, related studies have, for example, established correlations between default events across geographical neighborhoods \citep{calabrese2020spatial,medinaolivares2025joint}. More generally, graph-based algorithms, such as \glspl{gnn}, facilitate the processing of comprehensive dependency structures, providing modelers with considerable freedom in how to design an influence network. 
 
\citet{oskarsdottir2021multilayer} pioneered the use of multilayer network analysis in credit-risk modeling, showing that borrower connections can provide substantial predictive gains. This line of work has since developed toward graph-based and \gls{gnn} approaches for credit-risk prediction \citep{shi2024improved,zandi2025attention}.
The pursuit to incorporate auxiliary information in risk models beyond what classical scorecards embody has also inspired scholars to systematically explore novel sources of data, including social media postings \citep{decnudde2018what}, psychometric data \citep{djeundje2021enhancing}, and textual data \citep{stevenson2021value,kriebel2022credit,wu2025unleashing}. A distinctive advantage of such alternative data is that it may facilitate accurate risk assessment of \textit{thin-book clients} whose limited credit history may otherwise exclude them from accessing financial services \citep{oskarsdottir2019value}.

Facing increasing concern related to the omnipresence of advanced, \gls{ai}-based, decision models in lending (and beyond), much recent work concentrates on \gls{xai} techniques to ensure that the mechanisms underlying such models are well-understood \citep{debock2024explainable,martens2025beware}. Model explainability is a regulatory imperative and prerequisite for achieving higher-level objectives, including robustness, safety, and fairness \citep{barredoarrieta2020explainable}. Recent work on \gls{xai} for credit risk examines the interplay between \gls{ml}-based scorecards and \gls{xai} tools in general \citep[e.g.,][]{bastos2022explainable}, the robustness of explanations \citep{ballegeer2025evaluating}, or the moderating effect of longstanding challenges, such as class imbalance \citep{chen2024interpretable}. 

Benefiting from its strong theoretical foundations in cooperative game theory, the \gls{shap} framework for additive feature attribution has gained considerable popularity in the industry \citep{dutoit2023shapley,borgonovo2024many}. Unlike \gls{shap}, which decomposes a model's prediction, be it right or wrong, \citet{hue2025measuring} introduces an interesting challenger approach that inherits the same theoretical underpinnings but decomposes a scorecard's predictive performance. The corresponding insights might be even more meaningful to decision-makers as, on the one hand, performance statistics are easier to interpret than predictions, and, on the other hand, knowing which features drive performance provides directly actionable insights to improve the model. Apart from techniques to explain otherwise opaque \gls{ml} models, several studies devise new, intrinsically interpretable learning algorithms and demonstrate that these often avoid sacrificing (much) predictive accuracy. Representative examples include tree-based approaches \citep{decaigny2018new,carrizosa2025enhancing}, generalized additive models \citep{kraus2024interpretable}, and interpretable \gls{dl} techniques \citep{medinaolivares2024deep,zografopoulos2025industry}.  

Regulatory frameworks governing the use of \gls{ai} in consumer-facing applications, such as the \gls{eu} \gls{ai} Act, the \gls{eu} \gls{gdpr}, and recent revisions to the Basel capital accord, require \glspl{fi} to demonstrate that their risk models are not only explainable but also free from discriminatory bias against historically disadvantaged groups. Seminal papers provide evidence that scorecards exhibit such biases and that \gls{ml}-based approaches are especially vulnerable \citep{fu2021crowds,fuster2022predictably}. Focusing on retail lending decisions, \citet{kozodoi2022fairness} analyze available statistical indicators of algorithmic fairness and algorithms to mitigate disparate impact in the \gls{ml} pipeline for their suitability in credit scoring. \citet{hurlin2025fairness} introduce a powerful tool to test a given scorecard for algorithmic bias, enabling lenders to verify compliance (or the lack thereof) with regulatory requirements. 

In summary, \gls{ml} and \gls{ai} play a pivotal role in credit risk modeling. They continuously supply novel technologies that aid lenders in their quest for more effective, efficient, and ultimately profitable decision models that also comply with regulations. 

This paper contributes to the empirical credit risk modeling literature. Although we focus on established classification and regression settings for \gls{pd} and \gls{lgd} modeling, respectively, our goal extends beyond assessing yet another prediction method. Instead, we assess a \textit{novel paradigm} for risk prediction: the use of tabular data foundation models for zero-shot prediction. Adopting a zero-shot prediction method to support lending operations entails a fundamental shift in established risk modeling practices, moving from task-specific model training and calibration to the use of pretrained foundation models and \gls{icl}. Given these consequences and the prevalence of predictive modeling in credit risk, we consider a holistic evaluation of the new paradigm in canonical risk modeling tasks imperative. 

\Needspace{8\baselineskip}
\subsection{Deep learning for tabular data}
\label{subsec:deep_learning_for_tabular_data}
The study follows prior work and established industry practices of approaching \gls{pd} and \gls{lgd} modeling as classification and regression problems, respectively. Consequently, we use supervised \gls{ml} algorithms for labeled tabular data \citep{2016.BAESENS.creditriskanalyticsmeasurementtechniquesapplicationsexamplessas}.

\Gls{dl} has revolutionized various data modalities and represents the uncontested state-of-the-art in \gls{nlp} and computer vision, to name only a few \citep{2015.LECUN.deeplearning}. Given that natural language is a specific type of sequential data, it is not surprising that \gls{dl} has also become a \textit{go-to method} for time series forecasting \citep{benidis2022deep}. With respect to cross-sectional tabular data, however, classic \gls{ml} methods, particularly \glspl{gbm}, may be considered the de facto standard. Benchmarking experiments have shown their superiority over \gls{dl}-based approaches in that the former often provide more accurate predictions at lower computational costs and/or with less need for hyperparameter tuning \citep{2022.SHWARTZ-ZIV.tabulardatadeeplearningnotallyouneed,2022.GRINSZTAJN.whytreebasedmodelsstilloutperformdeeplearningtypicaltabulardata}. \citet{shmuel2025comprehensive} contextualize this finding by elaborating on the specific conditions (e.g., dataset characteristics) when \gls{dl} excels. Considering \gls{pd} modeling, however, \citet{2021.GUNNARSSON.deeplearningcreditscoringdont} argue that the potential of \gls{dl} lies in unlocking non-standard data sources (e.g., text) to augment risk models rather than serving as a vehicle for model estimation. 

We argue that the \citet{2021.GUNNARSSON.deeplearningcreditscoringdont} study represents the credit risk modeling space well, in particular for tabular data. \gls{dl} has provided excellent results in use cases involving semi-structured and unstructured data modalities \citep{bravo2026deep}, such as text data or networks \citep[e.g.,][]{stevenson2021value,zandi2025attention}, and settings that employ time-varying covariates \citep[e.g.,][]{korangi2023transformer,medinaolivares2024deep}, whereas \gls{gbm}-type algorithms, often paired with a post-hoc \gls{xai} approach, remain the standard solution when the goal is to develop highly accurate risk scorecards.\footnote{We do not mean to imply that \gls{gbm}-type approaches, or any \gls{ml}-based approach, can be considered an industry standard. Linear models are the only approach that can potentially claim this role.}

The unmatched success of \gls{dl} in unstructured data settings stands in sharp contrast to its moderate performance on tabular data, which has inspired much recent work aimed at closing the performance gap with \gls{gbm}-type approaches. 
Given that attention-based transformer networks were instrumental to the success of \gls{dl} in unstructured data settings, unlocking the power of attention for tabular data was a natural next step. 

The work by \citet{2021.GORISHNIY.revisitingdeeplearningmodelstabulardata}, introducing \gls{ftt}, may be seen as a seminal paper paving the way for more advanced \gls{dl}-based tabular data models. The authors demonstrate effective ways to utilize attention when working with tabular data, which is essential because it enables models to flexibly capture heterogeneous feature interactions common in tabular domains. Unlike convolutional or recurrent layers, attention can dynamically weight dependencies across diverse variables, reducing the need for manual feature engineering. Building on \gls{ftt}, subsequent work has refined attention variants for tabular data, confirming its role as a central design principle in closing the performance gap with \glspl{gbm} \citep{ye2024closer}. 

\citet{2026.JIANG.representationlearningtabulardatacomprehensivesurvey} provide a comprehensive survey of \gls{dl} models for tabular data and their evolution. The authors identify three main evolutionary stages of deep tabular learning. The first stage includes approaches similar to \gls{ftt}, which are trained and evaluated within a single distribution and focus on feature-level encoding, sample-level interactions, and objective-driven regularization \citep[e.g.,][]{huang2020tabtransformer,2021.ARIK.tabnetattentiveinterpretabletabularlearning}. These approaches introduced innovations such as feature tokenization and attention-based feature selection, but their generalization remained limited to the dataset at hand.

The second stage, transferable models, decouples representation learning from the downstream task, pretraining a network before fine-tuning it for prediction. One branch relies on self-supervised objectives that improve label efficiency but stay within a single table's feature space \citep[e.g.,][]{yoon2020vime,ucar2021subtab,somepalli2021saint}. A second branch confronts the feature- and label-space heterogeneity that blocks transfer across tables: TabLLM serializes each row into text, namely feature names, values, and a task description, recasting prediction as language generation and enabling \glspl{llm} in a few-shot mode \citep{hegselmann2023tabllm}.

The most recent, third stage is marked by foundation models for tabular data, which provide zero-shot predictive capabilities across diverse domains without the need for task-specific fine-tuning. Within this category, \glspl{pfn} stand out as a pioneering approach that harnesses (synthetic) pretraining and \gls{icl} with transformers to deliver robust generalization across a wide range of tabular problems \citep{2023.HOLLMANN.tabpfntransformerthatsolvessmalltabularclassificationproblemssecond,2025.QU.tabicltabularfoundationmodelincontextlearninglargedata,2025.MA.tabdptscalingtabularfoundationmodelsrealdata,2025.ZHANG.mitramixedsyntheticpriorsenhancingtabularfoundationmodels}. A defining architectural distinction separates these \glspl{tfm} from their language-model counterparts: whereas state-of-the-art \glspl{llm} rely on autoregressive \emph{decoders} to generate tokens sequentially under a causal mask, \glspl{tfm} utilize transformer \emph{encoders} that attend bidirectionally across the entire set of rows in a single pass. This design mirrors the structure of tabular data, in which rows carry no inherent sequential order. To achieve their zero-shot capabilities, \glspl{pfn} approximate Bayesian inference by pretraining on a large corpus of datasets, which are often synthetic and generated from \glspl{scm}, though in some cases derived from real-world data \citep{2025.MA.tabdptscalingtabularfoundationmodelsrealdata}, thereby endowing the transformer with a learned prior that enables strong \gls{icl} capabilities. Introducing TabPFNv2, the landmark paper by \citet{2025.HOLLMANN.accuratepredictionssmalldatatabularfoundationmodel} demonstrates the generality, predictive capability, and scalability of a \gls{pfn}-based zero-shot predictor. The authors conduct comprehensive empirical comparisons to demonstrate that their approach outperforms \gls{gbm} algorithms across a large set of regression and classification datasets with up to $10{,}000$ samples. However, because \gls{icl} requires the transformer to process the entire dataset simultaneously, TabPFNv2 was computationally constrained to moderately sized tables ($10{,}000$ samples). This scalability bottleneck has driven rapid subsequent innovations, culminating in architectures like TabPFN-3 and TabICLv2, which extend the operational capacity to $1{,}000{,}000$ observations on a single \gls{gpu} \citep{2026.GRINSZTAJN.tabpfn3technicalreport,2026.QU.tabiclv2betterfasterscalableopentabularfoundationmodel}. Foundation models for financial prediction are also beginning to extend beyond static tabular data. PRAGMA, for example, learns reusable representations from raw, multi-source banking event histories and reports strong relative improvements on internal tasks including credit scoring and fraud detection \citep{2026.OSTROUKHOV.pragmarevolutfoundationmodel}. This development is complementary to the \gls{pfn} paradigm considered here: PRAGMA targets longitudinal behavioral records, whereas our study evaluates zero-shot foundation models for canonical static \gls{pd} and \gls{lgd} tables.

The three stages illustrate how the field of deep tabular learning has shifted from approaches that adopt the classic \gls{ml} paradigm of training task-specific models to increasingly versatile architectures, which mimic the generality of contemporary \gls{ai} models. It is this generality, achieved through \gls{icl} and pretraining on large data collections, which we consider a \textit{paradigm shift} with potentially far-reaching implications for credit risk modeling. This prospect motivates the focal study, in which we undertake a comprehensive empirical evaluation of \glspl{pfn} for credit risk modeling. 

We acknowledge that our work does not advance deep tabular data learning methodology. Instead, we aim to contribute original empirical insights into how that field's latest innovations compare with established domain standards in credit risk modeling. Such evidence is valuable because the evaluation of \glspl{pfn} has not moved beyond the standard benchmarking datasets routinely used in \gls{ml}. These dataset collections are comprehensive and aim to be general by combining data from various fields. While clearly effective for demonstrating the potential of a new approach, decision-makers may appreciate targeted results that reflect the characteristics of their domain. Thus, our study provides a critical test of whether \glspl{pfn} can move from promising benchmarks to delivering real impact in credit risk modeling.

\clearpage
\section{Prior-data fitted networks}
\label{sec:prior-data_fitted_networks_pfns}
This section introduces the conceptual background of \glspl{pfn}. We first summarize their Bayesian intuition: a single pretrained model learns to make predictions from context examples by using regularities encoded in a pretraining prior. We then discuss TabPFN as the first tabular instantiation of this idea and review the architectural developments that made later \glspl{pfn} applicable to larger and more heterogeneous datasets. Finally, we explain how synthetic, real, and hybrid pretraining shape what these models can transfer to \gls{pd} and \gls{lgd} modeling, before summarizing the main variants in Table~\ref{tab:tfm_architectures}.

\Needspace{8\baselineskip}
\subsection{Theoretical foundations}
\label{subsec:theoretical_foundations}

\glspl{pfn} are pretrained models that approximate the Bayesian posterior predictive distribution at inference time, a paradigm first established in \emph{``Transformers Can Do Bayesian Inference''} by \citet{2021.MULLER.transformerscanbayesianinference}. The central idea is to train one transformer, denoted by $q_\theta$, not on one dataset, but across many supervised learning tasks sampled from a prior. In this sense, the problem is framed as meta-learning across datasets rather than learning from individual observations within a single dataset.

The prior encodes assumptions about plausible input--output relationships through a probability distribution $p(\phi)$ over data-generating mechanisms $\phi$, where each $\phi$ determines how features $x$ arise and how targets $y$ depend on them. Synthetic pretraining tasks are generated by first sampling a mechanism, $\phi \sim p(\phi)$, and then sampling a dataset from it, $\mathcal{D} \sim p(\mathcal{D}\mid \phi)$. 

For each sampled dataset, part of the data is used as labeled context and part is held out as query examples. Let
\[
\mathcal{D}_{\text{context}}=\{(x_i,y_i)\}_{i=1}^{n}
\]
denote the context set, and let $x_\ast$ denote a query point whose target $y_\ast$ is hidden. Under the pretraining prior $p(\phi)$, the Bayesian object of interest is the posterior predictive distribution
\[
p(y_\ast \mid x_\ast,\mathcal{D}_{\text{context}}),
\]
that is, the distribution of plausible target values $y_\ast$ conditional on $x_\ast$ and on the observed context set $\mathcal{D}_{\text{context}}$. During pretraining, the network $q_\theta$ repeatedly observes such context--query splits and learns to predict the hidden query targets from the context examples. After pretraining, the same learned predictor $q_\theta$ can be applied to a new dataset in a single forward pass, without task-specific parameter updates. In this sense, \glspl{pfn} amortize Bayesian inference: the costly learning of how to infer from data is shifted to the offline pretraining phase. In the limit of infinitely many sampled pretraining tasks, sufficient model capacity, and successful optimization, the learned predictor would recover the Bayes-optimal posterior predictive distribution implied by the prior \citep{2021.MULLER.transformerscanbayesianinference}. In practice, performance depends on finite model capacity, optimization, and the match between the pretraining prior and the empirical structure of the downstream data.

\clearpage
\Needspace{8\baselineskip}
\subsection{TabPFN: the tabular instantiation} 
\label{subsec:tabpfn_the_tabular_instantiation}

Whereas \citet{2021.MULLER.transformerscanbayesianinference} established the general principle that a transformer can approximate Bayesian inference from tasks sampled from a prior, \citet{2023.HOLLMANN.tabpfntransformerthatsolvessmalltabularclassificationproblemssecond} instantiated this idea for small tabular classification problems by designing a tabular pretraining prior and introducing TabPFN, the first tabular instantiation of the \gls{pfn} paradigm.

In the notation introduced above, TabPFN specifies the prior $p(\phi)$ as a mixture over \glspl{scm} and \glspl{bnn}, which serve as synthetic tabular data-generating mechanisms. Each sampled mechanism generates a synthetic dataset $\mathcal{D} \sim p(\mathcal{D}\mid\phi)$. Across millions of such datasets, TabPFN learns to infer the relationship between features and targets from labeled context examples and to predict the labels of held-out query examples \citep{2023.HOLLMANN.tabpfntransformerthatsolvessmalltabularclassificationproblemssecond}.

Because a \gls{pfn} approximates the posterior predictive distribution of its pretraining prior, both the Bayesian optimality of its predictions and the calibration that accompanies it hold only for tasks whose data-generating structure is adequately captured by that prior \citep{2021.MULLER.transformerscanbayesianinference,2023.HOLLMANN.tabpfntransformerthatsolvessmalltabularclassificationproblemssecond}. In real-world settings such as credit risk modeling, any systematic discrepancy between the synthetic pretraining assumptions and actual credit-risk data-generating processes, known as prior misspecification, can therefore degrade empirical calibration and accuracy on real data. A central lever for downstream performance is thus not only the model architecture but also the structural realism and diversity of the data-generating prior. This concern is acute in credit risk modeling, where borrower populations, default processes, recovery dynamics, and the bimodal loss distributions typical of \gls{lgd} modeling may differ substantially from the synthetic mechanisms used during pretraining. Whether a synthetic tabular prior transfers to \gls{pd} and \gls{lgd} modeling is therefore an empirical question, which motivates our domain-specific benchmark.

This prior dependence motivates empirical evaluation, but it does not diminish the practical appeal of the paradigm, for which two properties are salient. First, TabPFN performs \gls{icl}: it reuses the pretrained network to predict the labels for query observations of a new dataset in a single forward pass, without task-specific gradient training or hyperparameter tuning. Second, it can be applied to tabular datasets with varying numbers of observations and features, allowing the same model class to be used across heterogeneous credit portfolios. This flexibility is relevant for lenders, whose datasets often differ substantially in size, dimensionality, class balance, and distributional characteristics.

\Needspace{8\baselineskip}
\subsection{Architectural developments}
\label{subsec:architectural_developments}

The original TabPFN encodes each row, that is, a feature vector together with its label, as a single token and applies attention across the full set of context and query examples \citep{2023.HOLLMANN.tabpfntransformerthatsolvessmalltabularclassificationproblemssecond}. This delivers predictions for a new table in one forward pass, but makes the cost of attention grow quadratically in the number of rows. Scaling \gls{icl} to the larger and more heterogeneous tables found in real settings therefore became a central architectural challenge, which the literature has tackled with three broad strategies.

The first strategy keeps the model conditioned on the entire table but redesigns how attention is applied. In place of one token per row, TabPFNv2 assigns a representation to each cell and uses two-way attention, with each cell attending to the other features in its row and then to the same feature across all rows \citep{2025.HOLLMANN.accuratepredictionssmalldatatabularfoundationmodel}. This adds native support for regression, categorical features, and missing values, and lifts the supported scale to datasets of up to $10{,}000$ rows and $500$ features. TabPFN-2.5 retains this cell-level design but uses a deeper network and a richer synthetic prior, reaching up to $50{,}000$ rows and $2{,}000$ features \citep{2026.GRINSZTAJN.tabpfn25advancingstatearttabularfoundationmodels}. Mitra reuses the same two-way attention design and places its contribution in the prior, showing that its mixture of synthetic priors is model-agnostic, improving both \gls{oned} row attention and \gls{twod} cell attention \citep{2025.ZHANG.mitramixedsyntheticpriorsenhancingtabularfoundationmodels}.

The second strategy, introduced by TabICL, treats even two-way attention as too costly at scale and compresses each row before performing \gls{icl} \citep{2025.QU.tabicltabularfoundationmodelincontextlearninglargedata}. A first stage embeds each row with a column-wise Set transformer followed by row-wise attention, collapsing the feature dimension into a fixed-length vector, and a second stage runs dataset-wise \gls{icl} over these vectors rather than over raw cells. This raises the limit to $500{,}000$ rows. Its successor, TabICLv2, adds regression and reaches one million rows \citep{2026.QU.tabiclv2betterfasterscalableopentabularfoundationmodel}. The two lines then converge: TabPFN-3 abandons the alternating cell-level attention of its predecessors, which it likewise finds too costly as datasets grow, returns to row-level \gls{icl}, and adopts TabICL's two-stage row compression to scale to one million rows along a cell-budget frontier that trades rows against features \citep{2026.GRINSZTAJN.tabpfn3technicalreport}. The third strategy, taken by TabDPT, retrieves only each query's nearest rows as context, so the total dataset size no longer directly determines the attention cost \citep{2025.MA.tabdptscalingtabularfoundationmodelsrealdata}.

\Needspace{8\baselineskip}
\subsection{Pretraining data and prior design}
\label{subsec:pretraining_data_and_prior_design}

Beyond architectural scalability, a second design axis is the pretraining prior. Because the prior shapes the statistical patterns a \gls{pfn} learns to recognize and reuse on new tasks, prior design is arguably at least as important as architecture. Many subsequent \gls{pfn} variants therefore focus not only on scaling the model, but also on redesigning the synthetic or real-data distributions used during pretraining. Most models relied entirely on synthetic data-generating mechanisms. The original TabPFN sampled from a mixture of \glspl{scm} and \glspl{bnn} \citep{2023.HOLLMANN.tabpfntransformerthatsolvessmalltabularclassificationproblemssecond}, whereas TabPFNv2 used richer \glspl{scm} with more elaborate causal graphs and native generation of categorical features and missing values \citep{2025.HOLLMANN.accuratepredictionssmalldatatabularfoundationmodel}. TabICL extended this idea by incorporating tree-based \glspl{scm}, using XGBoost as a generating function to inject tree-based inductive biases \citep{2025.QU.tabicltabularfoundationmodelincontextlearninglargedata}. Mitra further systematized prior design by training on a curated mixture of \gls{scm} and tree-based priors, including \gls{gb}, \glspl{rf}, and \glspl{dt}, to improve sample efficiency \citep{2025.ZHANG.mitramixedsyntheticpriorsenhancingtabularfoundationmodels}. The common lesson is that no single synthetic generator is likely to cover the diversity of real tabular tasks, motivating increasingly diverse priors in TabICLv2, TabPFN-2.5, and TabPFN-3.

A complementary line of work introduces real datasets into pretraining. Real-TabPFN starts from synthetic-pretrained TabPFNv2 and continues pretraining on a curated collection of real OpenML and Kaggle tables, improving off-the-shelf classification performance without changing the underlying architecture \citep{2025.GARG.realtabpfnimprovingtabularfoundationmodelscontinuedpretrainingrealworlddata}. TabDPT goes further by pretraining on real data only, combining self-supervised column masking with retrieval-based context selection \citep{2025.MA.tabdptscalingtabularfoundationmodelsrealdata}. These approaches reflect the same underlying motivation: real tables may contain statistical structures that current synthetic priors fail to capture. For credit risk modeling, this issue is particularly important because the effectiveness of \glspl{pfn} depends on the extent to which their pretraining data---whether synthetic, real, or hybrid---reflect the empirical structure of \gls{pd} and \gls{lgd} datasets. Appendix~\ref{supp:data_leakage} audits direct overlap between our benchmark datasets and the documented pretraining corpora of Real-TabPFN and TabDPT.

\Needspace{8\baselineskip}
\subsection{Overview of PFNs}
\label{subsec:overview_of_pfns}
Table~\ref{tab:tfm_architectures} summarizes the main \gls{pfn} architectures discussed above, grouped by lineage and compared in terms of task support, pretraining prior, \gls{icl} architecture, and reported inference limits.

\begin{table}[ht!]
    \centering
    \caption{Overview of \glspl{pfn}, grouped by lineage.
        \emph{Max rows} and \emph{Max feat.} indicate the largest numbers of rows and features, respectively, that the model is built to handle at inference, as reported by the respective authors; these limits are soft.
        \protect\\ \mbox{C = classification}, \mbox{R = regression},
        \mbox{\gls{ssl}}.}
    \label{tab:tfm_architectures}
    \scriptsize
\setlength{\tabcolsep}{3pt}
\renewcommand{\arraystretch}{1.2}

    \begin{tabular}{@{}>{\raggedright\arraybackslash}p{0.18\textwidth} >{\raggedright\arraybackslash}p{0.05\textwidth} >{\raggedright\arraybackslash}p{0.16\textwidth} >{\raggedright\arraybackslash}p{0.368\textwidth} >{\raggedleft\arraybackslash}p{0.095\textwidth} >{\raggedleft\arraybackslash}p{0.08\textwidth}@{}}
        \toprule
        \textbf{Model} & \textbf{Tasks} & \textbf{Pretraining data / prior} & \textbf{\gls{icl} architecture} & \textbf{Max rows} & \textbf{Max feat.} \\
        \midrule
        \multicolumn{6}{l}{\textit{TabPFN line (Prior Labs)}} \\
        TabPFN \citep{2023.HOLLMANN.tabpfntransformerthatsolvessmalltabularclassificationproblemssecond} & C & Synthetic: \gls{scm} $+$ \gls{bnn} & Row $=$ token; \gls{oned} row-wise attn. & 1{,}000 & 100 \\
        TabPFNv2 \citep{2025.HOLLMANN.accuratepredictionssmalldatatabularfoundationmodel} & C, R & Synthetic: \gls{scm} & Cell $=$ token; \gls{twod} alternating attn.; pairs grouped & 10{,}000 & 500 \\
        Real-TabPFN \citep{2025.GARG.realtabpfnimprovingtabularfoundationmodelscontinuedpretrainingrealworlddata} & C & Hybrid: \gls{scm} $\to$ real (cont.\ pretrain.) & Cell $=$ token; \gls{twod} alternating (TabPFNv2 backbone, unchanged) & 10{,}000 & 500 \\
        TabPFN-2.5 \citep{2026.GRINSZTAJN.tabpfn25advancingstatearttabularfoundationmodels} & C, R & Synthetic: \gls{scm} (enriched) & Cell $=$ token; \gls{twod} alternating; triplets grouped, deeper net & 50{,}000 & 2{,}000 \\
        TabPFN-3 \citep{2026.GRINSZTAJN.tabpfn3technicalreport} & C, R & Synthetic: \gls{scm} (enriched) & Compress-first (col.\ embed $\to$ row compress $\to$ row \gls{icl}); triplets & 1{,}000{,}000\textsuperscript{a} & 200\textsuperscript{a} \\
        \midrule
        \multicolumn{6}{l}{\textit{TabICL line}} \\
        TabICL \citep{2025.QU.tabicltabularfoundationmodelincontextlearninglargedata} & C & Synthetic: \gls{scm} $+$ tree-based \gls{scm} & Compress-first (col.\ embed $\to$ row compress $\to$ row \gls{icl}) & 500{,}000 & 500\textsuperscript{c} \\
        TabICLv2 \citep{2026.QU.tabiclv2betterfasterscalableopentabularfoundationmodel} & C, R & Synthetic: \gls{scm} $+$ tree-based (redesigned engine) & Compress-first; triplets grouped, target-aware embed., \gls{qassmax} & 1{,}000{,}000 & 500\textsuperscript{c} \\
        \midrule
        \multicolumn{6}{l}{\textit{Others}} \\
        Mitra \citep{2025.ZHANG.mitramixedsyntheticpriorsenhancingtabularfoundationmodels} & C, R & Synthetic: \gls{scm} $+$ tree-based (\gls{gb}, \gls{rf}, \gls{dt}) & Cell $=$ token; \gls{twod} alternating (priors are architecture-agnostic) & 10{,}000 & 500 \\
        TabDPT \citep{2025.MA.tabdptscalingtabularfoundationmodelsrealdata} & C, R & Real: OpenML; \gls{ssl} (column masking) & Row $=$ token; \gls{oned} row-wise $+$ \gls{knn} retrieval & Unbounded\textsuperscript{b} & 100\textsuperscript{b} \\
        \bottomrule
        \multicolumn{6}{@{}p{\textwidth}@{}}{\scriptsize \textsuperscript{a} Soft cell budget: 1M rows @ 200 feat., \emph{or} 100k @ 2{,}000, \emph{or} 1k @ 20{,}000.}\\
        \multicolumn{6}{@{}p{\textwidth}@{}}{\scriptsize \textsuperscript{b} Per-query \gls{knn} retrieval, so training-set size is unbounded; features beyond $F_{\max}{=}100$ are \gls{pca}-reduced.}\\
        \multicolumn{6}{@{}p{\textwidth}@{}}{\scriptsize \textsuperscript{c} Pretrained on ${\leq}100$ features; up to 500 handled at inference.}\\
    \end{tabular}
\end{table}

\clearpage
\section{Experimental design}
\label{sec:experimental_design}

\Needspace{8\baselineskip}
\subsection{Research questions}
\label{subsec:research_questions}

To systematically assess the viability of \gls{pfn}-based \glspl{tfm} as a novel paradigm for credit risk modeling, our empirical study is guided by three core \glspl{rq}. These questions are designed to evaluate both the general predictive capabilities of these models and their robustness under structural constraints typical of financial datasets.

\begin{enumerate}
    \item[\textbf{\gls{rq} $1$}] \textbf{Predictive competitiveness:} To what extent do these models achieve competitive predictive performance in core credit risk modeling tasks, namely \gls{pd} classification and \gls{lgd} regression?
    
    \item[\textbf{\gls{rq} $2$}] \textbf{Robustness to data scarcity:} How does model performance evolve with dataset size? Specifically, do pretrained \glspl{tfm} attain stronger absolute predictive performance than conventional learners at common small-sample budgets?
    
    \item[\textbf{\gls{rq} $3$}] \textbf{Robustness to class imbalance:} How robust are these models to severe class imbalance---a common challenge in \gls{pd} modeling associated with low-default portfolios? 
\end{enumerate}

By addressing these questions, the subsequent experiments aim to identify not only whether \glspl{tfm} are competitive, but also under which specific portfolio characteristics they are most beneficial.

\Needspace{8\baselineskip}
\subsection{Data}
\label{subsec:data}
We evaluate \glspl{pfn} for \gls{pd} and \gls{lgd} estimation, corresponding to classification and regression tasks, respectively. Our experiments use both public and proprietary datasets spanning retail and small-business portfolios. Table~\ref{tab:data-summary} summarizes the key characteristics of the \gls{pd} and \gls{lgd} datasets, including sample size, dimensionality, minority class proportion, and data origin.

Our datasets vary substantially in the number of observations, the number of features, and the distribution of the target variable. For \gls{pd}, the default rate differs substantially across portfolios, ranging from approximately $6.7\%$ to $40.0\%$ (mean~$\approx 22.1\%$, median~$\approx 22.4\%$),  reflecting the heterogeneity of credit products and their associated risk, from secured mortgages to unsecured consumer loans and subprime credit.

\begin{figure}[ht!]
  \centering
  \includegraphics[width=\textwidth,height=0.90\textheight,keepaspectratio]{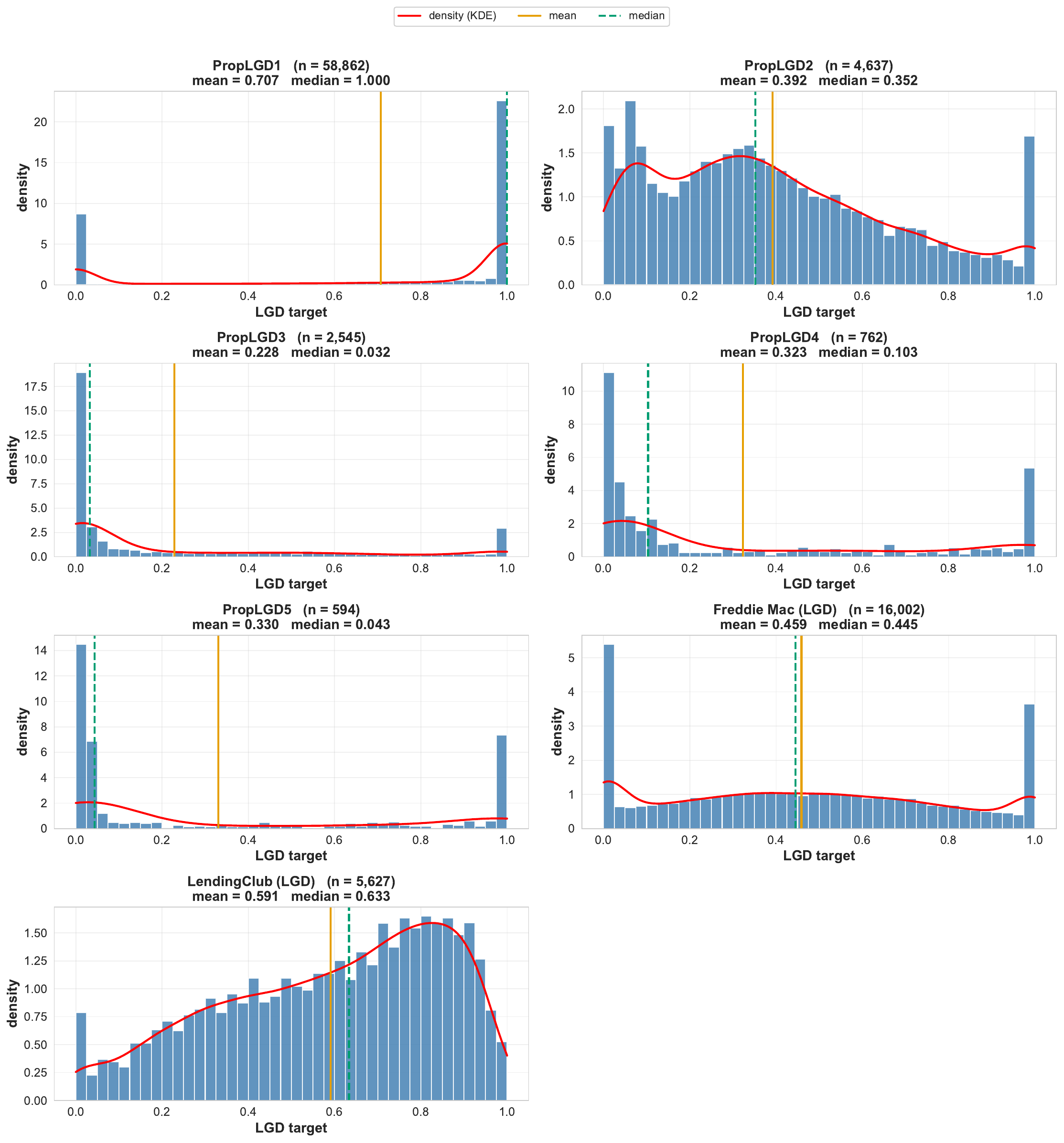}
  \caption{Distribution of the \gls{lgd} target in each of the seven \gls{lgd} datasets, after clipping to the unit interval. Red curves show kernel-density estimates; vertical lines mark the mean and median in each panel.}
  \label{fig:lgd_target_hists}
\end{figure}

Figure~\ref{fig:lgd_target_hists} shows the distribution of the target variable for each \gls{lgd} dataset. Many datasets exhibit the typical bimodal, zero-one-inflated distribution, with substantial probability mass at or near total recovery (\gls{lgd} close to zero) and at or near total loss (i.e., \gls{lgd} equal to or close to one). This bimodality of the \gls{lgd} distribution arises due to, on the one hand, the high risk of losing the entire outstanding amount in the event of default for unsecured exposures, while for secured loans, on the other hand, the full outstanding amount at time of default is often recovered. Compared to previous work of \citet{2012.LOTERMAN.benchmarkingregressionalgorithmslossgivendefaultmodeling}, our data collection exhibits more heterogeneity in the \gls{lgd} distribution: some portfolios are dominated by a single mode, whereas others place comparable mass at both boundaries. 

\begin{table}[ht]
\centering
\footnotesize
\setlength{\tabcolsep}{5pt}
\renewcommand{\arraystretch}{0.9}
\caption{Characteristics of the \gls{pd} and \gls{lgd} datasets. Minority \% represents the positive class rate (default) for \gls{pd} datasets and is undefined for \gls{lgd} datasets. Variable counts refer to the cleaned data before fold-wise feature filtering and, where applicable, \gls{pca}.}
\label{tab:data-summary}
\begin{tabular}{@{}>{\raggedright\arraybackslash}p{0.055\textwidth}>{\raggedright\arraybackslash}p{0.255\textwidth}>{\raggedleft\arraybackslash}p{0.105\textwidth}>{\centering\arraybackslash}p{0.07\textwidth}>{\raggedleft\arraybackslash}p{0.095\textwidth}>{\raggedright\arraybackslash}p{0.31\textwidth}@{}}
\toprule
ID & Dataset name & Observations & Variables & Minority \% & Source \\ \midrule
PD1  & Cobranded                  & 80,000  & 47    & 24.6\% & \href{https://www.kaggle.com/datasets/thedevastator/predicting-co-branded-credit-card-defaults-in-re}{Kaggle} \\
PD2  & German Credit              & 1,000   & 20    & 30.0\% & \href{https://archive.ics.uci.edu/dataset/144/statlog+german+credit+data}{\gls{uci}} \\
PD3  & \gls{gmsc} & 150,000 & 10    & 6.7\%  & \href{https://www.kaggle.com/c/GiveMeSomeCredit}{Kaggle} \\
PD4  & HackerEarth                & 532,428 & 35    & 23.6\% & \href{https://www.hackerearth.com/problem/machine-learning/bank-fears-loanliness/}{HackerEarth} \\
PD5  & \gls{hmeq}                       & 5,960   & 12    & 19.9\% & \href{http://www.creditriskanalytics.net}{creditriskanalytics.net} \\
PD6  & Home Credit                & 307,511 & 120   & 8.1\%  & \href{https://www.kaggle.com/c/home-credit-default-risk}{Kaggle} \\
PD7  & LendingClub (\gls{pd})           & 9,578   & 13    & 16.0\% & \href{https://www.kaggle.com/datasets/saramah/loan-data}{Kaggle} \\
PD8  & MicroFinance               & 158,700 & 2,986 & 37.8\% & \citep{kozodoi2025fighting} \\
PD9  & Myhom                      & 7,000   & 8     & 40.0\% & \href{https://www.kaggle.com/datasets/gauravduttakiit/loan-default-prediction}{Kaggle} \\
PD10 & Taiwan Credit Card         & 30,000  & 23    & 22.1\% & \href{https://archive.ics.uci.edu/dataset/350/default+of+credit+card+clients}{\gls{uci}}; \citep{yeh2009comparisons} \\
PD11 & Thomas                     & 1,225   & 14    & 26.4\% & \citep{thomas2002credit} \\
PD12 & Vehicle Loan               & 233,154 & 35    & 21.7\% & \href{https://www.kaggle.com/datasets/mamtadhaker/lt-vehicle-loan-default-prediction}{Kaggle} \\
PD13 & PropPD1                    & 100,000 & 16    & 22.6\% & Proprietary \\
PD14 & PropPD2                    & 105,471 & 759   & 9.3\%  & Proprietary \\ \midrule
LGD1 & Freddie Mac (\gls{lgd})          & 16,002  & 20    &        & \href{https://www.freddiemac.com/research/datasets/sf-loanlevel-dataset}{Freddie Mac} \\
LGD2 & LendingClub (\gls{lgd})          & 5,627   & 17    &        & \href{https://www.kaggle.com/datasets/wordsforthewise/lending-club}{Kaggle} \\
LGD3 & PropLGD1                   & 58,862  & 9     &        & Proprietary \\
LGD4 & PropLGD2                   & 4,637   & 52    &        & Proprietary \\
LGD5 & PropLGD3                   & 2,545   & 2     &        & Proprietary \\
LGD6 & PropLGD4                   & 762     & 202   &        & Proprietary \\
LGD7 & PropLGD5                   & 594     & 256   &        & Proprietary \\
\bottomrule
\end{tabular}
\end{table}

A serious risk when benchmarking foundation models using publicly available data concerns leakage. If the pretraining corpus of a foundation model included some of our benchmarking datasets, this model would face an unfair advantage over the other contestants. To assess this risk, we audit the pretraining protocols of the foundation models employed, and report detailed results in Appendix~\ref{supp:data_leakage}. The key finding of the analysis is threefold: i) most foundation models were pretrained exclusively on synthetic data, ii) two models were pretrained partly (Real-TabPFN) or wholly (TabDPT) on real data and warrant closer inspection, iii) the overlap between the pretraining corpus of the two affected models is small. The empirical results reported in  Section~ \ref{subsec:pd_and_lgd_modeling_benchmark} further shows that the two affected approaches perform much worse than many alternative learners, far away from the leading group. Therefore, we conclude that there is no material risk of leakage impeding the benchmarking results, and refer readers to Appendix~\ref{supp:data_leakage} for details. 

\Needspace{8\baselineskip}
\subsection{Data preprocessing}
\label{subsec:data_preprocessing}
We use \gls{talent} \citep{2025.LIU.talenttabularanalyticslearningtoolbox} as the benchmarking backbone for the experiments. \gls{talent} provides a unified interface, ensuring that heterogeneous learners are compared under consistent data-handling conventions. However, we perform dataset-specific cleaning before passing a dataset through the common modeling pipeline. This one-off curation step consists of deterministic operations that do not interfere with train/test splitting: removal of identifier or leakage-prone columns, dataset-specific transformations, and, for \gls{lgd}, clipping of the target to the valid unit interval $[0,1]$.\footnote{Only one of our datasets contains raw \gls{lgd} values outside the unit interval.}

The subsequent fold-wise pipeline applies the same rules across datasets. Within that pipeline, all learned preprocessing parameters (e.g., feature means) are estimated exclusively from the training partition of each cross-validation fold and applied to the corresponding validation and test partitions, thereby avoiding data leakage. Near-constant features are identified from the training partition and removed from all three partitions. Extreme numerical outliers are removed from the training partition only, while extreme validation and test values are winsorized using training-fold quantiles. For datasets with more than $99$ retained features, numerical and integer-encoded categorical features are combined, mean-imputed, and reduced to at most $99$ principal components, with both imputation and \gls{pca} fitted on the training partition. For the other datasets, we impute numerical missing values with the training-fold median, and assign a separate level to categorical missing values. We then encode categorical features as integer indices for methods that handle them natively, or via one-hot encoding for methods that require real-valued inputs. Finally, numerical features are standardized to zero mean and unit variance, except for the \glspl{tfm}, which receive the same fold-wise processed inputs without external standardization and apply their own internal normalization. Where required by a learner's architectural constraints, a method-specific cap is applied only to its training context; the validation and test partitions remain unchanged.

\Needspace{8\baselineskip}
\subsection{Benchmarking protocol}
\label{subsec:benchmarking_protocol}
To assess \gls{rq} $1$, we compare a diverse panel of competitive learners, ranging from classical statistical methods to state-of-the-art deep tabular architectures. For \gls{pd}, we compare $33$ methods, including \glspl{tfm}, several classical methods, particularly tree-based ensembles (XGBoost, LightGBM, CatBoost), and a wide array of \gls{dl} models, including transformers (e.g., \gls{ftt}, ExcelFormer) and specialized architectures such as TabNet and TabM. For \gls{lgd}, the benchmark comprises $27$ methods, including the foundation models that support regression, classical regression baselines, \gls{gb} approaches, and advanced \gls{dl} architectures such as ModernNCA, ResNet, and PTARL. Table~\ref{tab:learners} summarizes the full set of methods considered across both tasks.

\begin{table}[ht!]
\centering
\footnotesize
\setlength{\tabcolsep}{5pt}
\renewcommand{\arraystretch}{0.9}
\caption{Overview of benchmarked methods for \gls{pd} (classification) and \gls{lgd} (regression). Abbreviation: \gls{mlp}.}
\label{tab:learners}
\begin{tabular}{@{}lcc@{\hspace{2em}}lcc@{}}
\toprule
\textbf{Method} & \textbf{\gls{pd}} & \textbf{\gls{lgd}} & \textbf{Method} & \textbf{\gls{pd}} & \textbf{\gls{lgd}} \\
\midrule
\textit{\textbf{Foundation Models}}            &            &            & \textit{\textbf{Deep Tabular (\gls{mlp} \& Spec.)}} &            &            \\
TabPFN                               & \checkmark &            & \gls{mlp}                                  & \checkmark & \checkmark \\
TabPFNv2                             & \checkmark & \checkmark & ResNet                               &            & \checkmark \\
TabPFN-2.5                           & \checkmark & \checkmark & SNN                                  & \checkmark &            \\
TabPFN-3                             & \checkmark & \checkmark & RealMLP                              & \checkmark & \checkmark \\
Real-TabPFN                          & \checkmark &            & \gls{mlp}-PLR                              & \checkmark & \checkmark \\
TabICL                               & \checkmark &            & DANets                               & \checkmark &            \\
TabICLv2                             & \checkmark & \checkmark & SwitchTab                            & \checkmark &            \\
TabDPT                               & \checkmark & \checkmark & TabNet                               & \checkmark & \checkmark \\
Mitra                                & \checkmark & \checkmark & DCN2                                 & \checkmark & \checkmark \\
\textit{\textbf{Tree Boosting}}                &            &            & TabM                                 & \checkmark & \checkmark \\
CatBoost                             & \checkmark & \checkmark & TANGOS                               & \checkmark & \checkmark \\
LightGBM                             & \checkmark & \checkmark & ModernNCA                            &            & \checkmark \\
XGBoost                              & \checkmark & \checkmark & \textit{\textbf{Classical \gls{ml}}}                &            &            \\
\textit{\textbf{Deep Tabular (Transformers)}}  &            &            & Logistic Regression                  & \checkmark &            \\
\gls{ftt}                 & \checkmark & \checkmark & Linear Regression                    &            & \checkmark \\
AutoInt                              & \checkmark & \checkmark & \gls{knn}            & \checkmark & \checkmark \\
ExcelFormer                          & \checkmark & \checkmark & \gls{rf}                        & \checkmark & \checkmark \\
AMFormer                             & \checkmark &            & \makecell[l]{\gls{svm}\\\gls{svr}}     & \checkmark & \checkmark \\
T2G-Former                           & \checkmark & \checkmark & Naive Bayes                          & \checkmark &            \\
PTARL                                &            & \checkmark & \gls{ncm}             & \checkmark &            \\
\bottomrule
\end{tabular}
\end{table}

We evaluate performance on the held-out test partitions from five-fold cross-validation. For tunable learners, we select hyperparameters independently within each fold using Optuna \citep{2019.AKIBA.optunanextgenerationhyperparameteroptimizationframework} with $20$ trials on a validation split comprising $20\%$ of the training fold. We run \glspl{tfm} in their default in-context configuration without downstream hyperparameter tuning.

In the main text, we report one primary metric per task: \gls{auc} for \gls{pd}, because it evaluates ranking performance independently of a classification threshold, and $R^2$ for \gls{lgd}, because it is scale-normalized. Both are comparable across datasets that differ in size and class balance, and standard in the credit scoring benchmarking tradition \citep{2015.LESSMANN.benchmarkingstateoftheartclassificationalgorithmscreditscoringupdateresearch}. All remaining performance metrics are reported in Appendix~\ref{supp:metrics}. For \gls{pd} these comprise additional discrimination measures (Gini, the \gls{ks} statistic, and adjusted \gls{ap}, $\mathrm{AP}_{\mathrm{adj}}$), proper scoring rules (the Brier score and log loss), and threshold-based measures obtained by maximizing $F_1$ on the validation split ($F_1$, \gls{mcc}, accuracy, balanced accuracy, precision, and recall). Adjusted \gls{ap} accounts for the prevalence-dependent random-ranking reference of \gls{ap} and is defined in Appendix~\ref{supp:metrics}. For \gls{lgd} they comprise \gls{mse}, \gls{rmse}, \gls{mae}, \gls{medae}, maximum error, the Pearson and Spearman correlations, and explained variance. All \gls{lgd} predictions are clipped to the unit interval $[0,1]$ before evaluation. A formal statistical analysis, comprising rank-based omnibus tests, \gls{cd} diagrams, and pairwise comparisons, a Bayesian signed-rank analysis with a \gls{rope}, win/loss matrices, and a champion-level comparison is provided in Appendix~\ref{supp:stats}.

In addition to conventional performance measures, we assess mean calibration at the portfolio level using the pooled out-of-fold predictions for each dataset. For \gls{pd}, we compare the observed default rate with the mean predicted \gls{pd}; for \gls{lgd}, we compare the mean realized \gls{lgd} with the mean predicted \gls{lgd}. We report differences as observed minus predicted. Positive values therefore indicate underprediction, whereas negative values indicate overprediction. When summarizing results across datasets, each dataset receives equal weight.

\clearpage
\subsection{Auxiliary experiments}
\label{subsec:auxiliary_experiments}
A feature of particular interest for credit risk modeling concerns the ability of \glspl{tfm} to handle data scarcity. Therefore, we augment the main benchmark with two auxiliary experiments that probe \emph{how} the leading models behave when data regimes change. We reuse the main protocol --- five-fold cross-validation with the same train/validation/test splits and the same metrics --- but disable \gls{hpo} (default settings only) and restrict the pool to the strongest foundation models together with two reference baselines: CatBoost and a logistic or linear regression model. Crucially, every reduced dataset is built by \emph{nested, fixed-seed} subsampling: at each step, the same rows are kept for all methods, and smaller samples are strict subsets of larger ones, so the resulting trends reflect the change in data, rather than variation due to random sampling.

\Needspace{6\baselineskip}
\subsubsection{Small-data robustness} 
\label{subsec:small-data_robustness}
To respond to \gls{rq} $2$, the first auxiliary experiment assesses how learner performance evolves with dataset size. We cap the data to a target size \emph{before} splitting and sweep this size downward in fine increments (\gls{pd}: $30{,}000 \rightarrow 100$ in steps of $25$; \gls{lgd}: $4{,}600 \rightarrow 100$ in steps of $10$). The upper bound doubles as the inclusion threshold: only datasets with at least that number of observations are included in the experiment (9 for \gls{pd}, and 4 for \gls{lgd}), so every performance curve is defined over the full sweep of selected datasets. Because eligibility is determined by the upper sweep bound, this experiment evaluates nested subsamples of datasets that are natively large; it does not directly evaluate portfolios that are intrinsically small.

\Needspace{6\baselineskip}
\subsubsection{Class-imbalance robustness (PD only)} 
\label{subsec:class-imbalance_robustness_pd_only}
In the second auxiliary experiment, aimed at addressing \gls{rq} $3$, we assess how robust \glspl{tfm} are to class imbalance by progressively removing minority-class (default) observations. We reduce the positive rate from $0.15$ to $0.0025$ in steps of $0.0005$ ($\approx 300$ imbalance levels per dataset and method). We retain only datasets with a minority class proportion above $0.15$ and with at least $30{,}000$ observations, so that every dataset can be carried through the full sweep while still containing minority cases at the lowest imbalance setting. This results in $6$ datasets being included in this experiment. Note that, for each target rate, the minority class is subsampled across the entire dataset (training, validation, and test partitions). This design creates class imbalance artificially by removing default observations from otherwise larger portfolios; it does not reproduce all characteristics of naturally low-default portfolios. Appendix~\ref{supp:data_eligibility} reports the complete dataset-eligibility table and explains how the inclusion thresholds determine the experimental samples.

\clearpage
\section{Results and discussion}
\label{sec:results_and_discussion}

\Needspace{8\baselineskip}
\subsection{PD and LGD prediction benchmark}
\label{subsec:pd_and_lgd_modeling_benchmark}

We begin with the headline result of the benchmark. Across both prediction tasks, \gls{pd} (classification) and \gls{lgd} (regression), a \gls{tfm} achieves the best average performance, and in both cases, it is the same model: TabPFN-3. 
Recall that foundation models achieve the reported performance off the shelf: \gls{gb} and deep-tabular baselines are tuned through \gls{hpo}, whereas the foundation models are applied in their default in-context configuration, without gradient-based training or task-specific tuning. We consider it a notable achievement that an untuned, in-context learner surpasses carefully tuned boosting ensembles, the de facto standard in credit risk modeling. The improvement is small for \gls{pd} (an \gls{auc} of $0.764$ against CatBoost's $0.750$) but consistent, and substantial for \gls{lgd}, where the best foundation model raises $R^2$ from $0.487$ to $0.537$, a relative gain of roughly $10\%$. 

\begin{figure}[ht!]
 \centering
 \includegraphics[width=1.0\textwidth,height=0.90\textheight,keepaspectratio]{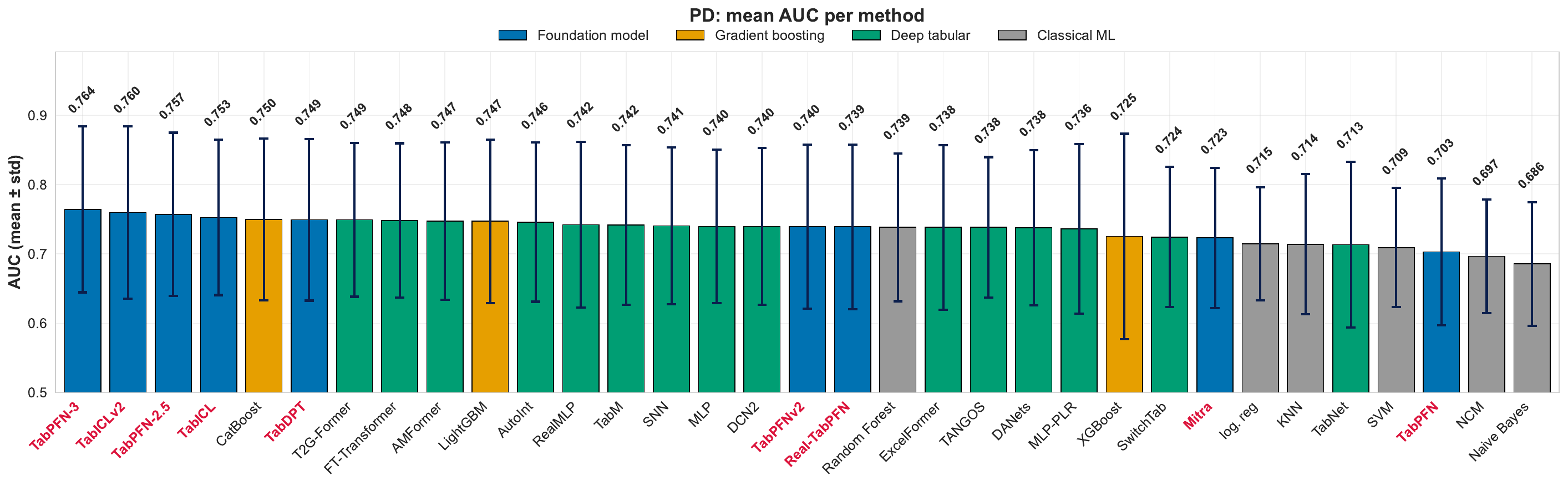}
\caption{Bars are ordered best to worst; whiskers show pooled fold-level standard deviations. Bar fill encodes the model class. Foundation models are highlighted in red.}
 \label{fig:pd_bar_mean_auc}
\end{figure}

Figure~\ref{fig:pd_bar_mean_auc} reports the average \gls{auc} of each classifier across the five folds of all \gls{pd} datasets. The four highest-ranked methods are foundation models: TabPFN-3 ($0.764$), TabICLv2 ($0.760$), TabPFN-2.5 ($0.757$), and TabICL ($0.753$). The strongest conventional learner is CatBoost ($0.750$), which the leading foundation models edge out. The absolute differences at the top of the ranking are small, but consistent, suggesting that \glspl{tfm} set a new benchmark. However, performance among foundation models is far from uniform. The most recent generation of models (TabPFN-3, TabICLv2, and TabPFN-2.5) forms the leading group, TabPFNv2 and its continued-pretraining variant Real-TabPFN sit mid-table, and Mitra and the original TabPFN rank among the weakest learners for default prediction. 

\begin{figure}[ht!]
 \centering
 \includegraphics[width=1\textwidth,height=0.90\textheight,keepaspectratio]{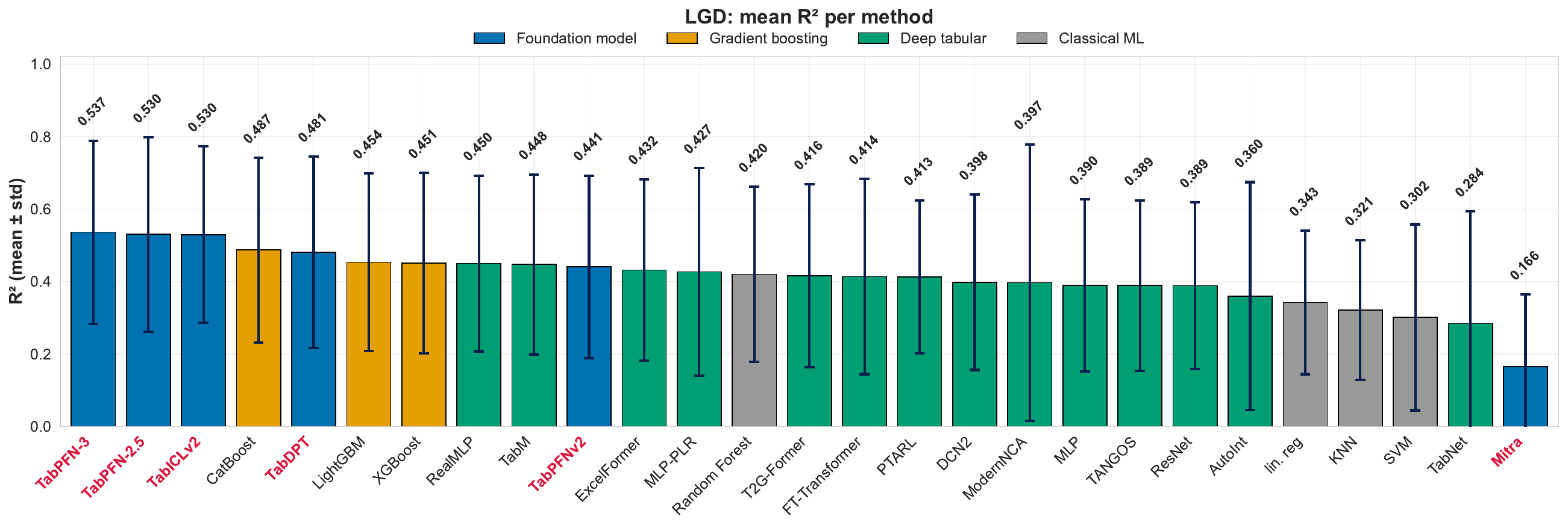}
\caption{Bars are ordered best to worst; whiskers show pooled fold-level standard deviations. Bar fill encodes the model class. Foundation models are highlighted in red.}
 \label{fig:lgd_bar_mean_r2}
\end{figure}

The same pattern, in sharper form, holds for \gls{lgd}. Figure~\ref{fig:lgd_bar_mean_r2} reports the average $R^2$ across \gls{lgd} datasets. TabPFN-3 again ranks first ($R^2=0.537$), followed by TabPFN-2.5 ($0.531$) and TabICLv2 ($0.530$). The three best methods are foundation models: the same three leading models as for \gls{pd} modeling. The first conventional learner is CatBoost ($0.487$), ranked fourth, followed by TabDPT ($0.481$) in fifth place. The margin over the boosting baselines is wider for \gls{lgd} than for \gls{pd}. This is notable because \gls{lgd} modeling is widely regarded as the harder problem, a difficulty usually attributed to the irregular bimodal shape of the target loss distributions. The within-segment heterogeneity is also stronger for \gls{lgd}: while the TabPFN and TabICL line dominates the top of the table, Mitra is the single weakest method overall ($R^2=0.166$).

The appendices augment the headline results of Figures~\ref{fig:pd_bar_mean_auc}~and~\ref{fig:lgd_bar_mean_r2}. Appendix~\ref{supp:stats} provides a full statistical analysis for our main performance indicators (\gls{auc} for \gls{pd} and $R^2$ for \gls{lgd}), with results of rank-based and pairwise tests on the significance of performance differences across all learners and selected champions. Appendix~\ref{supp:metrics} reports average (over datasets) results for other performance indicators, including Gini, \gls{ks}, adjusted \gls{ap} ($\mathrm{AP}_{\mathrm{adj}}$)) and the Brier score for \gls{pd}, and \gls{mse}, \gls{rmse}, \gls{mae}, and medium/maximum average errors for \gls{lgd}. Appendix~\ref{supp:viz} complements these aggregate results with per-dataset performance matrices and direct comparisons of TabPFN-3 with CatBoost and the linear baselines. 

Taken together, the observed results point to a shift in the state of the art in credit risk prediction. \gls{dt} boosting ensembles, long considered the strongest method, are narrowly surpassed for \gls{pd} and decisively outperformed for \gls{lgd} by \glspl{tfm}, which attain this performance without task-specific training or tuning. This advantage is most pronounced for \gls{lgd}, where the leading foundation models open a wide gap over boosting, and more modest for \gls{pd}, where the best methods are separated by small margins. This is not an artifact of the performance metric used. As the additional performance measures in Appendix~\ref{supp:metrics} show, the foundation models lead across the board. For \gls{pd}, they top the ranking not only in terms of \gls{auc}, but also for other discrimination measures, error measures, and calibration measures. Hence, their advantage extends from discrimination to overall probabilistic accuracy, although the Brier score does not reveal whether mean \gls{pd} is systematically under- or overpredicted. The same holds for \gls{lgd}, where foundation models record the lowest prediction error (\gls{rmse} and \gls{mae}) and the highest rank correlation with realized losses (Spearman). 

\begin{figure}[ht!]
  \centering
  \includegraphics[width=0.8\textwidth,height=0.90\textheight,keepaspectratio]{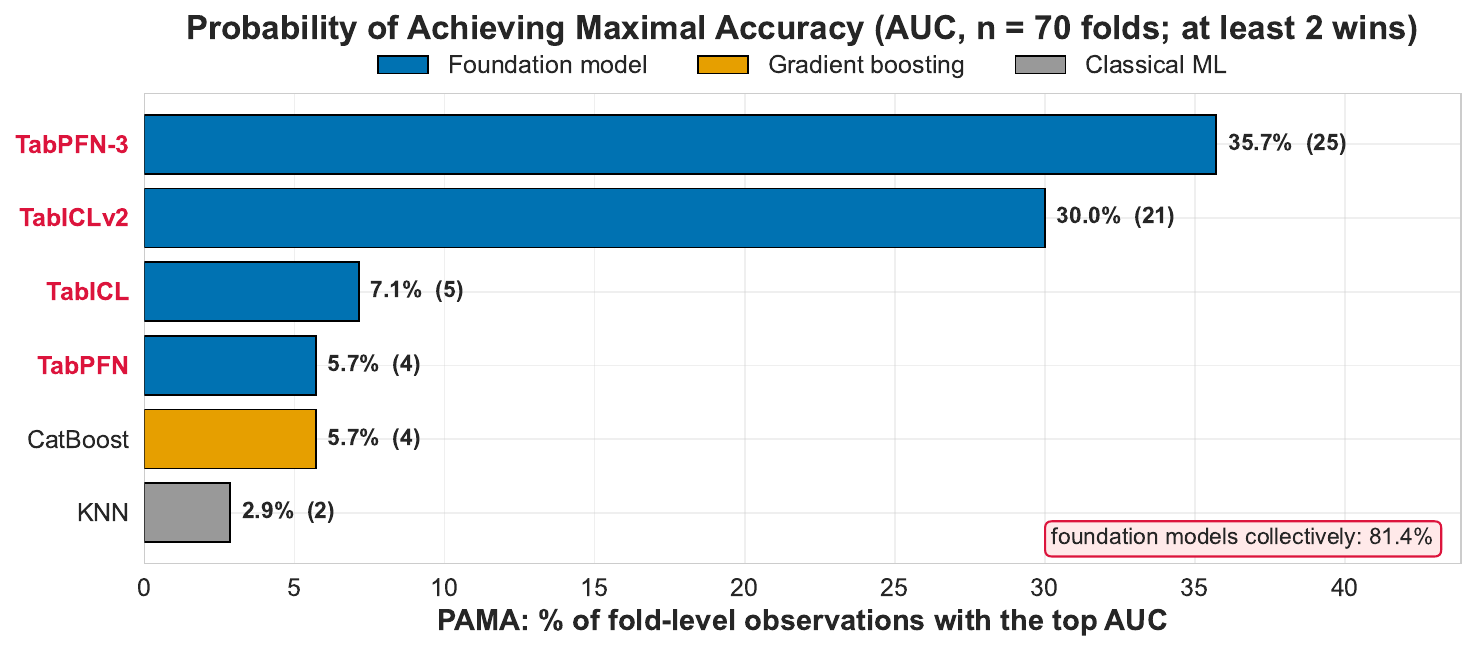}
\caption{Only methods with at least two winning fold-level observations are shown. Foundation models are highlighted in red. Comparison across all benchmarked learners.}
  \label{fig:pd_pama}
\end{figure}

\begin{figure}[ht!]
  \centering
  \includegraphics[width=0.8\textwidth,height=0.90\textheight,keepaspectratio]{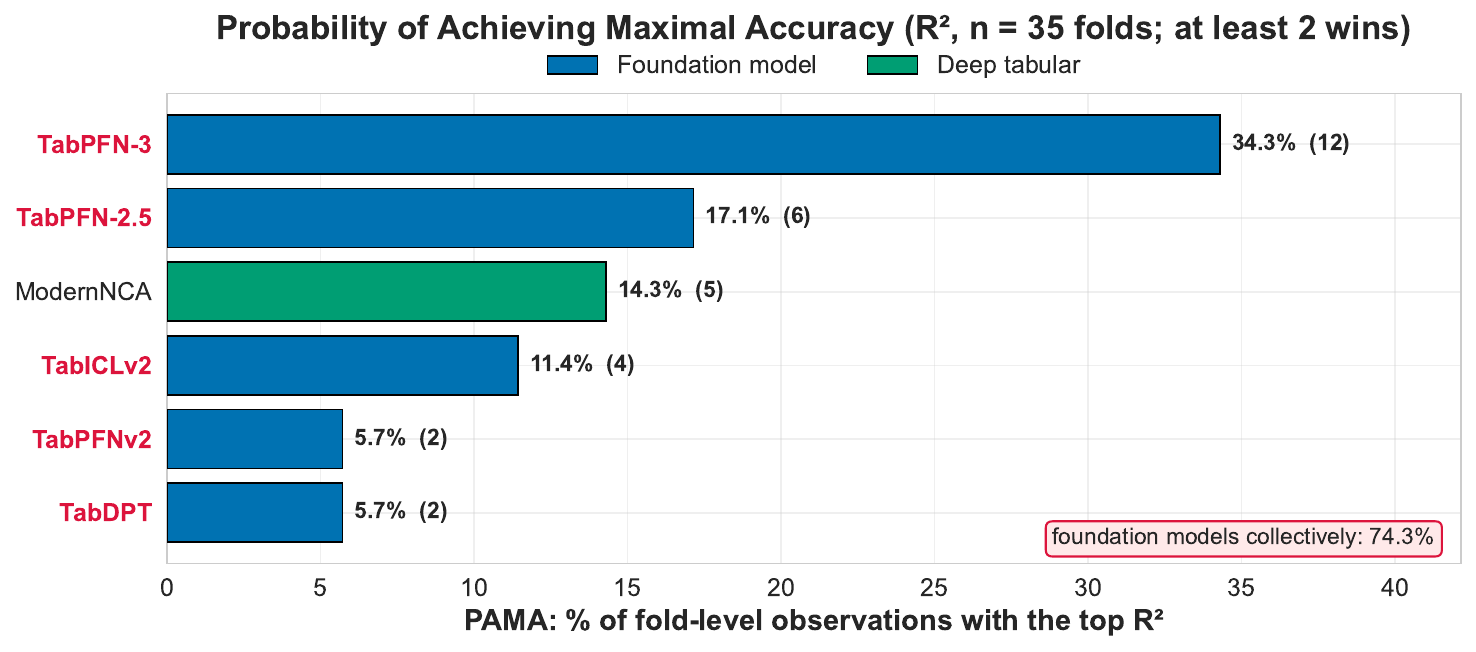}
\caption{Only methods with at least two winning fold-level observations are shown. Foundation models are highlighted in red. Comparison across all benchmarked learners.}
  \label{fig:lgd_pama}
\end{figure}

In the following, we further summarize the likelihood that a given method \textit{wins} a dataset--fold observation using the \gls{pama}, computed as the relative frequency with which a learner achieves the single best performance (in a given metric) across all cross-validation splits and datasets \citep{2014.FERNANDEZ-DELGADO.weneedhundredsclassifierssolverealworldclassificationproblems}. We use \gls{auc} for \gls{pd} and $R^2$ for the \gls{lgd} part. \gls{pama} provides an intuitive, rank-based summary of competitiveness that is robust to outlying datasets and makes no assumption about the distribution of performance differences. Figures~\ref{fig:pd_pama} and~\ref{fig:lgd_pama} report the \gls{pama} results for the \gls{pd} and \gls{lgd} benchmarks, respectively. For \gls{pd}, two foundation models separate themselves from the field. TabPFN-3 attains the highest \gls{auc} on $35.7\%$ of fold-level observations ($25$ of $70$) and TabICLv2 on $30.0\%$ ($21$ of $70$); the best conventional learner, CatBoost, leads on only $5.7\%$ ($4$ of $70$), tied with the original TabPFN. Aggregating over the segment, foundation models achieve the best score on $81.4\%$ of all folds ($57$ of $70$).

For \gls{lgd}, \gls{pama} paints a similar picture, although the wins are concentrated among five of the six evaluated foundation models. TabPFN-3 again leads, attaining the best $R^2$ on $34.3\%$ of folds ($12$ of $35$), followed by TabPFN-2.5 at $17.1\%$ ($6$ of $35$), TabICLv2 at $11.4\%$ ($4$ of $35$), and TabPFNv2 and TabDPT at $5.7\%$ ($2$ of $35$) each. The strongest conventional learner is ModernNCA, which attains the best $R^2$ on $14.3\%$ of folds ($5$ of $35$); no other conventional learner exceeds a single fold win. Foundation models collectively achieve the best score on $74.3\%$ of all folds ($26$ of $35$). Appendix~\ref{supp:ranks} reports the complete average-rank results and the corresponding per-dataset rank matrices for both tasks.

Notably, the results from \gls{pama} analysis also contextualize previous findings concerning average learner performance in that they show how the large error bars in Figure~\ref{fig:pd_bar_mean_auc} and Figure~\ref{fig:lgd_bar_mean_r2} stem mainly from performance variations across datasets rather than performance differences across learners, emphasizing how selected \glspl{tfm} perform slightly but consistently better than other, established alternatives.
Taken together, the results observed thus far provide strong evidence that recent \glspl{tfm} are a powerful prediction tool for credit risk management. 

\Needspace{6\baselineskip}
\subsubsection{Bias in mean predictions}
\label{subsubsec:bias_in_mean_predictions}
\begin{figure}[ht!]
    \centering
    \includegraphics[width=1.0\textwidth,height=0.90\textheight,keepaspectratio]{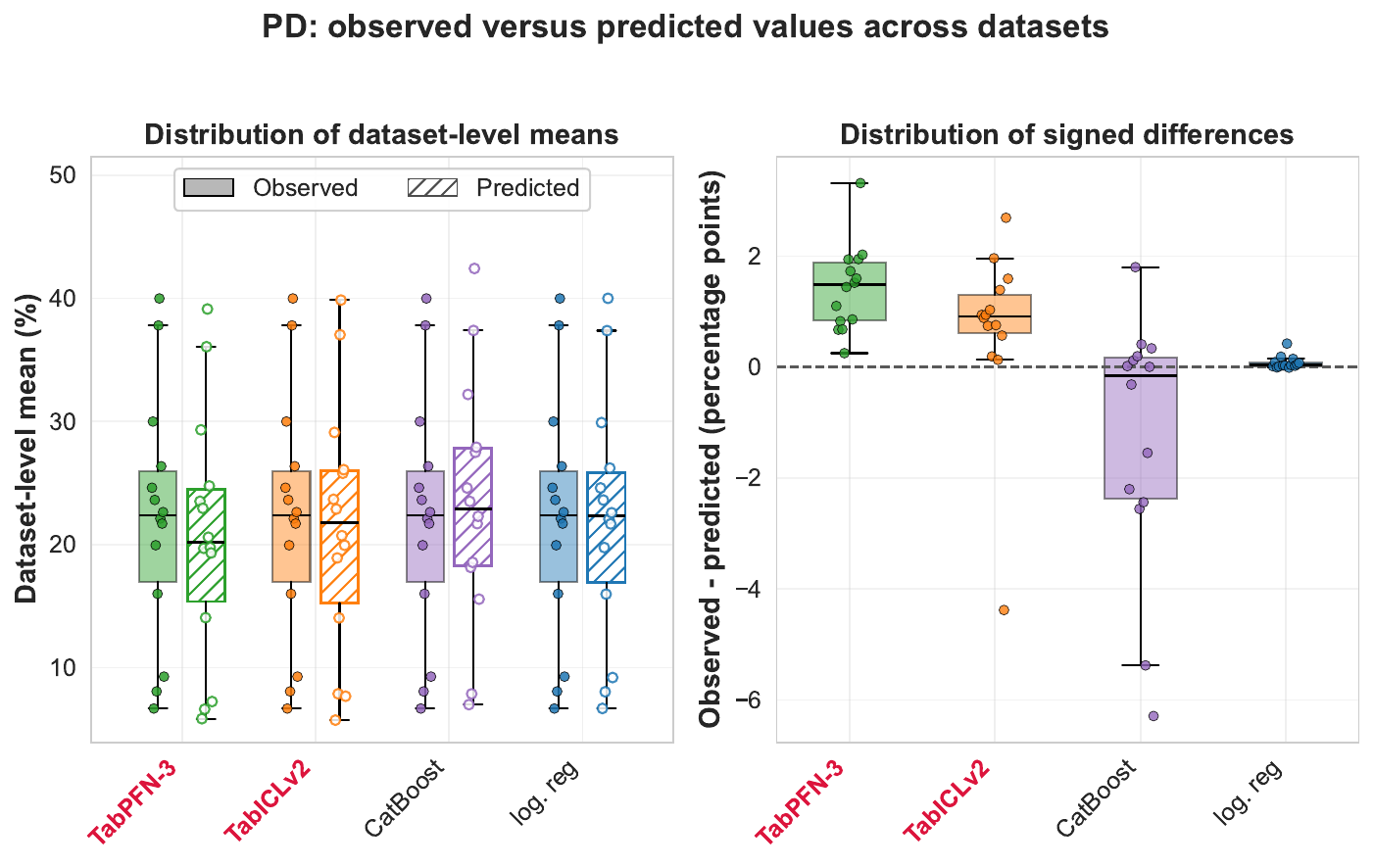}
    \caption{Calibration across the 14 \gls{pd} datasets. Left: dataset-level observed default rates and mean predicted default probabilities, one point per dataset. Right: the distribution of their signed differences in percentage points, with the dashed line marking zero bias. Positive observed-minus-predicted values indicate underprediction, negative values overprediction.}
    \label{fig:experiment1-pd-pd-selected-calibration-summary}
\end{figure}

\begin{figure}[ht!]
    \centering
    \includegraphics[width=1.0\textwidth,height=0.90\textheight,keepaspectratio]{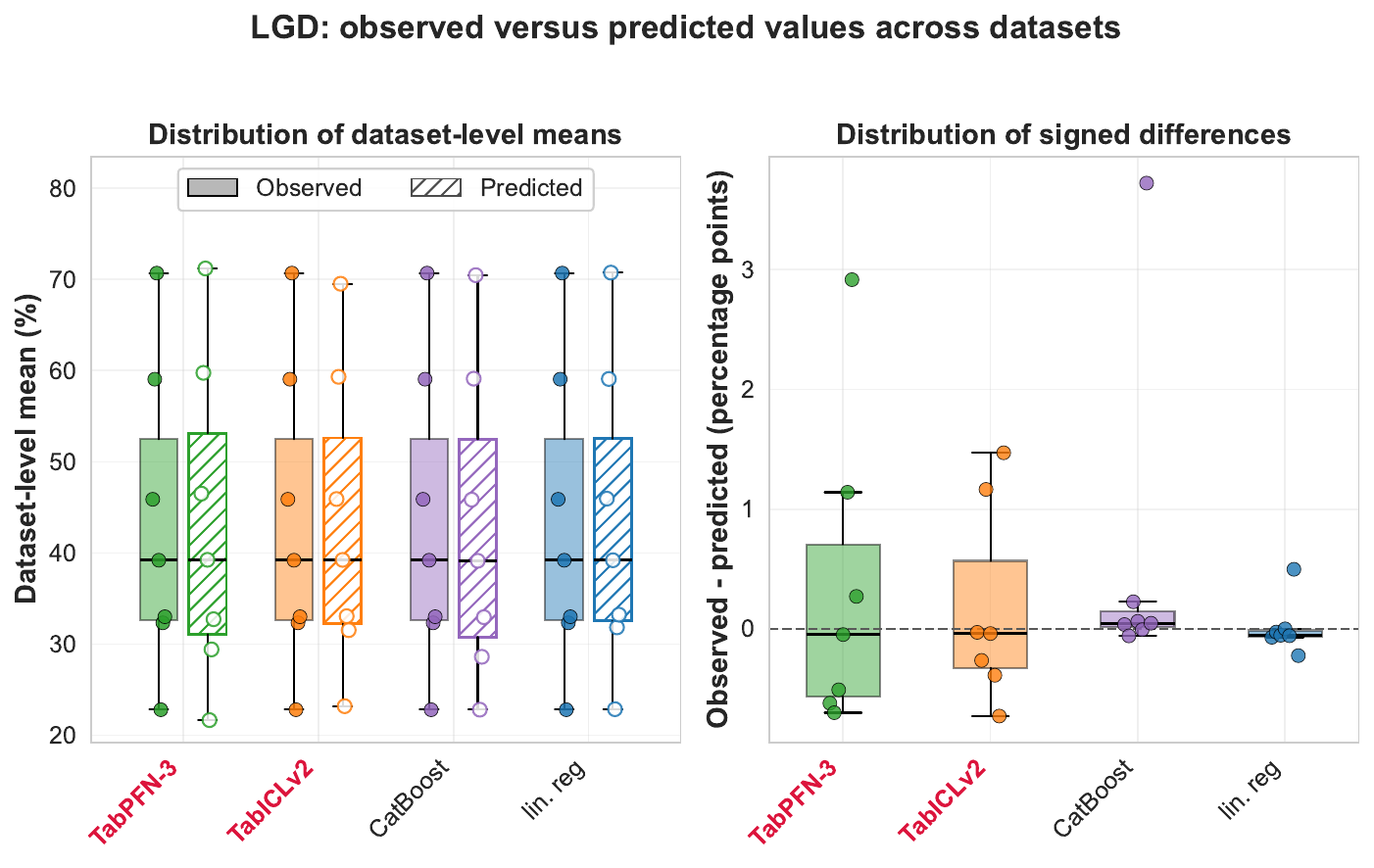}
    \caption{Calibration across the 7 \gls{lgd} datasets. Left: dataset-level observed and predicted mean \gls{lgd}, one point per dataset. Right: the distribution of their signed differences in percentage points, with the dashed line marking zero bias. Positive observed-minus-predicted values indicate underprediction, negative values overprediction.}
    \label{fig:experiment1-lgd-lgd-selected-calibration-summary}
\end{figure}

The headline benchmark evaluates predictive performance primarily through \gls{auc} for \gls{pd} and \(R^2\) for \gls{lgd}, while Appendix~\ref{supp:metrics} reports a broader set of discrimination, scoring, and prediction-error measures. These results characterize how accurately the methods rank observations and predict individual outcomes, but they do not isolate whether predictions are systematically too low or too high at the portfolio level. Figures~\ref{fig:experiment1-pd-pd-selected-calibration-summary} and~\ref{fig:experiment1-lgd-lgd-selected-calibration-summary} complement these results by examining the signed bias in dataset-level mean predictions for TabPFN-3, TabICLv2, CatBoost, and logistic or linear regression. Bias is defined as the observed minus the predicted mean, such that positive values indicate underprediction and negative values indicate overprediction. The left panels compare the observed and predicted means for each dataset, while the right panels show the corresponding distributions of signed bias. The more detailed prediction distributions underlying these summaries are reported in Appendix~\ref{sec:supp-prediction-densities}.

For \gls{pd}, logistic regression is nearly unbiased at the aggregate level, with a mean signed bias of \(+0.08\) percentage points. By contrast, both foundation models exhibit a systematic tendency to underpredict mean default rates. The mean signed bias is \(+1.42\) percentage points for TabPFN-3 and \(+0.68\) percentage points for TabICLv2. TabPFN-3 underpredicts the mean default rate in all 14 datasets, while TabICLv2 does so in 13 of 14; the sole exception is Vehicle Loan, for which TabICLv2 overpredicts by \(4.38\) percentage points. CatBoost shows the opposite aggregate tendency, with a mean signed bias of \(-1.27\) percentage points, although the direction of its bias varies across datasets. Thus, the strong predictive performance of the foundation models is accompanied by a clear tendency to understate the dataset-level default rate, particularly for TabPFN-3.

Figure~\ref{fig:experiment1-pd-pd-selected-calibration-summary} suggests that decision-makers must select specific use cases for foundation models carefully. A tendency to underestimate portfolio-level default rates is less relevant for an approval scorecard but a serious concern for an internal rating model used for risk management. Taking this latter perspective, one might prefer the more prudent CatBoost model although it is more biased. On the other hand, standard risk management practices involve a multi-layered architecture in which prediction model facilitate an ordering obligations based on their estimated default-risk, and estimates of portfolio-level default rates emerge from a dedicated calibration step \citep{2017..baseliiifinalisingpostcrisisreforms}. More generally, potential distortion in probabilistic forecasts of a learning algorithm can be addressed through post-hoc recalibration, using established methods, such as, Platt-scaling or isotonic regression \citep{bequeApproachesCreditScorecard2017}.

For \gls{lgd}, the aggregate directional biases are considerably smaller. The mean signed bias is \(+0.35\) percentage points for TabPFN-3, \(+0.17\) for TabICLv2, \(+0.58\) for CatBoost, and \(+0.01\) for linear regression, indicating little systematic under- or overprediction on average. These aggregate values nevertheless conceal dataset-specific biases. The most pronounced underprediction occurs for PropLGD4, where the signed bias reaches \(+3.72\) percentage points for CatBoost and \(+2.92\) percentage points for TabPFN-3. Hence, unlike for \gls{pd}, the \gls{lgd} results do not reveal a strong general tendency toward either underprediction or overprediction, although localized biases remain for particular portfolios.

\Needspace{8\baselineskip}
\subsection{Auxiliary experiments}
\label{subsec:auxiliary_experiments_results}

Having established that foundation models are competitive across the board, we next examine \emph{when} their advantage is largest. The two auxiliary experiments described in Section~\ref{subsec:auxiliary_experiments} address this question by varying the data regime along two axes that are considered to be common challenges in credit risk modeling: the sample size (\gls{rq} $2$) and, for \gls{pd}, the severity of class imbalance (\gls{rq} $3$). 

\Needspace{6\baselineskip}
\subsubsection{Data efficiency and the small-data regime}
\label{subsec:data_efficiency_and_the_small-data_regime}
\begin{figure}[ht!]
 \centering
 \includegraphics[width=0.8\textwidth,height=0.90\textheight,keepaspectratio]{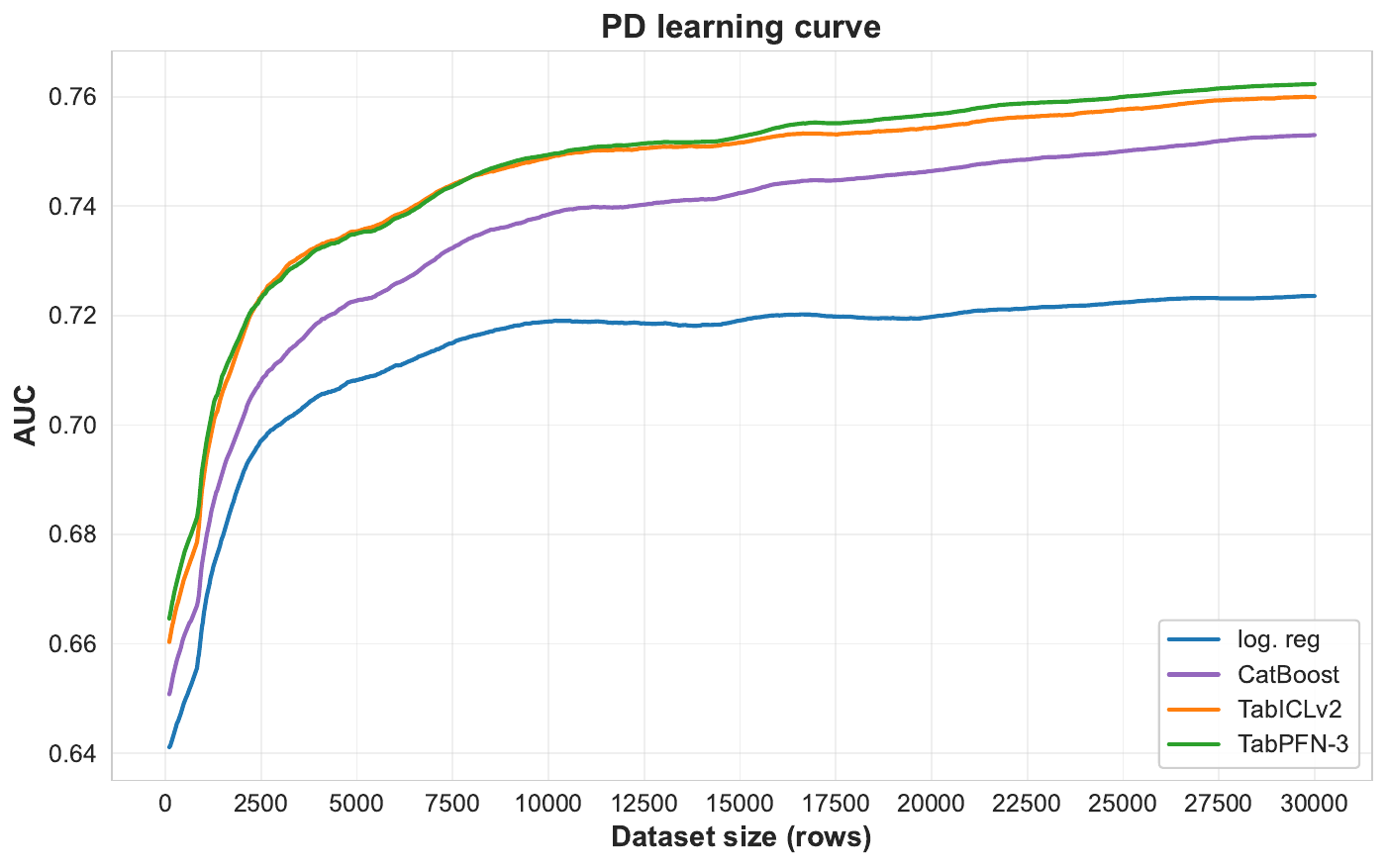}
\caption{Curves average scores across datasets at each sweep value. Moving-average lines use the shorter window.}
 \label{fig:pdlearningcurves}
\end{figure}

\begin{figure}[ht!]
 \centering
 \includegraphics[width=0.8\textwidth,height=0.90\textheight,keepaspectratio]{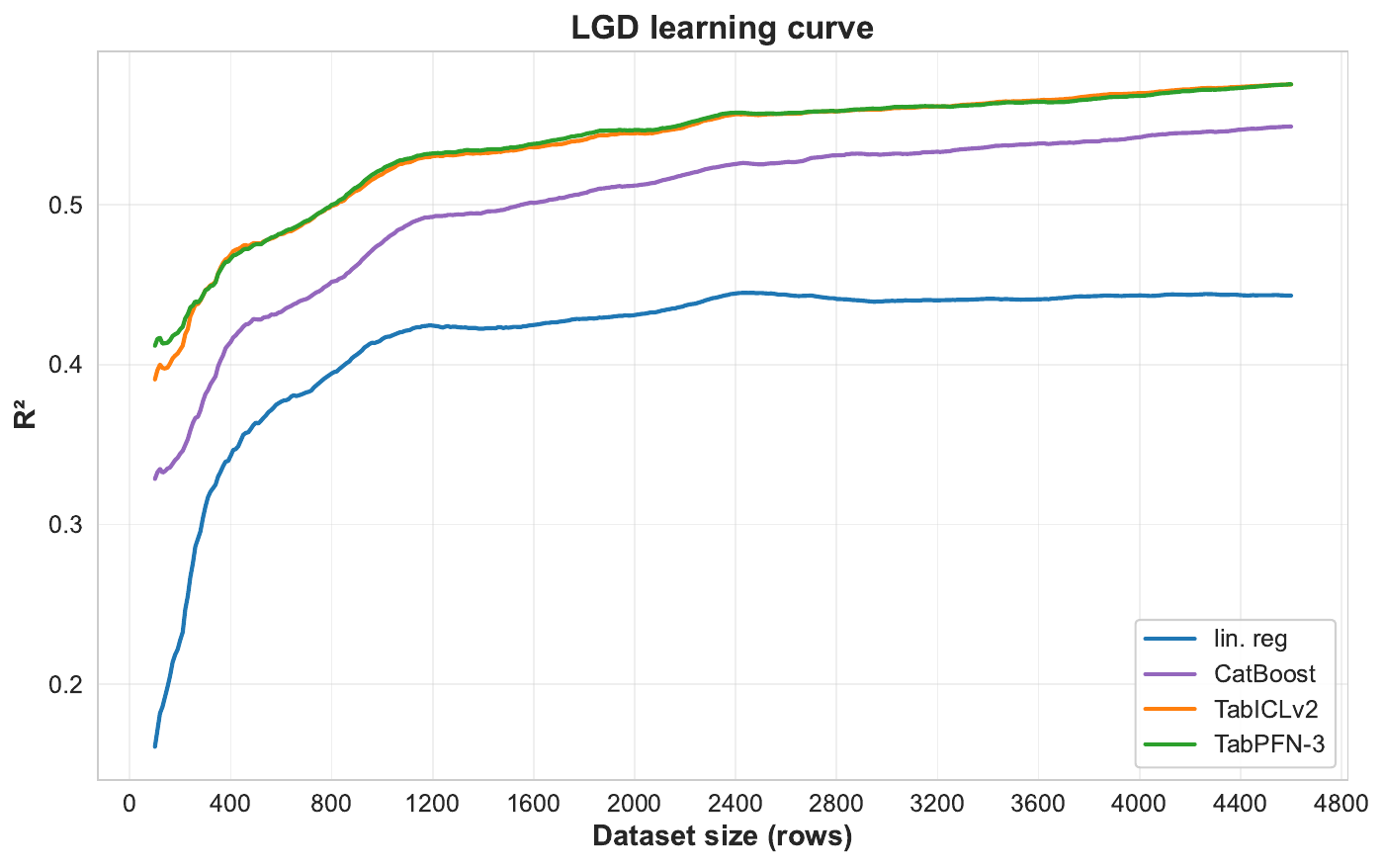}
\caption{Curves average scores across datasets at each sweep value. Moving-average lines use the shorter window.}
 \label{fig:lgdlearningcurves}
\end{figure}

Figures~\ref{fig:pdlearningcurves} and~\ref{fig:lgdlearningcurves} report the learning curves for \gls{pd} and \gls{lgd}, tracing mean \gls{auc} and $R^2$ as the capped dataset size (before splitting) grows from $100$ observations to the largest common size of the available datasets ($30{,}000$ for \gls{pd} over $9$ datasets, $4{,}600$ for \gls{lgd} over $4$ datasets). The curves show an overall upward trend for all methods, but the most relevant pattern is comparative: the foundation models are strongest in the low-to-moderate data regime, where the performance gap to the conventional baselines is largest, and this gap narrows as more data become available.

For \gls{pd}, the smallest sample sizes are noisy, but once a minimal context is available, the foundation models pull ahead. At $750$ observations, TabPFN-3 and TabICLv2 reach \gls{auc} values of $0.700$ and $0.693$, compared with $0.677$ for CatBoost and $0.673$ for logistic regression. By $30{,}000$ observations, they still lead, with \gls{auc} values of $0.763$ and $0.760$, but the gap to CatBoost ($0.754$) has narrowed. The pattern is even clearer for \gls{lgd}. At only $130$ observations, TabPFN-3 and TabICLv2 already achieve $R^2$ values of $0.449$ and $0.407$, corresponding to $78\%$ and $71\%$ of their own best performance over the sweep, while CatBoost reaches $0.325$ ($59\%$ of its own best) and linear regression $0.169$ ($38\%$). At the largest sample size, the foundation models still lead, but the gap is smaller: TabPFN-3 and TabICLv2 reach $R^2$ values of $0.576$ and $0.577$, compared with $0.550$ for CatBoost. This indicates that the foundation models achieve a large share of their eventual performance at much smaller sample sizes, with both models following similar trajectories and therefore appearing highly data-efficient.

The reported curves in Figures~\ref{fig:pdlearningcurves} and~\ref{fig:lgdlearningcurves} provide direct evidence in support of the following answer to \gls{rq}~$2$: the pretraining of foundation models translates into markedly higher sample efficiency, and their advantage is greatest in the low-to-moderate data regime; this indicates a strong potential for adoption in small portfolios, e.g., for project lending or newly launched products where conventional learners might struggle more due to data scarcity. The same ordering appears at full dataset size, where the relative gain of TabPFN-3 over CatBoost declines with the number of rows on both tasks. This is the complementary view to the subsampling design, which by construction evaluates portfolios that are natively large; Appendix~\ref{supp:viz} reports the association and its limitations.

\Needspace{6\baselineskip}
\subsubsection{Robustness to class imbalance}
\label{subsec:robustness_to_class_imbalance}
\begin{figure}[ht!]
 \centering
 \includegraphics[width=0.8\textwidth,height=0.90\textheight,keepaspectratio]{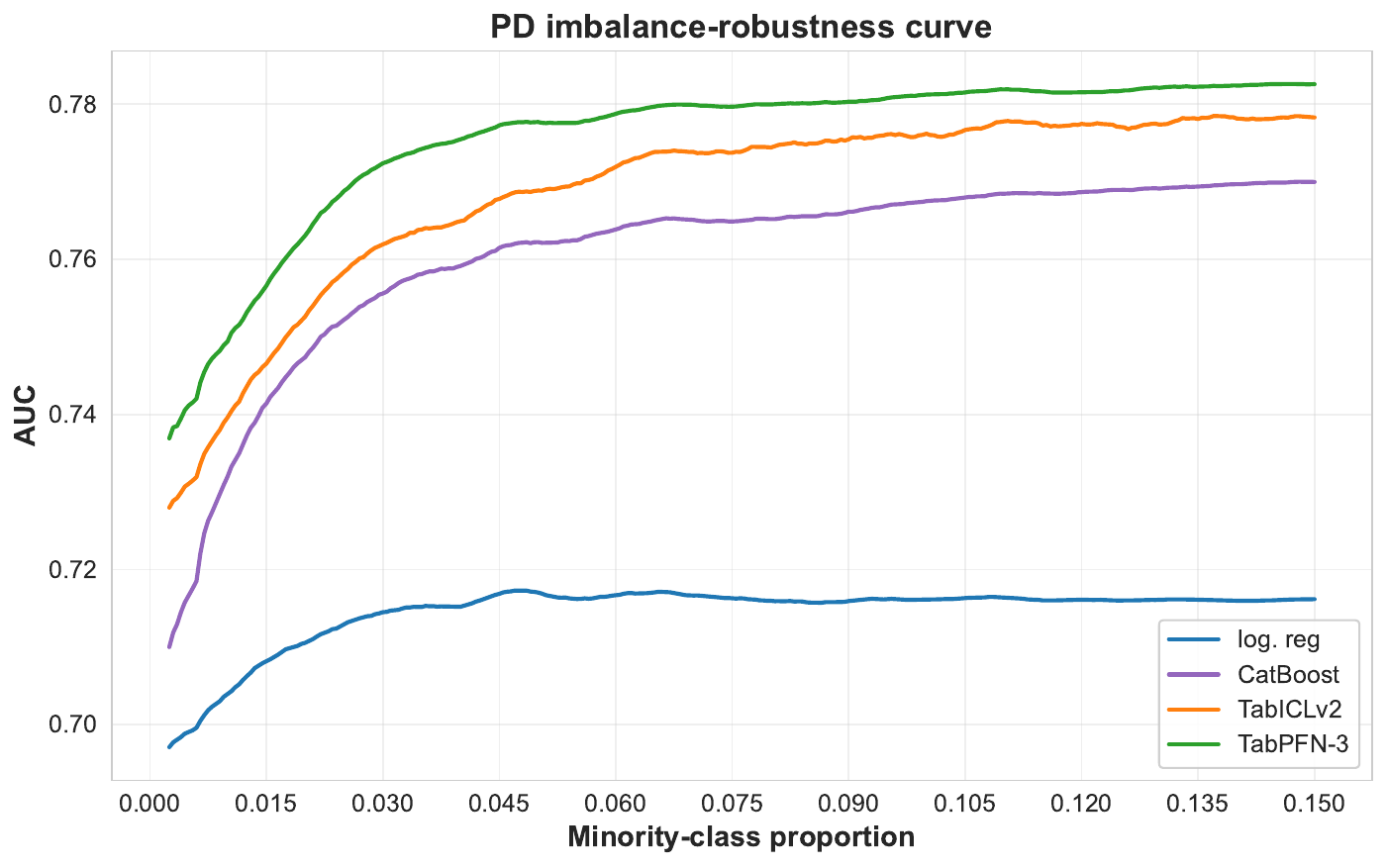}
\caption{Curves average scores across datasets at each sweep value. Moving-average lines use a shorter window.}
 \label{fig:pd_imbalance_auc}
\end{figure}

The second auxiliary experiment focuses on \gls{pd} modeling, particularly low-default portfolios, by incrementally reducing the default rate from $0.15$ to $0.0025$. Figure~\ref{fig:pd_imbalance_auc} reports mean \gls{auc} across the six included datasets as a function of the minority-class proportion. The central finding is that the foundation models retain the highest absolute performance throughout the sweep, and their advantage over CatBoost becomes larger as defaults become rarer. At a $15\%$ default rate, TabPFN-3 and TabICLv2 reach \gls{auc} values of $0.783$ and $0.778$, respectively, compared with $0.770$ for CatBoost and $0.716$ for logistic regression. At the most extreme $0.25\%$ default rate, TabPFN-3 and TabICLv2 remain at $0.718$ and $0.715$, whereas CatBoost falls to $0.683$ and logistic regression reaches $0.689$. Thus, the gap between TabPFN-3 and CatBoost increases from roughly $0.012$ \gls{auc} points at the least imbalanced setting to roughly $0.035$ at the most imbalanced setting. CatBoost is, in fact, the most imbalance-sensitive method in the panel; for the most extreme setting, its performance drops below logistic regression ($0.689$), which is otherwise the weakest method\footnote{This crossing is based on the unsmoothed sweep values; it is not visually apparent in Figure~\ref{fig:pd_imbalance_auc}, which shows moving-average trends.}. Among the high-performing methods, the foundation models are affected least, with TabPFN-3 retaining about $92\%$ of its least-imbalanced \gls{auc} at the $0.25\%$ setting, compared with $88.7\%$ for CatBoost. This is exactly the behavior one would hope for in low-default modeling: as defaults become rarer, the boosting baseline, which must partition on an ever smaller set of positive examples, deteriorates fastest, while the in-context foundation models remain comparatively stable.

A second observation concerns the performance of the two foundation models. They are close on \gls{auc} throughout, with TabPFN-3 consistently stronger than TabICLv2, especially in the severely imbalanced band of roughly $1\%$ to $5\%$ positives (a delta in \gls{auc} of about $0.010$, against $0.004$ at the least imbalanced setting and a near tie at the $0.25\%$ extreme). Taken together, the class imbalance experiment motivates the following answer to \gls{rq}~$3$: foundation models maintain the highest absolute \gls{auc} under severe class imbalance, with TabPFN-3 strongest throughout. This property is of particular relevance for modeling low-default portfolios.

\clearpage
\section{Management implications and research directions}
\label{sec:implications_and_rd}
The empirical results demonstrate the potential of \glspl{tfm} for credit risk prediction, particularly in data-constrained settings. However, their functioning and use differ fundamentally from conventional credit risk models. This section complements the quantitative assessment with a qualitative appraisal of the implications for credit risk modeling and management, and maps out selected directions for research.

\Needspace{8\baselineskip}
\subsection{Implications for credit risk management}
\label{subsec:managerial_implications}
This subsection discusses managerial implications of the rapidly evolving \gls{tfm}-paradigm and the observed benchmarking results. We consider how \glspl{tfm} may affect the economics of model development, the sources of competitive advantage among \glspl{fi}, and the organization and governance of the broader credit risk ecosystem.

\Needspace{6\baselineskip}
\subsubsection{Model development and competitive capabilities}
\label{subsec:from_predictive_performance_to_competitive_capabilities}
A first implication concerns the role of predictive modeling capabilities as a source of competitive advantage. Historically, advances in credit risk analytics have often been associated with the adoption of increasingly sophisticated modeling techniques. \glspl{fi} that were able to deploy superior predictive models could potentially benefit from improved risk assessment, more accurate capital allocation, and more efficient lending decisions. FinTechs exemplify this strategy. Exploiting reduced regulatory requirements as non-bank \glspl{fi}, many FinTechs leverage advanced \gls{ai} tools to gain market share and competitive advantages in the lending business. The results reported here suggest that \glspl{tfm} represent yet another step in the evolution of prediction technology. In particular, the strong performance of recent \gls{tfm} generations suggests that this class of models is still evolving rapidly and may not yet have reached the performance plateau observed for many mature \gls{ml} approaches (e.g., \gls{gb}).

At the same time, the long-term strategic implications of this development may differ from the immediate performance gains observed in empirical benchmarks. Foundation models are designed to be broadly applicable and reusable across tasks. If \glspl{tfm} continue to improve and become widely available, advanced predictive capabilities may become accessible to a larger number of institutions, gradually commoditizing parts of model development. Competitive advantage would then derive less from the ability to develop a marginally more accurate predictive model and more from the complementary capabilities required to use that model effectively: identifying suitable applications, integrating predictions into lending processes, validating performance for the relevant portfolio, and maintaining appropriate governance and monitoring. For example, explainability and fairness remain essential requirements for responsible credit risk modeling and decisioning. The use of a \gls{tfm} does not remove or fundamentally alter these obligations, which remain with the deploying institution. Extending the perimeter of \glspl{tfm} to reflect corresponding requirements in their prediction also creates opportunity for research, which we discuss in Section~\ref{subsec:research-agenda}.

Furthermore, the emergence of \glspl{tfm} does not diminish the importance of institution-specific data. The predictive performance of a \gls{tfm} depends on the quality, relevance, and representativeness of the information supplied for \gls{icl}. Our results demonstrate that \glspl{tfm} can leverage limited data more effectively than many conventional approaches, but they do not imply that data quality ceases to matter or that performance transfers uniformly across portfolios. On the contrary, as predictive technologies become more widely accessible, the ability to generate, curate, and govern high-quality data may become an even more important source of differentiation among \glspl{fi}. Recent industrial evidence points in the same direction: PRAGMA's reported gains stem from pretraining on proprietary, multi-source banking transactions and event histories, illustrating that foundation-model capabilities may amplify rather than replace the strategic value of rich institution-specific data \citep{2026.OSTROUKHOV.pragmarevolutfoundationmodel}.

The relevant economic comparison also extends beyond predictive accuracy and the effort required to fit an individual model. Depending on how a \gls{tfm} is provided, institutions may incur licensing fees, substantial inference costs, or expenses associated with adapting their data and technical infrastructure. Reliance on an external model or application programming interface may additionally create vendor dependence, switching costs, and uncertainty about future access, pricing, or model updates. Conversely, reusable foundation models may reduce repeated expenditures on method selection, \gls{hpo}, and redevelopment across portfolios. Assessing their managerial value therefore requires a lifecycle perspective that compares acquisition, integration, validation, operation, monitoring, and replacement costs with the benefits of improved predictions and shorter development cycles.

These considerations suggest that \glspl{tfm} may change the composition rather than eliminate the importance of modeling capabilities. Whether their diffusion primarily benefits institutions with lean and highly automated processes, established lenders combining rich proprietary data with mature governance structures, or specialized technology providers remains an open question. 

\Needspace{6\baselineskip}
\subsubsection{Governance and the credit risk ecosystem}
\label{subsec:implications_for_the_credit_risk_ecosystem}

Beyond the level of individual institutions, the emergence of \glspl{tfm} may have broader implications for the organization and governance of the credit risk ecosystem. Credit risk modeling is characterized by a highly decentralized structure in which \glspl{fi} develop, validate, maintain, and periodically redevelop proprietary models, often to address highly similar prediction problems. Foundation models introduce the possibility of a different organizational logic, where substantial effort is concentrated on the development, pretraining, and validation of a small number of general-purpose predictive systems that can subsequently be deployed across a wide range of downstream applications.

Such a development could have important consequences for regulatory supervision and model governance. Current supervisory frameworks are largely built around institution-specific models whose properties must be validated separately by each institution and reviewed by regulators. The emergence of widely adopted foundation models raises the possibility that some evidence concerning their general properties, including robustness, stress behavior, and known failure modes, could be generated centrally and reused across applications. Common documentation and validation resources could reduce duplication and promote more consistent assessments. They could not, however, replace institution-specific validation, because realized performance also depends on the target portfolio, data construction, outcome definitions, implementation choices, and intended use. Moreover, externally developed models may change through new weights, software versions, or service updates, requiring governance processes that identify material changes and determine when renewed validation is necessary.

A related implication concerns the long-standing regulatory challenge of achieving comparability of risk assessments across \glspl{fi}. Differences in internally developed models have repeatedly been identified as a source of variation in risk estimates and risk-weighted assets, leading to several revisions of the Basel framework since its inception. To the extent that multiple institutions rely on common foundation-model infrastructures, part of this variability could be reduced. This raises the prospect of a future ecosystem in which standardization and comparability are improved without necessarily sacrificing predictive performance or risk sensitivity. At the same time, common modeling infrastructures cannot ensure comparable risk estimates when institutions differ in their data, definitions, calibration procedures, or use of model outputs.

The emergence of \glspl{tfm} may also influence the roles of other actors in the credit risk ecosystem. Credit bureaus and specialized service providers, which often hold large volumes of data spanning institutions or countries, have traditionally created value through data provision, score development, and analytical services. In a foundation-model paradigm, these actors may increasingly participate in the development, maintenance, certification, and distribution of pretrained predictive systems. This shift could create new forms of specialization and facilitate access to advanced modeling capabilities, particularly for smaller institutions. It could also concentrate technical knowledge, data, and market power among a limited number of providers, thereby increasing common dependencies and the potential systemic relevance of shared model weaknesses.

Taken together, these considerations suggest that the significance of \glspl{tfm} may ultimately extend beyond their immediate predictive performance. Our results establish that they are a competitive modeling approach for contemporary credit risk prediction, but their managerial value will depend on the full set of capabilities and costs associated with their deployment. Their longer-term importance may lie in accelerating the diffusion of advanced predictive technologies while shifting the principal sources of differentiation toward proprietary data, institutional implementation, validation, governance, and integration. At the ecosystem level, the same development may enable greater standardization and reuse while creating new dependencies that institutions, service providers, and supervisors must manage.

\Needspace{8\baselineskip}
\subsection{Research directions}
\label{subsec:research-agenda}

This invited review evaluates whether \glspl{tfm} transfer successfully to credit risk prediction. Accordingly, we do not attempt to formulate a comprehensive research agenda for credit risk management. Instead, we derive three connected directions from the paper's conceptual arguments and empirical findings. First, research should establish the boundary conditions of the observed performance gains and explain why pretraining is especially beneficial when observations or defaults are scarce. Second, it should determine how \glspl{tfm} can be calibrated, validated, and governed as pretrained systems whose predictions depend jointly on an external prior and institution-specific context data. Third, it should examine whether the paradigm extends beyond cross-sectional \gls{pd} and \gls{lgd} prediction to the dynamic, normative, and decision-dependent problems encountered in lending. Progress in all three directions requires evaluation infrastructure that makes the provenance of data and pretrained models transparent.

\Needspace{6\baselineskip}
\subsubsection{Why, when, and when not do tabular foundation models work?}
\label{subsubsec:why-when-tfms-work}
Our benchmark establishes that recent \glspl{tfm} can predict \gls{lgd} and \gls{pd} accurately, but do not explain the underlying mechanisms. Understanding them is scientifically important and practically necessary: without an account of what knowledge is transferred and when it is useful, an institution has limited grounds for deciding whether benchmark results are externally valid and informative about a new portfolio.

The Bayesian interpretation of \glspl{pfn} provides one possible explanation. Pretraining exposes the model to tasks drawn from a prior distribution and teaches it to approximate posterior predictive inference when a new labeled dataset is supplied as context; performance should therefore depend on whether the prior assigns sufficient probability to structures resembling the target task. A complementary frequentist interpretation views a pretrained \gls{pfn} as a predictor whose tuning parameters have been selected across many tasks before it encounters the target dataset \citep{2023.NAGLER.statisticalfoundationspriordatafittednetworks}. Its small-sample performance may then reflect a favorable bias--variance trade-off induced by cross-task pre-tuning. More generally, observed gains could arise from the prior, architectural inductive biases, regularization through pretraining, context construction, internal preprocessing, retrieval, or ensembling.

Comparisons between complete systems cannot separate these explanations because current \glspl{tfm} differ along most of these dimensions simultaneously. Mitra emphasizes mixtures of synthetic priors, Real-TabPFN introduces continued pretraining on real tables, and TabDPT combines real-data pretraining with retrieval-based inference \citep{2025.ZHANG.mitramixedsyntheticpriorsenhancingtabularfoundationmodels,2025.MA.tabdptscalingtabularfoundationmodelsrealdata,2025.GARG.realtabpfnimprovingtabularfoundationmodelscontinuedpretrainingrealworlddata}. Full factorial experiments would require repeated pretraining and may be prohibitively expensive. Therefore, we invite research on selective ablations with computationally cheaper designs: controlled target tasks with known data-generating mechanisms, smaller proxy models, intermediate checkpoints, and interventions on the composition, size, and labels of the context set. Such studies could quantify how much a prediction depends on pretrained regularities rather than local evidence and test whether the mechanisms responsible for sample efficiency also explain the models' robustness to rare defaults and their particularly large gains for \gls{lgd}.

The related question of \emph{when} transfer works should be framed as compatibility between the pretraining distribution and the target portfolio. Sample size, dimensionality, and class balance provide useful initial descriptors, but compatibility may also concern nonlinearities and interactions, target shape, missingness and measurement processes, selection mechanisms, temporal stability, and the definitions of default and loss. Meta-analyses across heterogeneous portfolios should examine which observable task characteristics predict the incremental performance of a \gls{tfm} over a strong task-specific learner. The objective is not merely to identify settings in which average performance is high, but to establish empirically testable boundary conditions for the value of pretraining.

Further, the research agenda must consider negative transfer. An external prior is valuable when it supplies relevant information that the target data cannot estimate reliably, but harmful when it favors relationships that are inappropriate for a particular portfolio. Potential sources of mismatch in credit risk include institution-specific underwriting and collection policies, selective observation of accepted applicants, changes in borrower populations, economic regime shifts, and differences in outcome definitions. Research should develop diagnostics that detect when a target task lies outside the effective support of the pretraining distribution and distinguish prior mismatch from insufficient context or ordinary estimation error. Such diagnostics should guide an operational response: revising or localizing the context set, recalibrating or adapting the model, combining it with a locally trained learner, or reverting to task-specific estimation.

Context design is central to this problem because the context set is the application-specific component through which a \gls{pfn} learns about a portfolio. Research should examine whether contexts should prioritize recent or locally similar observations, preserve portfolio-level target frequencies, or be constructed separately for borrower segments and individual queries. It should also quantify sensitivity to individual context observations, row order, label noise, and changes in context composition. These questions are especially relevant in low-default portfolios, where a small number of observations may materially influence both discrimination and probability levels.

Finally, research should determine the appropriate scope of pretraining. General-purpose \glspl{tfm} seek regularities that transfer across arbitrary tables, whereas credit-specific models could encode recurring structures across institutions, products, risk parameters, and economic conditions. Greater domain specificity may improve relevance, but it may also reproduce historical biases, institutional practices, or definitions that do not transfer. Comparisons between general and credit-specific priors should establish which aspects of credit-risk structure are sufficiently stable to share. Synthetic generation offers one route for encoding hypothesized structures, while federated or other privacy-preserving approaches may permit learning from confidential institutional data without pooling raw observations \citep{nareshPrivacypreservingFederatedCredit2026}. The research target is what information should be transferred and at what level of specificity.

\Needspace{6\baselineskip}
\subsubsection{Calibration, validation, and evaluation infrastructure}
\label{subsubsec:validation-infrastructure}

Strong ranking performance does not by itself make a \gls{tfm} suitable for credit-risk use. In our benchmark, TabPFN-3 systematically underpredicts average \gls{pd}, despite leading on discrimination. Future studies should determine whether such bias arises from the pretraining distribution, context composition, class-prior handling, ensembling, or other elements of the inference pipeline. They should also assess whether established recalibration methods remain effective with few defaults and stable across time, borrower segments, and economic conditions. Related work should investigate predictive uncertainty, including uncertainty arising from limited context, sensitivity to context selection, and incompatibility between the pretrained prior and the target portfolio.

The foundation-model paradigm also changes the object of validation. Conventional validation focuses mainly on a model estimated from institutional data. For a \gls{pfn}, evidence must distinguish among the fixed pretrained system, its institution-specific context and preprocessing, and the resulting application. System-level validation should document general capabilities, limitations, pretraining provenance, and known failure modes. Application-level validation should establish that the target portfolio and context are appropriate, that discrimination and calibration satisfy the requirements of the intended use, and that predictions remain stable under plausible changes in data and context. Lifecycle validation must additionally account for changes to pretrained weights, software, preprocessing, retrieval, and surrounding ensembles, any of which may alter predictions without conventional redevelopment.

Interpretability should be considered within this validation framework. The relevant question is not only whether feature attributions can be computed, but whether they are stable when the context changes, whether they primarily reflect local portfolio evidence or relationships inherited from pretraining, and whether identified relationships can be challenged by validators and domain experts. Similar distinctions apply to robustness and fairness: a general property reported for the pretrained system does not establish that it holds for a particular institution, population, or decision process.

These questions require evaluation infrastructure tailored to pretrained tabular models. Data provenance, transformations, sampling periods, outcome definitions, and possible pretraining exposure must be documented so that transfer can be distinguished from memorization or leakage. Model artifacts, checkpoints, inference configurations, and context-construction procedures should be versioned. Benchmarks should include temporal and cross-institutional holdouts, controlled changes in target prevalence and context composition, and portfolios selected to vary systematically in their distance from the pretraining distribution. Secure benchmark environments may permit fixed pretrained models to be evaluated on confidential institutional data without disclosing the underlying portfolios.

Existing data initiatives illustrate both the opportunity and the limitations. For example, the \emph{Global Credit Data} pools anonymized internal data from member banks and provides long histories for \gls{pd}, \gls{lgd}, and \gls{ead} modeling under common definitions and quality controls \footnote{\url{https://globalcreditdata.org}}. Similarly, the European DataWarehouse\footnote{\url{https://eurodw.eu}} collects and validates standardized loan-level information across European securitized and private portfolios. Such infrastructures could support analysis across products, institutions, jurisdictions, and economic conditions that conventional public benchmarks cannot provide. Access restrictions, confidentiality, and differences between recorded and required variables nevertheless imply that open datasets, controlled-access repositories, institutional collaborations, and confidential benchmark environments will remain complementary. Their common objective should be a reproducible evaluation of both the pretrained system and its portfolio-specific use.

\Needspace{6\baselineskip}
\subsubsection{Extending the scope of foundation-model credit risk analytics}
\label{subsubsec:specialized-foundation-models}

Once the mechanisms, boundary conditions, and validation requirements of existing \glspl{tfm} are better understood, a further question is whether the paradigm can address credit-risk problems beyond static \gls{pd} classification and \gls{lgd} regression. Three extensions concern distinct dimensions of the modeling problem: survival and longitudinal models change the structure of the predicted outcome; fairness requirements concern properties that predictions and decisions should satisfy; and causal models change the inferential question from forecasting outcomes to estimating the effects of interventions. Recent prototypes show that each dimension can, in principle, be represented through pretraining, but none yet constitutes an established credit-risk solution.

A first direction concerns survival and longitudinal modeling. Default, recovery, prepayment, and completion of the workout process are naturally represented as time-to-event or multistate processes \citep{2016.BAESENS.creditriskanalyticsmeasurementtechniquesapplicationsexamplessas,medinaolivares2025joint}. SurvPFN demonstrates that a \gls{pfn} can be pretrained on synthetic survival tasks, incorporate censoring, and produce continuous-time predictive distributions \citep{bohmSurvPFNFoundationModels2026}. Credit applications, however, involve competing events, transitions among account states, and evolving inputs such as repayments, transactions, arrears, and macroeconomic conditions. Research should test whether pretraining improves the estimation of rare transitions and newly observed portfolios, whether it preserves calibration across economic regimes, and whether a common model can combine static borrower characteristics with irregular longitudinal histories more effectively than separately developed task-specific models.

A second direction concerns fairness. Lending decisions affect access to finance and may disadvantage protected groups, while the meaning of fairness depends on the product, jurisdiction, affected population, and decision context. FairPFN illustrates that a model can be pretrained on synthetic causal structures to attenuate the influence of a protected attribute without requiring the target dataset's causal graph as an input \citep{robertson2024fairpfn}. This demonstrates methodological feasibility, but does not resolve the normative choice among incompatible fairness criteria or guarantee transfer under causal misspecification. Research should examine multiple and intersecting protected attributes, alternative sources of direct and indirect discrimination, sensitivity to the assumed causal structures, and stability under population change. Corresponding work can clarify whether pretrained models can implement institutionally specified constraints transparently and consistently.

A third direction moves from predicting risk to estimating the consequences of credit decisions. Lenders intervene through pricing, credit limits, contract terms, restructuring offers, and collection strategies \citep{fahnerEstimatingCausalEffects2012}. Selecting among these actions requires counterfactual evidence rather than predictions under the historically observed lending process. CausalPFN-style models suggest that treatment-effect estimation can be amortized across simulated data-generating processes, in a manner related to the use of synthetic causal structures in FairPFN \citep{2025.BALAZADEHMERESHT.causalpfnamortizedcausaleffectestimationincontextlearning}. Yet pretraining cannot make causal identification automatic. Approval policies determine which outcomes are observed, relevant confounders may be unavailable, and interventions can be continuous, dynamic, and interdependent \citep{kozodoi2025fighting,BOCKELRICKERMANN2025114378,alfonsosanchez2024optimizing}. Research must therefore establish which identification assumptions are encoded in the prior, how they can be communicated and challenged, and whether the resulting models improve treatment-effect estimation and decision quality in realistic pricing, limit-management, restructuring, and collection settings.

These extensions should remain connected to the core questions developed above. Their value depends on prior--task compatibility, their outputs require application-specific calibration and validation, and their evaluation demands longitudinal, action, and protected-group data that are often unavailable in public benchmarks. The central research opportunity is therefore not merely to create credit-specific versions of SurvPFN, FairPFN, CausalPFN, or other instances of \gls{pfn}-type estimators. It is to determine whether foundation-style pretraining transfers reliably to the dynamic, normative, and decision-dependent conditions of real lending, and to identify when specialized or conventional approaches remain preferable.

\clearpage
\section{Conclusions}
\label{sec:conclusions}

This invited review examines the emergence of \glspl{tfm} as an important development in tabular \gls{ml} and assesses their relevance to credit risk modeling. We review the conceptual foundations, architectural evolution, and pretraining strategies of \gls{pfn}-based models and complement this synthesis with a domain-specific empirical evaluation. The review explains how \glspl{tfm} shift part of learning from task-specific model fitting to reusable pretraining and in-context inference, while the empirical analysis tests whether the resulting capabilities transfer to \gls{pd} and \gls{lgd} prediction under conditions characteristic of credit portfolios.

Regarding predictive competitiveness (\gls{rq}~$1$), the results favor the most recent foundation models. TabPFN-3 ranks first for both tasks, attaining the highest average \gls{auc} for \gls{pd} classification and the highest average $R^2$ for \gls{lgd} regression. Foundation models as a group achieve the best score on $81.4\%$ of \gls{pd} folds and $74.3\%$ of \gls{lgd} folds. Their margin over the strongest \gls{gb} baseline is narrow for \gls{pd} but substantial for \gls{lgd}. However, this result is driven by the latest TabPFN and TabICL generations rather than by the model class as a whole. Moreover, strong discrimination does not ensure unbiased probability levels: TabPFN-3 underpredicts the mean default rate across all $14$ \gls{pd} datasets, whereas the corresponding \gls{lgd} biases are small. Consequently, its use for applications requiring probability levels, including provisioning or capital calculations, requires an appropriate calibration step.

The auxiliary experiments provide initial evidence on how the observed gains vary with common credit risk modeling characteristics. Regarding sample efficiency (\gls{rq}~$2$), TabPFN-3 and TabICLv2 lead in the low-to-moderate data regime for both tasks, with their advantage over CatBoost generally narrowing as more observations become available. Regarding class imbalance (\gls{rq}~$3$), the foundation models retain the highest absolute \gls{auc} as defaults become increasingly rare, while the gap between TabPFN-3 and CatBoost widens. Taken together, these findings indicate that pretraining is especially valuable when the target portfolio contains limited information, whether because few observations are available or because the event of interest is rare. This makes recent \glspl{tfm} particularly relevant to low-default portfolios, specialized segments, and newly launched products.

The reviewed literature and the empirical evidence suggest that \glspl{tfm} represent more than another family of tabular prediction algorithms. By combining regularities acquired across pretraining tasks with evidence supplied through an institution-specific context, they introduce a distinct modeling paradigm whose performance depends on both sources of information. Their strong results demonstrate the potential value of this transfer, but variation across model generations and the observed \gls{pd} bias also reveal current limits. Understanding compatibility between the pretrained system and the target portfolio, including the possibility of negative transfer, is therefore central to explaining when these models should be preferred to conventional task-specific methods.

The broader significance of these findings is that \glspl{tfm} change where learning takes place and, consequently, how credit risk modeling must be organized and studied. Their adoption may reduce the importance of repeatedly developing models from scratch, while increasing the importance of understanding how pretrained knowledge interacts with institution-specific data and requirements. This perspective connects the managerial and research agendas: both depend on distinguishing the general capabilities of a pretrained system from its suitability for a particular portfolio.
Recent \glspl{tfm} should therefore be regarded as serious candidates for contemporary credit risk prediction, particularly in information-constrained settings, while their wider adoption depends on a better understanding of the mechanisms, limits, and reliability of transfer.

\medskip
\noindent\textbf{Data and code availability.} This study uses publicly available and proprietary datasets. Public source locations are provided in Table~\ref{tab:data-summary}. Code for model evaluation, statistical analysis, and figure generation is available at \url{https://github.com/andreasgoethals/TabPFNCredit}.

\medskip
\noindent\textbf{Funding.} The computational resources and services used in this work were provided by the \gls{vsc}, funded by the Flemish Government.

\medskip
\noindent\textbf{Declarations of interest.} None.

\medskip
\noindent\textbf{Acknowledgments.} The benchmark was implemented and run using \gls{talent}, a unified benchmarking framework for tabular learning \citep{2025.LIU.talenttabularanalyticslearningtoolbox}.

\medskip
\noindent\textbf{Declaration of generative \gls{ai} and \gls{ai}-assisted technologies in the manuscript preparation process.} During the preparation of this work the authors used Claude in order to develop code and to streamline and improve writing of the manuscript. After using this tool, the authors reviewed and edited the content as needed and take full responsibility for the content of the published article.

\clearpage
\bibliographystyle{plainnat}
\bibliography{references}

\clearpage
\appendix
\counterwithin{figure}{section}
\counterwithin{table}{section}
\numberwithin{equation}{section}
\section*{Appendices}

These appendices provide the analyses and documentation that support the paper but are not reported in full in the main text. Appendix~\ref{supp:stats} presents the statistical analysis, and Appendix~\ref{supp:metrics} reports performance measures beyond the two primary measures, \gls{auc} for \gls{pd} classification and $R^2$ for \gls{lgd} regression. Appendix~\ref{supp:viz} provides additional visualizations and supporting results, Appendix~\ref{sec:supp-prediction-densities} shows the per-dataset prediction distributions underlying the mean-bias analysis, and Appendix~\ref{supp:ranks} reports the complete rank-based summary. Finally, Appendix~\ref{supp:data} documents the target distributions, eligibility for the auxiliary experiments, and overlap between the evaluation datasets and the documented pretraining corpora of models pretrained on real data.

\section{Statistical analysis}
\label{supp:stats}

\Needspace{8\baselineskip}
\subsection{Statistical methodology}
We compare the methods across datasets by combining an established significance-testing protocol with a Bayesian analysis and a descriptive summary, and we apply the same procedure to \gls{pd} classification and \gls{lgd} regression. The backbone is the protocol of \citet{2006.DEMSAR.statisticalcomparisonsclassifiersmultipledatasets}: a Friedman omnibus test of whether the methods differ at all, followed, only if it rejects, by pairwise rank comparisons with a correction for multiple testing, reported as a \gls{cd} diagram and a matrix of adjusted $p$-values. Classical significance tests of this kind report only whether a difference is unlikely to have arisen by chance, not how large or how probable it is, and they lose power when the datasets are few, so we complement them with the Bayesian signed-rank test, which returns the probability that one method is better or that two methods are practically equivalent, and with a descriptive Win/Loss matrix. Finally, because the full method panel is large relative to the number of datasets, we also report an exploratory comparison of a compact, data-selected panel of family champions.

The omnibus test asks whether the methods differ at all. We use the Friedman test \citep{1937.FRIEDMAN.useranksavoidassumptionnormalityimplicitanalysisvariance}, which ranks the methods on each dataset and tests whether the average ranks depart from equality, in the less conservative $F$-distributed form derived by \citet{1980.IMAN.approximationscriticalregionfbietkanstatistica}. The average ranks on which these tests operate are reported in full in Appendix~\ref{supp:ranks}.

When the omnibus test rejects, we examine pairwise differences using two complementary post-hoc procedures applied to the same average ranks. The \gls{cd} diagram, introduced for this purpose by \citet{2006.DEMSAR.statisticalcomparisonsclassifiersmultipledatasets} on the basis of the all-pairwise rank test of \citet{1963.NEMENYI.distributionfreemultiplecomparisons}, places the methods on the rank axis and connects methods that are not separated by the Nemenyi test. Alongside it, we report a matrix of pairwise $p$-values adjusted with the step-down Shaffer procedure of \citet{2008.GARCIA.extensionstatisticalcomparisonsclassifiersmultipledatasetsallpairwisecomparisons}. This procedure can be more powerful than applying one common critical distance to every pair \citep{2010.GARCIA.advancednonparametrictestsmultiplecomparisonsdesignexperimentscomputationalintelligencedataminingexperimentalanalysispower} and is the all-pairwise rank counterpart of the Holm correction \citep{1979.HOLM.simplesequentiallyrejectivemultipletestprocedure}.

The Bayesian signed-rank test, introduced by \citet{2014.BENAVOLI.bayesianwilcoxonsignedranktestbaseddirichletprocess} and applied here following the tutorial of \citet{2017.BENAVOLI.timechangetutorialcomparingmultipleclassifiersbayesiananalysis}, is the Bayesian counterpart of the Wilcoxon signed-rank test \citep{1945.WILCOXON.individualcomparisonsrankingmethods}. For a pair of methods it returns the posterior probability that one is better, that the other is better, or that the two are practically equivalent, the last defined by a \gls{rope} of $\pm 0.01$ in the metric (\gls{auc} or $R^2$). Unlike a significance test it can therefore support, and not only reject, the hypothesis that a difference is too small to matter in practice.

We then report the Win/Loss matrix, which is descriptive rather than inferential: each cell counts the datasets on which one method beats another, shaded by the net margin, and so conveys the direction and consistency of head-to-head dominance that a single significance verdict discards.

The benchmark contains many methods (33 for \gls{pd} classification, 27 for \gls{lgd} regression) but comparatively few datasets (14 and 7). On a panel of that size the all-pairwise post-hoc has little power and the \gls{cd} grows so large that the diagram cannot separate the leading methods from one another. We therefore repeat the Nemenyi and Shaffer comparisons on a reduced panel of champions: the single best-performing method of each modeling family on the primary metric, together with the linear model that remains the standard reference in credit scoring practice. For \gls{pd} classification the champions are TabPFN-3 (the best foundation model), CatBoost (the best \gls{gb} ensemble), T2G-Former (the best deep tabular network), and logistic regression; for \gls{lgd} regression they are TabPFN-3, CatBoost, RealMLP (the best deep tabular network), and linear regression. This reduced panel provides a focused comparison with the strongest representative of each established family. Because the champions are selected using the same benchmark on which they are compared, however, this analysis is exploratory and its adjusted $p$-values do not account for model-selection uncertainty. The linear model is retained for its role as the industry benchmark rather than for its performance.

\Needspace{8\baselineskip}
\subsection{PD classification}

The Friedman test strongly rejects equal performance on the full panel of 33 methods ($F_F = 10.32$, $p = 6.9 \times 10^{-36}$; $N = 14$, $k = 33$), providing evidence that performance differs across methods. Localizing the differences on this panel is, however, uninformative. The Nemenyi \gls{cd} spans roughly 13.9 rank positions, so the \gls{cd} diagram merges almost every method into a single indistinguishable group, and the Shaffer correction flags only 61 of the $\binom{33}{2} = 528$ pairs, essentially all of them separations of the leading methods from the weakest (Naive Bayes, TabNet, \gls{knn}, the \gls{ncm} classifier) rather than distinctions among the competitive methods at the top. We therefore reserve the \gls{cd} diagram and the adjusted $p$-value matrix for the exploratory champion comparison and characterize the full panel with the Bayesian test and the Win/Loss matrix.

\begin{figure}[ht!]
  \centering
  \includegraphics[width=\textwidth,height=0.90\textheight,keepaspectratio]{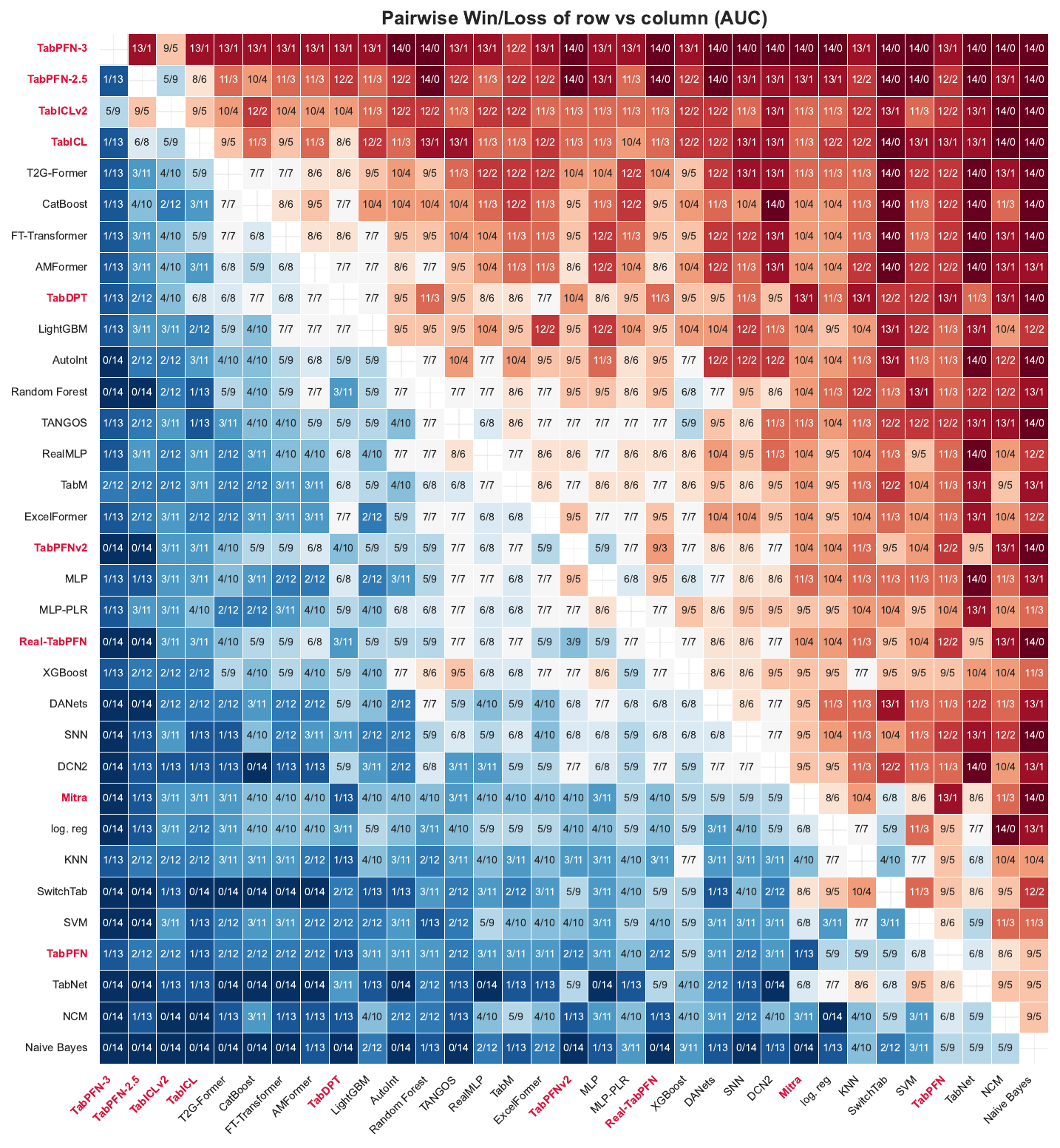}
  \caption{Pairwise Win/Loss matrix for the full \gls{pd} panel (33 methods, $N = 14$ datasets). Each cell 'W/L' is the number of datasets on which the row method beats versus loses to the column method, with ties omitted. Cells compare row methods against column methods; color encodes the win-loss margin (red: the row method dominates; blue: it is dominated). Methods are ordered by mean rank, so the strongest sit top-left. Foundation models are highlighted in red.}
  \label{fig:wlpd}
\end{figure}

The Bayesian signed-rank test (\gls{rope} $\pm 0.01$ \gls{auc}) compares TabPFN-3 against its closest rivals. Against TabICLv2, the next foundation model, it assigns probability 0.76 to practical equivalence and 0.24 to TabPFN-3 being better, with no posterior mass, at the reported precision, on TabICLv2 being better. Against CatBoost, the strongest \gls{gb} model, it assigns probability 0.57 to TabPFN-3 being better and 0.43 to practical equivalence, with no posterior mass, at the reported precision, on CatBoost being better. Against logistic regression, it assigns probability 0.87 to TabPFN-3 being better. These comparisons favor TabPFN-3 over the linear baseline, while its difference from CatBoost frequently lies within the \gls{rope}. The Win/Loss matrix (Figure~\ref{fig:wlpd}) shows a similar descriptive pattern: TabPFN-3 and TabICLv2 post positive net win rates against nearly every competitor.

\begin{figure}[ht!]
  \centering
  \begin{subfigure}{0.48\textwidth}
    \centering
    \includegraphics[width=\textwidth,height=0.90\textheight,keepaspectratio]{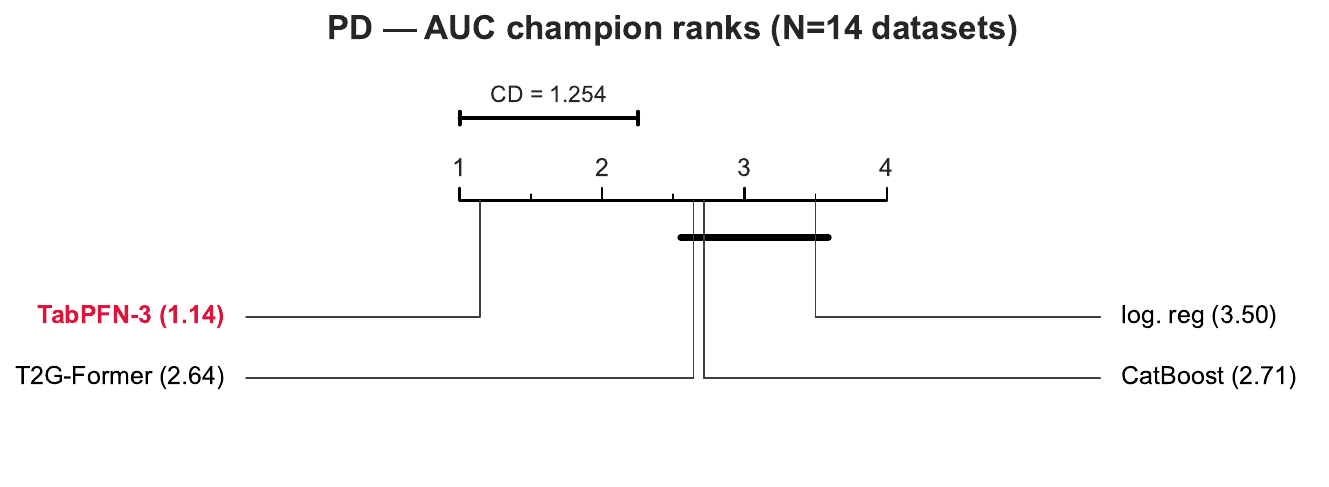}
    \caption{Lower ranks are better; horizontal bars connect methods not separated by the Nemenyi test. Foundation models are highlighted in red.}
    \label{fig:pd_cd}
  \end{subfigure}\hfill
  \begin{subfigure}{0.48\textwidth}
    \centering
    \includegraphics[width=\textwidth,height=0.90\textheight,keepaspectratio]{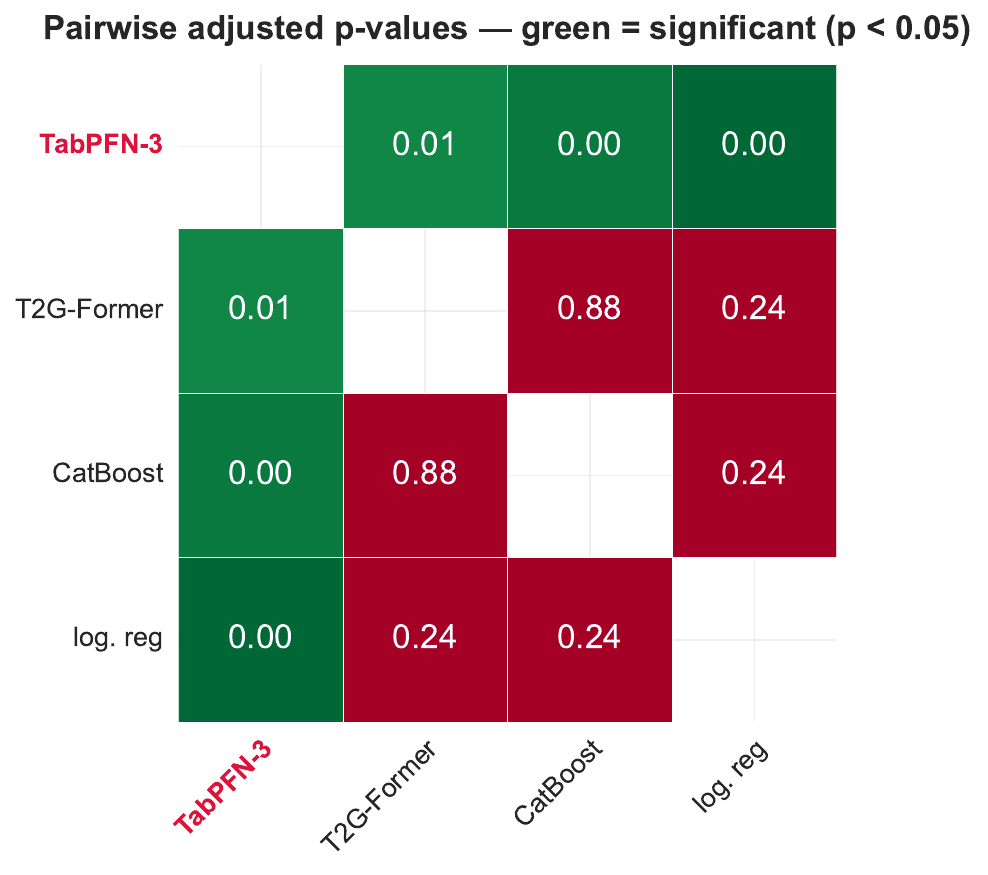}
    \caption{Cells contain adjusted \(p\)-values; green cells mark significant pairwise differences. Foundation models are highlighted in red.}
    \label{fig:pd_sig}
  \end{subfigure}
  \caption{Post-hoc champion-level \gls{pd} comparison of the four selected family representatives (TabPFN-3, CatBoost, T2G-Former, logistic regression) on \gls{auc}. (a) Nemenyi \gls{cd} diagram of the average ranks (1 = best); methods connected by a bar are not significantly separated. (b) Matrix of Shaffer-adjusted pairwise $p$-values; a cell is green where the two selected methods differ significantly and red where they do not. The foundation model TabPFN-3 is shown in red.}
  \label{fig:pd_champion}
\end{figure}

Within the exploratory four-champion comparison (TabPFN-3, CatBoost, T2G-Former, and logistic regression), the \gls{cd} diagram (Figure~\ref{fig:pd_cd}) places TabPFN-3 ahead of the other three, which are not separated from one another by the Nemenyi test. The Shaffer-adjusted matrix (Figure~\ref{fig:pd_sig}) reports adjusted \(p\)-values below the conventional threshold for the three comparisons involving TabPFN-3. These reduced-panel results do not account for the data-dependent selection of the family champions.

\clearpage

\Needspace{8\baselineskip}
\subsection{LGD regression}

The Friedman test rejects equal performance on the full panel of 27 methods ($F_F = 8.64$, $p = 3.3 \times 10^{-19}$; $N = 7$, $k = 27$). As for \gls{pd} classification, the full-panel post-hoc is uninformative: the Nemenyi \gls{cd} spans about 15.7 rank positions and the Shaffer correction flags only 24 of the $\binom{27}{2} = 351$ pairs, again strong-versus-weak contrasts (the leading foundation models against \gls{svm}, \gls{knn}, TabNet, and linear regression). With only seven datasets, the pairwise tests have limited power; a nonsignificant adjusted comparison should therefore be interpreted as inconclusive rather than as evidence of equality.

\begin{figure}[ht!]
  \centering
  \includegraphics[width=\textwidth,height=0.90\textheight,keepaspectratio]{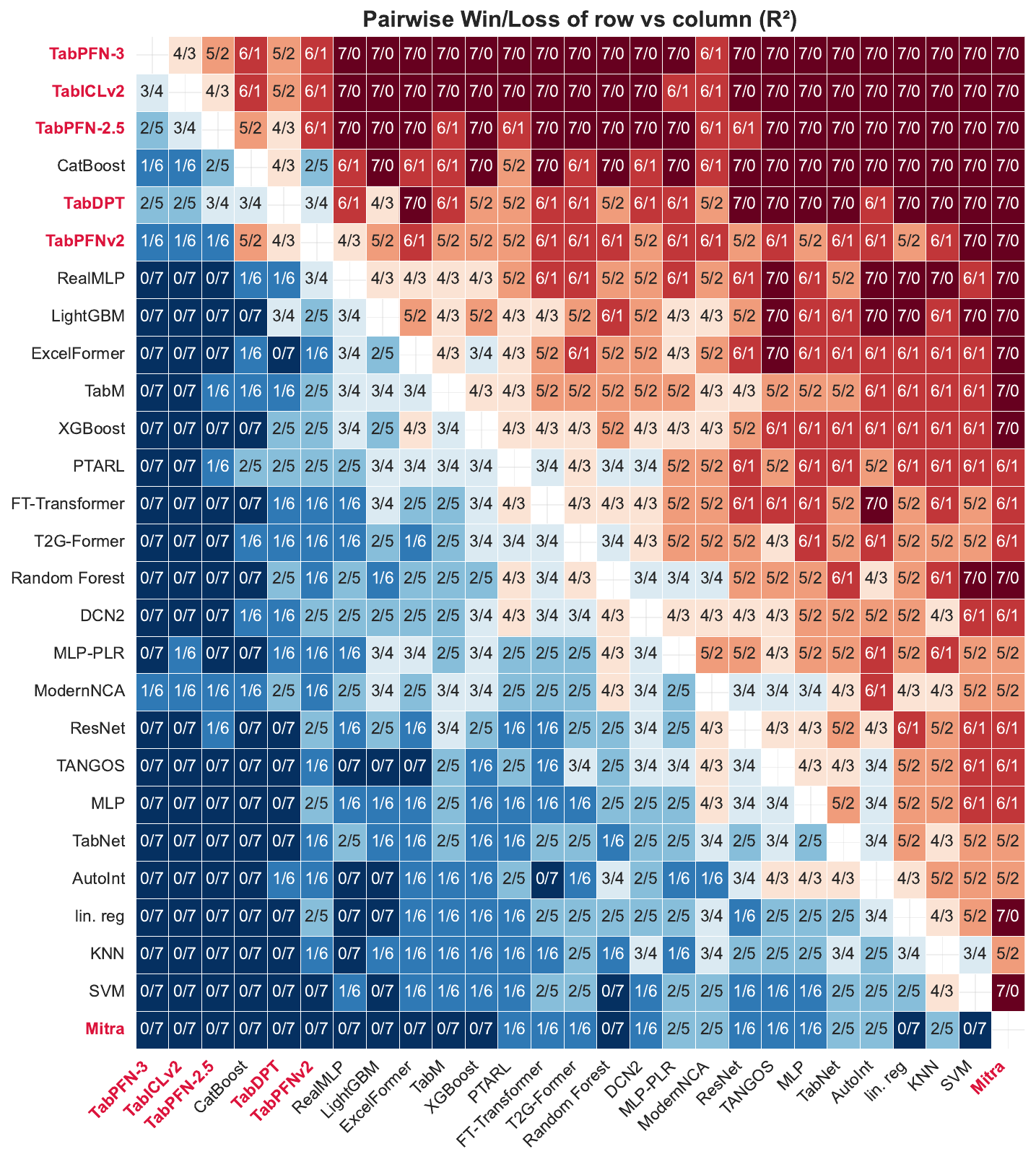}
  \caption{Pairwise Win/Loss matrix for the full \gls{lgd} panel (27 methods, $N = 7$ datasets). Each cell 'W/L' is the number of datasets on which the row method beats versus loses to the column method, with ties omitted. Cells compare row methods against column methods; color encodes the win-loss margin (red: the row method dominates; blue: it is dominated). Methods are ordered by mean rank, so the strongest sit top-left. Foundation models are highlighted in red.}
  \label{fig:wllgd}
\end{figure}

The Bayesian signed-rank test (\gls{rope} $\pm 0.01$ $R^2$) provides a complementary probabilistic comparison. For TabPFN-3 against TabICLv2, it assigns probability 0.59 to practical equivalence, 0.38 to TabPFN-3 being better, and 0.03 to TabICLv2 being better. Against CatBoost, it assigns probability 0.75 to TabPFN-3 being better, 0.20 to practical equivalence, and 0.05 to CatBoost being better; against linear regression it assigns probability 0.97 to TabPFN-3 being better, with no posterior mass, at the reported precision, on linear regression being better. Thus, where the adjusted pairwise test does not separate TabPFN-3 from CatBoost, the Bayesian analysis assigns a high posterior probability to TabPFN-3 being better. The Win/Loss matrix (Figure~\ref{fig:wllgd}) shows a similar descriptive pattern, with the leading foundation models posting strong head-to-head counts and Mitra the conspicuous exception at the bottom.

\begin{figure}[ht!]
  \centering
  \begin{subfigure}{0.48\textwidth}
    \centering
    \includegraphics[width=\textwidth,height=0.90\textheight,keepaspectratio]{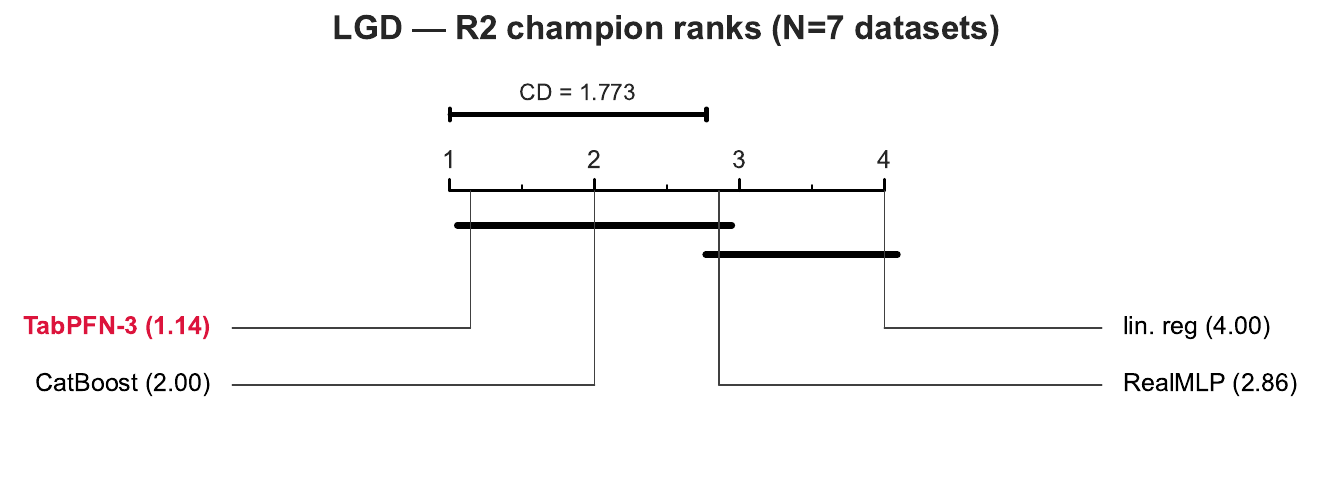}
    \caption{Lower ranks are better; horizontal bars connect methods not separated by the Nemenyi test. Foundation models are highlighted in red.}
    \label{fig:lgd_cd}
  \end{subfigure}\hfill
  \begin{subfigure}{0.48\textwidth}
    \centering
    \includegraphics[width=\textwidth,height=0.90\textheight,keepaspectratio]{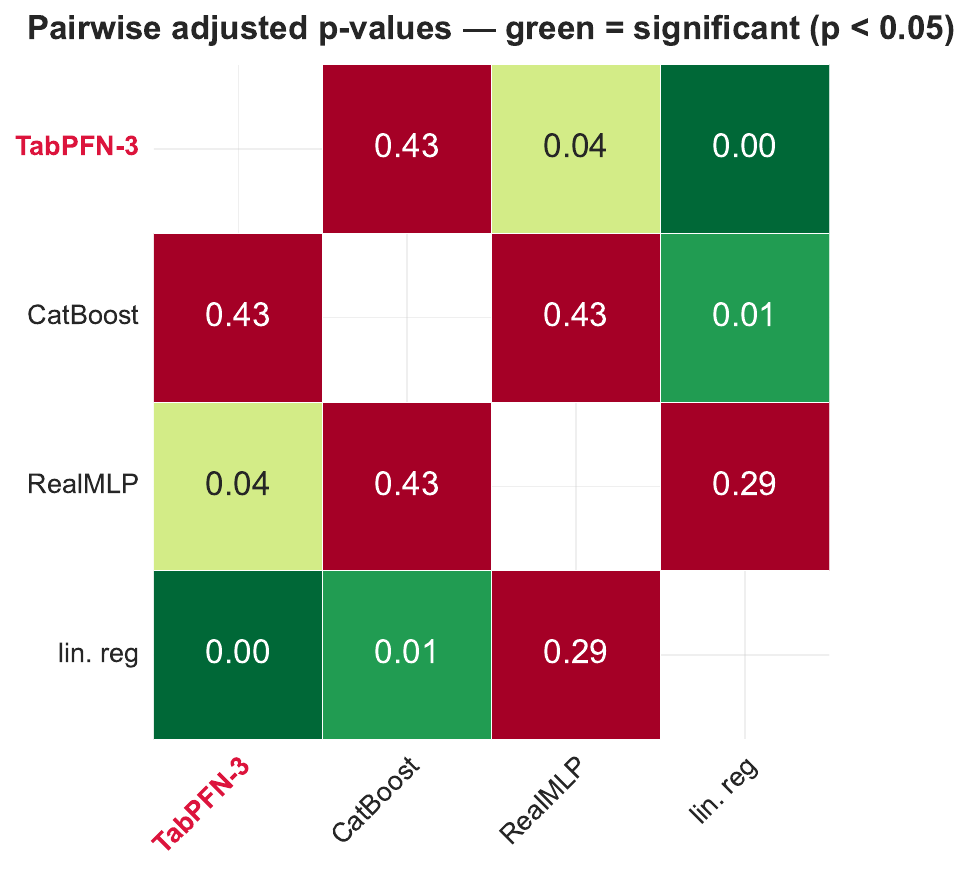}
    \caption{Cells contain adjusted \(p\)-values; green cells mark significant pairwise differences. Foundation models are highlighted in red.}
    \label{fig:lgd_sig}
  \end{subfigure}
  \caption{Post-hoc champion-level \gls{lgd} comparison of the four selected family representatives (TabPFN-3, CatBoost, RealMLP, linear regression) on $R^2$. (a) Nemenyi \gls{cd} diagram of the average ranks (1 = best); methods connected by a bar are not significantly separated. (b) Matrix of Shaffer-adjusted pairwise $p$-values; a cell is green where the two selected methods differ significantly and red where they do not. TabPFN-3 is shown in red.}
  \label{fig:lgd_champion}
\end{figure}

Within the exploratory four-champion comparison (TabPFN-3, CatBoost, RealMLP, and linear regression), the \gls{cd} diagram (Figure~\ref{fig:lgd_cd}) separates TabPFN-3 from linear regression, while RealMLP and CatBoost fall within the \gls{cd}. The Shaffer-adjusted matrix (Figure~\ref{fig:lgd_sig}) reports adjusted \(p\)-values below the conventional threshold for TabPFN-3 against RealMLP and linear regression, and for CatBoost against linear regression. The Bayesian analysis additionally assigns a high posterior probability to TabPFN-3 outperforming CatBoost, although the adjusted frequentist comparison does not separate them.

Across both tasks, the omnibus tests provide evidence of performance differences, while the full-panel post-hoc comparisons primarily separate the leading methods from the weakest because the panel is large relative to the number of datasets. The Bayesian comparisons generally favor TabPFN-3 over the selected conventional baselines, although practical equivalence retains substantial probability in some comparisons. The reduced champion analysis provides a focused but exploratory comparison because the family representatives were selected on the same benchmark.

\clearpage

\section{Additional performance measures}
\label{supp:metrics}

The main text reports a single primary measure per task (\gls{auc} for \gls{pd} classification and $R^2$ for \gls{lgd} regression). A single measure can, however, mask trade-offs, so we report a broader battery here, grouped by the aspect of performance it captures. The leading foundation models remain among the strongest methods across most discrimination, probabilistic-scoring, error, fit, and correlation measures, while other methods lead on some threshold-based, robust, or worst-case measures.

\Needspace{8\baselineskip}
\subsection{PD classification measures}
For \gls{pd} classification we group the measures into three families. The discrimination measures (\gls{auc}, the Gini coefficient, the \gls{ks} statistic, and adjusted \gls{ap}) assess how well the predicted scores rank defaulters above non-defaulters and are the quantities of primary interest in credit scoring. \gls{ap} summarizes the precision--recall curve as a recall-weighted mean of precision across score thresholds. Precision--recall measures provide a useful complementary perspective in imbalanced classification because they explicitly reflect the quality of predictions for the positive class \citep{2015.SAITO.precisionrecallplotmoreinformativerocplotwhenevaluatingbinaryclassifiersimbalanceddatasets}. Unlike \gls{auc}, AP depends on the positive-class prevalence $\pi$: the random-ranking reference value of AP is $\pi$. We therefore report adjusted \gls{ap}, following the chance adjustment applied to AP by \citet{2016.CAMPOS.evaluationunsupervisedoutlierdetection},
\[
    \mathrm{AP}_{\mathrm{adj}}
    =
    \frac{\mathrm{AP}-\pi}{1-\pi},
\]
where $\pi$ is the fold-specific default rate. The adjustment maps the random-ranking reference to zero and perfect AP to one, while values below the random reference can be negative. It aligns these reference points across datasets but does not make AP fully invariant to class prevalence; precision-based metrics remain sensitive to the class prior \citep{2020.SIBLINI.mastermetricscalibration}. The proper scoring rules (the Brier score and log loss) assess overall probabilistic accuracy; they combine calibration and refinement rather than isolating calibration. The threshold-based measures (accuracy, balanced accuracy, $F_1$, precision, recall, and the \gls{mcc}) summarize the confusion matrix at a single operating point; we set that threshold for each model by maximizing the $F_1$ score on the validation split. Table~\ref{tab:metricspd} reports all of them. TabPFN-3 attains the best value on every discrimination measure and on both proper scoring rules, and on most threshold-based measures; the isolated exception, top recall for XGBoost, reflects a different operating point (XGBoost gains recall by predicting more positives, at a cost in precision). The ordering among the foundation models is the same as on \gls{auc}, indicating that the ranking is not an artifact of the primary measure.

\begin{table}[ht!]
\centering
\caption{Average performance of all methods on the \gls{pd} (classification) benchmark across 14 datasets and 5 folds. The tunable methods are tuned by \gls{hpo} on a per-fold validation split; the \glspl{tfm} are run untuned in their default in-context configuration. Arrows indicate optimization direction ($\downarrow$ lower is better; all others higher is better). The best value in each column is shown in \textbf{bold}.}
\label{tab:metricspd}
\resizebox{\textwidth}{!}{
\begin{tabular}{l c c c c c c c c c c c c}
\toprule
\multirow{2}{*}{Method}
  & \multicolumn{4}{c}{Discrimination}
  & \multicolumn{2}{c}{Proper scoring rules}
  & \multicolumn{6}{c}{Classification (threshold-based)} \\
\cmidrule(lr){2-5}\cmidrule(lr){6-7}\cmidrule(lr){8-13}
  & \gls{auc} & Gini & \gls{ks} & AP$_{\text{adj}}$
  & Brier$\downarrow$ & LogLoss$\downarrow$
  & Acc & Bal-Acc & F1 & Precision & Recall & \gls{mcc} \\
\midrule
\multicolumn{13}{l}{\textit{Foundation models}} \\
TabPFN-3      & \textbf{0.7644} & \textbf{0.5288} & \textbf{0.4277} & \textbf{0.3779} & \textbf{0.1251} & \textbf{0.3886} & \textbf{0.7271} & \textbf{0.6920} & \textbf{0.5294} & \textbf{0.4602} & 0.6746 & \textbf{0.3600} \\
TabICLv2      & 0.7597 & 0.5194 & 0.4182 & 0.3728 & 0.1266 & 0.3928 & 0.7151 & 0.6843 & 0.5224 & 0.4522 & 0.6783 & 0.3462 \\
TabPFN-2.5    & 0.7572 & 0.5145 & 0.4137 & 0.3696 & 0.1267 & 0.3953 & 0.7267 & 0.6824 & 0.5190 & 0.4566 & 0.6526 & 0.3454 \\
TabICL        & 0.7526 & 0.5053 & 0.3994 & 0.3540 & 0.1567 & 0.4918 & 0.7167 & 0.6738 & 0.5064 & 0.4409 & 0.6485 & 0.3285 \\
TabDPT        & 0.7493 & 0.4986 & 0.3950 & 0.3521 & 0.1302 & 0.4051 & 0.7085 & 0.6750 & 0.5050 & 0.4396 & 0.6584 & 0.3273 \\
TabPFNv2      & 0.7395 & 0.4790 & 0.3823 & 0.3433 & 0.1315 & 0.4109 & 0.6983 & 0.6681 & 0.5030 & 0.4381 & 0.6612 & 0.3194 \\
Real-TabPFN   & 0.7393 & 0.4786 & 0.3828 & 0.3427 & 0.1315 & 0.4110 & 0.6961 & 0.6674 & 0.5029 & 0.4378 & 0.6646 & 0.3201 \\
Mitra         & 0.7232 & 0.4464 & 0.3461 & 0.3052 & 0.1390 & 0.4324 & 0.6451 & 0.6457 & 0.4706 & 0.3936 & 0.6815 & 0.2734 \\
TabPFN        & 0.7031 & 0.4063 & 0.3152 & 0.2729 & 0.1426 & 0.4455 & 0.6498 & 0.6353 & 0.4571 & 0.3742 & 0.6622 & 0.2489 \\
\midrule
\multicolumn{13}{l}{\textit{\gls{gb}}} \\
CatBoost      & 0.7496 & 0.4992 & 0.3982 & 0.3521 & 0.1333 & 0.4144 & 0.7131 & 0.6750 & 0.5079 & 0.4408 & 0.6550 & 0.3281 \\
LightGBM      & 0.7472 & 0.4943 & 0.3954 & 0.3505 & 0.1320 & 0.4185 & 0.6940 & 0.6720 & 0.5070 & 0.4368 & 0.6793 & 0.3224 \\
XGBoost       & 0.7253 & 0.4506 & 0.3611 & 0.3345 & 0.1346 & 0.4238 & 0.6343 & 0.6555 & 0.4989 & 0.4292 & \textbf{0.7264} & 0.2971 \\
\midrule
\multicolumn{13}{l}{\textit{Classical \gls{ml}}} \\
Random Forest & 0.7385 & 0.4770 & 0.3747 & 0.3316 & 0.1358 & 0.4225 & 0.6950 & 0.6601 & 0.4880 & 0.4216 & 0.6476 & 0.3004 \\
Logistic Reg. & 0.7147 & 0.4294 & 0.3334 & 0.2675 & 0.1439 & 0.4459 & 0.6665 & 0.6407 & 0.4592 & 0.3766 & 0.6484 & 0.2573 \\
\gls{knn}           & 0.7140 & 0.4279 & 0.3399 & 0.2914 & 0.1385 & 0.5766 & 0.6643 & 0.6495 & 0.4782 & 0.4005 & 0.6763 & 0.2804 \\
\gls{svm}           & 0.7092 & 0.4185 & 0.3270 & 0.2619 & 0.1435 & 0.4454 & 0.6620 & 0.6402 & 0.4578 & 0.3734 & 0.6513 & 0.2557 \\
\gls{ncm}           & 0.6967 & 0.3934 & 0.3097 & 0.2424 & 0.2188 & 0.6263 & 0.6474 & 0.6315 & 0.4491 & 0.3622 & 0.6589 & 0.2379 \\
Naive Bayes   & 0.6856 & 0.3713 & 0.2966 & 0.2268 & 0.3138 & 2.3542 & 0.6156 & 0.6185 & 0.4328 & 0.3412 & 0.6729 & 0.2142 \\
\midrule
\multicolumn{13}{l}{\textit{\gls{dl}}} \\
T2G-Former    & 0.7493 & 0.4986 & 0.3948 & 0.3489 & 0.1312 & 0.4141 & 0.7097 & 0.6750 & 0.5073 & 0.4397 & 0.6600 & 0.3295 \\
FTT           & 0.7483 & 0.4965 & 0.3949 & 0.3458 & 0.1313 & 0.4127 & 0.7010 & 0.6740 & 0.5052 & 0.4344 & 0.6701 & 0.3251 \\
AMFormer      & 0.7473 & 0.4945 & 0.3931 & 0.3505 & 0.1319 & 0.4193 & 0.7117 & 0.6739 & 0.5049 & 0.4419 & 0.6508 & 0.3274 \\
AutoInt       & 0.7459 & 0.4918 & 0.3919 & 0.3467 & 0.1320 & 0.4234 & 0.7044 & 0.6724 & 0.5055 & 0.4377 & 0.6617 & 0.3247 \\
RealMLP       & 0.7423 & 0.4847 & 0.3869 & 0.3441 & 0.1652 & 0.5153 & 0.7084 & 0.6705 & 0.5032 & 0.4447 & 0.6483 & 0.3247 \\
TabM          & 0.7418 & 0.4837 & 0.3840 & 0.3388 & 0.1334 & 0.4165 & 0.7073 & 0.6671 & 0.4965 & 0.4370 & 0.6429 & 0.3161 \\
SNN           & 0.7406 & 0.4811 & 0.3825 & 0.3331 & 0.1404 & 0.5463 & 0.7056 & 0.6662 & 0.4936 & 0.4316 & 0.6422 & 0.3136 \\
\gls{mlp}           & 0.7399 & 0.4798 & 0.3804 & 0.3355 & 0.1343 & 0.4701 & 0.6941 & 0.6668 & 0.4981 & 0.4350 & 0.6603 & 0.3174 \\
DCN2          & 0.7399 & 0.4798 & 0.3821 & 0.3322 & 0.1344 & 0.5115 & 0.7009 & 0.6674 & 0.4982 & 0.4363 & 0.6529 & 0.3166 \\
ExcelFormer   & 0.7384 & 0.4769 & 0.3798 & 0.3329 & 0.1341 & 0.4182 & 0.6978 & 0.6658 & 0.4970 & 0.4310 & 0.6539 & 0.3125 \\
TANGOS        & 0.7384 & 0.4768 & 0.3750 & 0.3266 & 0.1364 & 0.4355 & 0.6904 & 0.6623 & 0.4902 & 0.4178 & 0.6613 & 0.3049 \\
DANets        & 0.7380 & 0.4761 & 0.3749 & 0.3339 & 0.1344 & 0.4369 & 0.6926 & 0.6658 & 0.4950 & 0.4340 & 0.6576 & 0.3145 \\
\gls{mlp}-PLR       & 0.7364 & 0.4728 & 0.3840 & 0.3389 & 0.1351 & 0.4566 & 0.6905 & 0.6646 & 0.4974 & 0.4310 & 0.6605 & 0.3132 \\
SwitchTab     & 0.7245 & 0.4489 & 0.3499 & 0.3032 & 0.1387 & 0.4318 & 0.6814 & 0.6524 & 0.4761 & 0.4050 & 0.6496 & 0.2849 \\
TabNet        & 0.7135 & 0.4270 & 0.3373 & 0.2980 & 0.1387 & 0.4326 & 0.6641 & 0.6426 & 0.4715 & 0.4074 & 0.6466 & 0.2709 \\
\bottomrule
\end{tabular}
}
\end{table}

\clearpage
\Needspace{8\baselineskip}
\subsection{LGD regression measures}
For \gls{lgd} regression we report two families of measures. The error and fit measures ($R^2$, the \gls{mse}, the \gls{rmse}, the \gls{mae}, the \gls{medae}, and the maximum error) summarize how close the predictions are to the realized losses; $R^2$, the \gls{mae}, and the \gls{medae} are the quantities most commonly emphasized in the \gls{lgd} literature. The correlation and explained-variance measures (the Pearson and Spearman correlations and the explained variance) summarize how well the predictions track the realized losses. Table~\ref{tab:metricslgd} reports all of them. The three leading foundation models (TabPFN-3, TabPFN-2.5, and TabICLv2) occupy the top of $R^2$, the \gls{rmse}, both correlations, and the explained variance; TabPFN-3 is best on most of these measures, while TabPFN-2.5 has the lowest \gls{mae}. ModernNCA has the lowest \gls{medae}, a robust measure of typical error, whereas Mitra has the lowest maximum error, a worst-case measure. These exceptions do not alter the broader pattern across fit, mean-error, and correlation measures.

\begin{table}[ht!]
\centering
\caption{Average performance of all methods on the \gls{lgd} (regression) benchmark across 7 datasets and 5 folds. The tunable methods are tuned by \gls{hpo} on a per-fold validation split; the \glspl{tfm} are run untuned in their default in-context configuration. Arrows indicate optimization direction ($\downarrow$ lower is better; all others higher is better). The best value in each column is shown in \textbf{bold}.}
\label{tab:metricslgd}
\resizebox{\textwidth}{!}{
\begin{tabular}{l c c c c c c c c c}
\toprule
\multirow{2}{*}{Method}
  & \multicolumn{6}{c}{Error / Fit}
  & \multicolumn{3}{c}{Correlation \& Variance} \\
\cmidrule(lr){2-7}\cmidrule(lr){8-10}
  & $R^2$ & MSE$\downarrow$ & RMSE$\downarrow$ & MAE$\downarrow$ & MedAE$\downarrow$ & MaxErr$\downarrow$
  & Pearson & Spearman & Expl.\ Var \\
\midrule
\multicolumn{10}{l}{\textit{Foundation models}} \\
TabPFN-3      & \textbf{0.5366} & \textbf{0.0555} & \textbf{0.2234} & 0.1551 & 0.1022 & 0.8898 & \textbf{0.7139} & \textbf{0.6686} & \textbf{0.5393} \\
TabPFN-2.5    & 0.5305 & 0.0561 & 0.2237 & \textbf{0.1534} & 0.0992 & 0.8906 & 0.7069 & 0.6618 & 0.5355 \\
TabICLv2      & 0.5298 & 0.0565 & 0.2263 & 0.1608 & 0.1108 & 0.8863 & 0.7088 & 0.6632 & 0.5318 \\
TabDPT        & 0.4809 & 0.0633 & 0.2372 & 0.1705 & 0.1190 & 0.8875 & 0.6709 & 0.6379 & 0.4844 \\
TabPFNv2      & 0.4408 & 0.0710 & 0.2522 & 0.1890 & 0.1433 & 0.8740 & 0.6534 & 0.6053 & 0.4597 \\
Mitra         & 0.1656 & 0.1033 & 0.3118 & 0.2578 & 0.2137 & \textbf{0.7596} & 0.5644 & 0.5120 & 0.2631 \\
\midrule
\multicolumn{10}{l}{\textit{\gls{gb}}} \\
CatBoost      & 0.4872 & 0.0620 & 0.2359 & 0.1728 & 0.1242 & 0.9052 & 0.6790 & 0.6305 & 0.4907 \\
LightGBM      & 0.4538 & 0.0670 & 0.2471 & 0.1813 & 0.1325 & 0.9076 & 0.6561 & 0.6080 & 0.4611 \\
XGBoost       & 0.4509 & 0.0673 & 0.2472 & 0.1854 & 0.1398 & 0.8998 & 0.6499 & 0.6065 & 0.4555 \\
\midrule
\multicolumn{10}{l}{\textit{Classical \gls{ml}}} \\
Random Forest & 0.4203 & 0.0718 & 0.2565 & 0.1913 & 0.1396 & 0.8845 & 0.6295 & 0.5810 & 0.4268 \\
Linear Reg.   & 0.3426 & 0.0824 & 0.2766 & 0.2126 & 0.1719 & 0.9334 & 0.5709 & 0.5323 & 0.3448 \\
\gls{knn}           & 0.3210 & 0.0822 & 0.2785 & 0.2193 & 0.1762 & 0.8487 & 0.5644 & 0.5197 & 0.3275 \\
\gls{svm}           & 0.3017 & 0.0871 & 0.2836 & 0.2078 & 0.1462 & 0.9241 & 0.5601 & 0.5282 & 0.3259 \\
\midrule
\multicolumn{10}{l}{\textit{\gls{dl}}} \\
RealMLP       & 0.4502 & 0.0678 & 0.2484 & 0.1752 & 0.1191 & 0.9212 & 0.6556 & 0.6137 & 0.4560 \\
TabM          & 0.4477 & 0.0679 & 0.2486 & 0.1785 & 0.1259 & 0.9265 & 0.6554 & 0.6057 & 0.4519 \\
ExcelFormer   & 0.4320 & 0.0706 & 0.2535 & 0.1787 & 0.1226 & 0.9361 & 0.6354 & 0.5891 & 0.4398 \\
\gls{mlp}-PLR       & 0.4271 & 0.0707 & 0.2507 & 0.1716 & 0.1054 & 0.9340 & 0.6464 & 0.5953 & 0.4428 \\
T2G-Former    & 0.4162 & 0.0732 & 0.2580 & 0.1739 & 0.1031 & 0.9419 & 0.6493 & 0.5999 & 0.4311 \\
FTT           & 0.4140 & 0.0734 & 0.2577 & 0.1735 & 0.1029 & 0.9362 & 0.6461 & 0.6038 & 0.4282 \\
DCN2          & 0.3983 & 0.0762 & 0.2630 & 0.1881 & 0.1353 & 0.9388 & 0.6172 & 0.5775 & 0.4038 \\
\gls{mlp}           & 0.3897 & 0.0767 & 0.2646 & 0.1848 & 0.1177 & 0.9261 & 0.6211 & 0.5809 & 0.3986 \\
TANGOS        & 0.3894 & 0.0764 & 0.2645 & 0.1845 & 0.1168 & 0.9288 & 0.6173 & 0.5767 & 0.4004 \\
ResNet        & 0.3891 & 0.0769 & 0.2652 & 0.1864 & 0.1222 & 0.9414 & 0.6216 & 0.5729 & 0.3979 \\
AutoInt       & 0.3602 & 0.0805 & 0.2690 & 0.1876 & 0.1194 & 0.9351 & 0.6028 & 0.5666 & 0.4023 \\
TabNet        & 0.2843 & 0.0927 & 0.2879 & 0.2250 & 0.1748 & 0.9349 & 0.4861 & 0.4591 & 0.2919 \\
\midrule
\multicolumn{10}{l}{\textit{Specialized / other}} \\
PTARL         & 0.4129 & 0.0739 & 0.2605 & 0.1950 & 0.1464 & 0.9077 & 0.6287 & 0.5900 & 0.4160 \\
ModernNCA     & 0.3974 & 0.0752 & 0.2488 & 0.1657 & \textbf{0.0968} & 0.9356 & 0.6313 & 0.5970 & 0.4114 \\
\bottomrule
\end{tabular}
}
\end{table}

\clearpage
\section{Additional visualizations and supporting results}
\label{supp:viz}

This section collects additional visualizations that complement the main text. We first give a dataset-level view of performance across the full method panel, then two head-to-head comparisons of the strongest foundation model, TabPFN-3, against the \gls{gb} champion (CatBoost) and against the linear industry baseline (logistic regression for \gls{pd} classification, linear regression for \gls{lgd} regression), and finally an analysis of how the relative advantage of TabPFN-3 over the \gls{gb} champion varies with dataset size.

\Needspace{8\baselineskip}
\subsection{Dataset-level performance matrix}
\label{supp:viz_heatmap}
\begin{figure}[ht!]
  \centering
  \includegraphics[width=\textwidth,height=0.90\textheight,keepaspectratio]{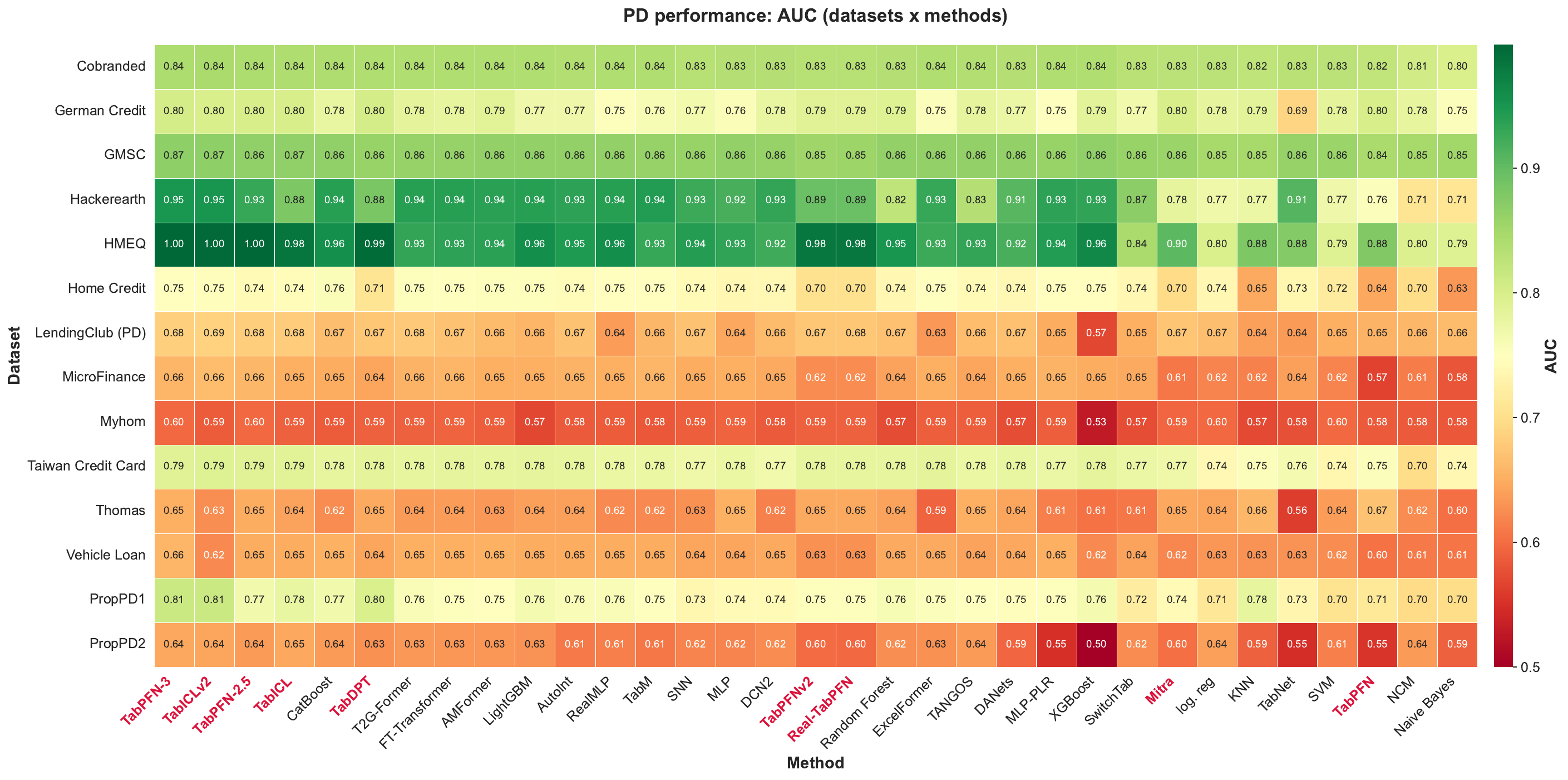}
  \caption{Per-dataset \gls{auc} of each method (columns) on every \gls{pd} dataset (rows). Cell values are five-fold means; method columns are ordered best to worst and shaded green (best) to red (worst). Foundation models are highlighted in red.}
  \label{fig:heatmap_pd}
\end{figure}

\begin{figure}[ht!]
  \centering
  \includegraphics[width=\textwidth,height=0.90\textheight,keepaspectratio]{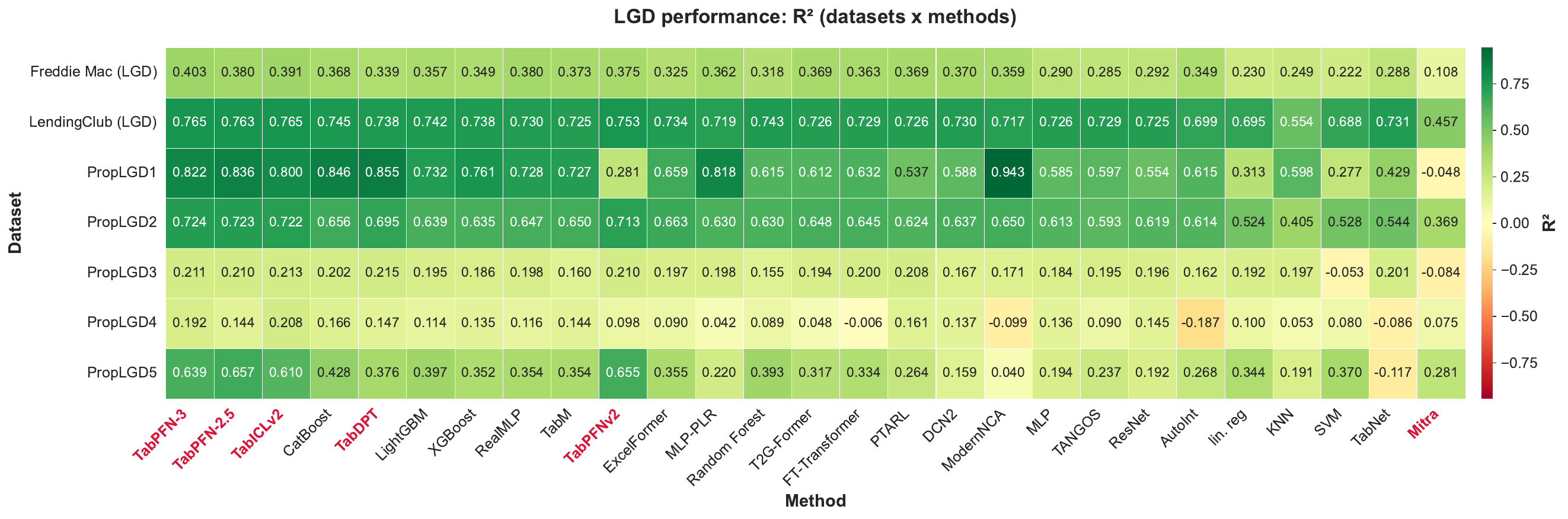}
  \caption{Per-dataset $R^2$ of each method (columns) on every \gls{lgd} dataset (rows). Cell values are five-fold means; method columns are ordered best to worst and shaded green (best) to red (worst). Foundation models are highlighted in red.}
  \label{fig:heatmap_lgd}
\end{figure}

The aggregate rankings in the main text average over datasets and can obscure dataset-level variation. Figures~\ref{fig:heatmap_pd} and~\ref{fig:heatmap_lgd} report the per-dataset performance of every method, \gls{auc} for \gls{pd} classification and $R^2$ for \gls{lgd} regression, with methods ordered best-first and cells shaded from green (best) to red (worst). The matrices confirm that the advantage of the leading foundation models is broad rather than driven by a few datasets, and they also expose the heterogeneity discussed in the main text, including the small number of datasets on which the gap to the strongest baselines narrows.

\clearpage

\Needspace{8\baselineskip}
\subsection{TabPFN-3 versus the gradient-boosting champion}
\label{supp:viz_catboost}
\begin{figure}[ht!]
  \centering
  \includegraphics[width=1\textwidth,height=0.90\textheight,keepaspectratio]{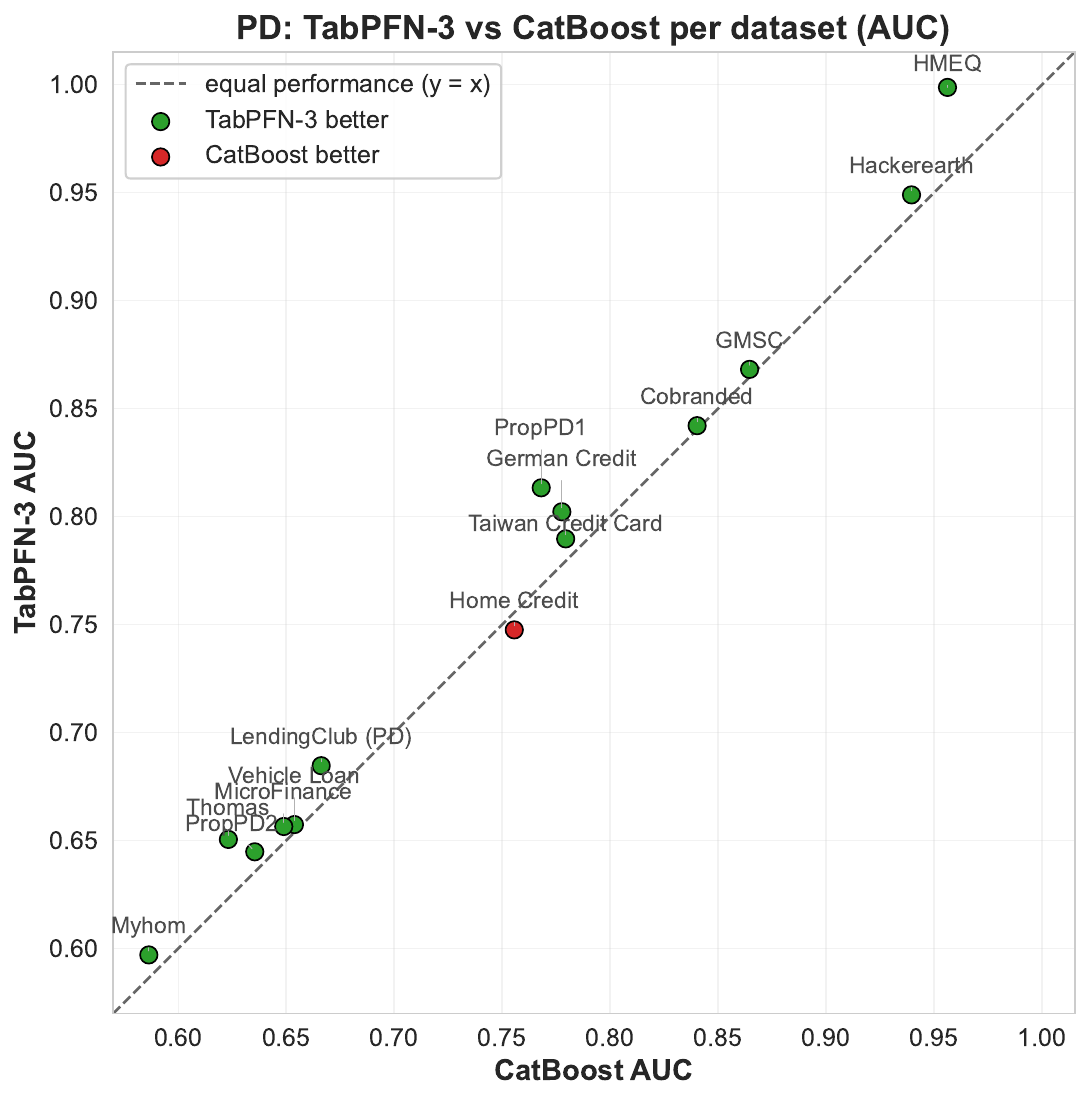}
  \caption{Per-dataset head-to-head of TabPFN-3 against CatBoost on \gls{pd} classification (\gls{auc}). Each point is a dataset; the diagonal marks equal performance, with green points favoring TabPFN-3 and red points favoring CatBoost.}
  \label{fig:vs_catboost_pd}
\end{figure}

\begin{figure}[ht!]
  \centering
  \includegraphics[width=1\textwidth,height=0.90\textheight,keepaspectratio]{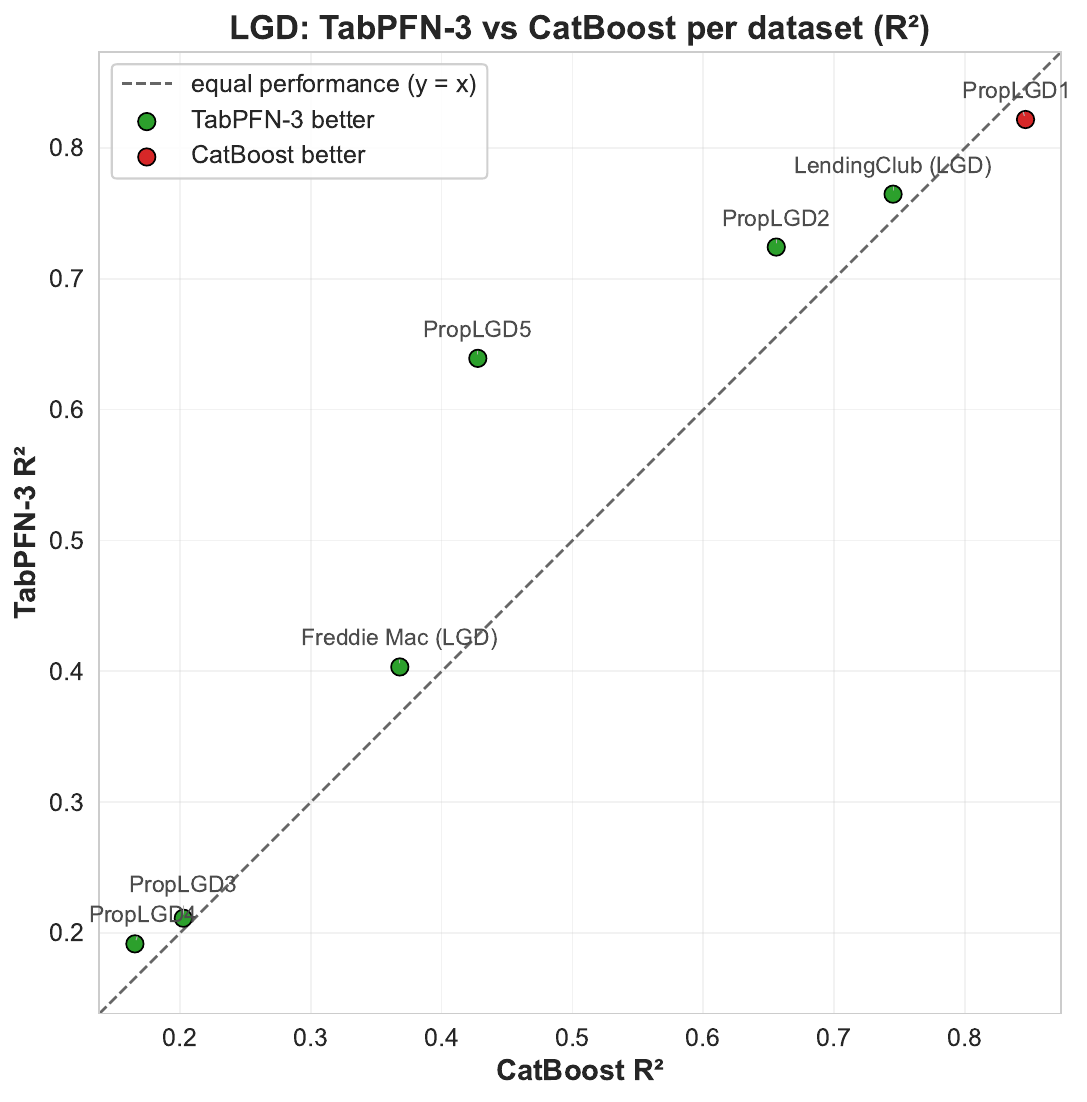}
  \caption{Per-dataset head-to-head of TabPFN-3 against CatBoost on \gls{lgd} regression ($R^2$). Each point is a dataset; the diagonal marks equal performance, with green points favoring TabPFN-3 and red points favoring CatBoost.}
  \label{fig:vs_catboost_lgd}
\end{figure}

\gls{gb} ensembles, and CatBoost in particular, are the strongest conventional benchmark in both tasks. Figures~\ref{fig:vs_catboost_pd} and~\ref{fig:vs_catboost_lgd} compare TabPFN-3 against CatBoost dataset by dataset, with the diagonal marking equal performance, so that points above it are datasets on which TabPFN-3 wins. TabPFN-3 attains the higher mean \gls{auc} on 13 of the 14 \gls{pd} datasets and the higher mean $R^2$ on 6 of the 7 \gls{lgd} datasets. The advantage is therefore consistent in sign on all but one dataset per task. Its mean is larger for \gls{lgd} regression, although the \gls{lgd} differences vary substantially across datasets.

\clearpage

\Needspace{8\baselineskip}
\subsection{TabPFN-3 versus the linear industry baseline}
\label{supp:viz_linear}
\begin{figure}[ht!]
  \centering
  \includegraphics[width=1\textwidth,height=0.90\textheight,keepaspectratio]{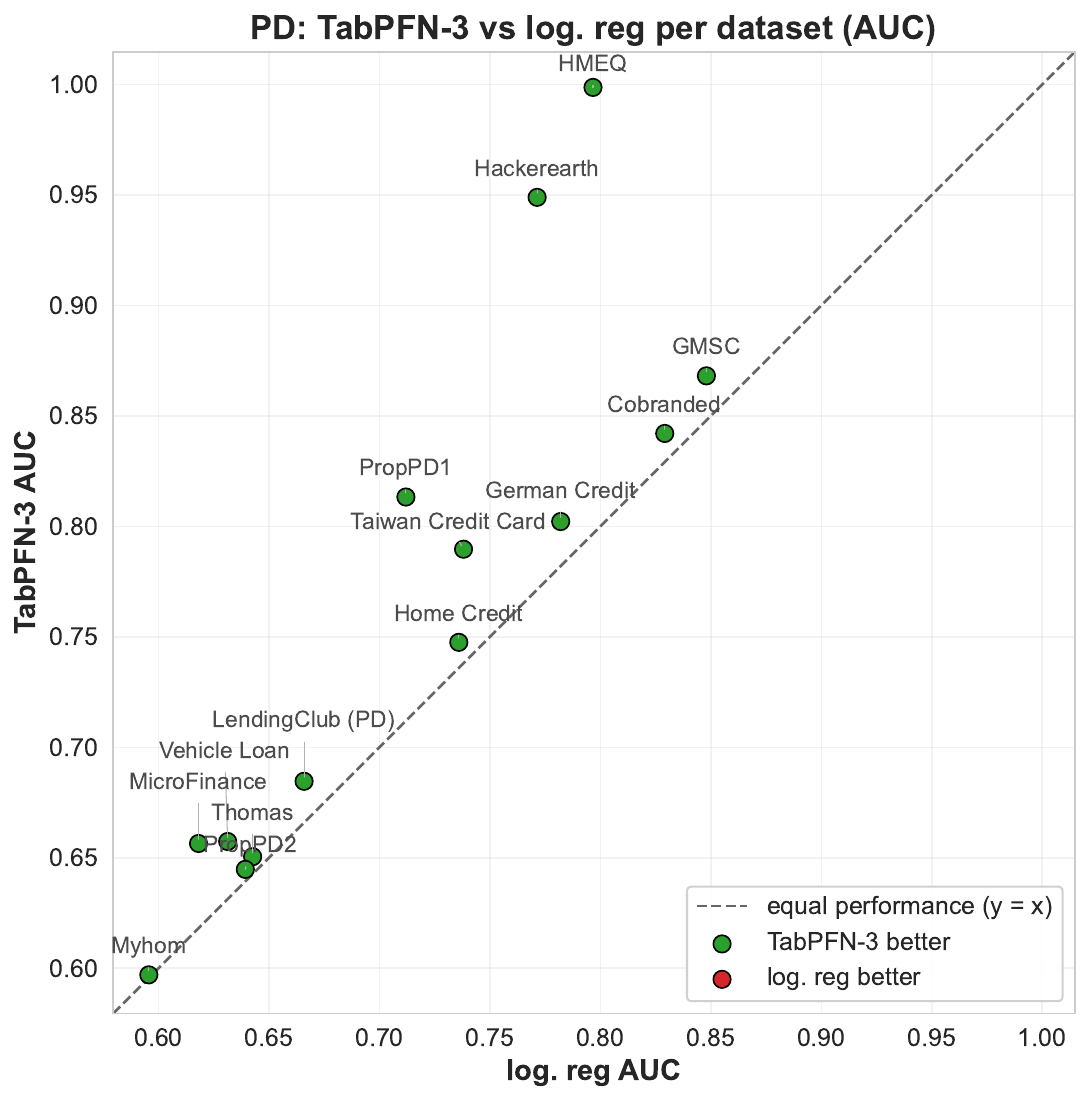}
  \caption{Per-dataset head-to-head of TabPFN-3 against logistic regression on \gls{pd} classification (\gls{auc}). Each point is a dataset; the diagonal marks equal performance, with green points favoring TabPFN-3 and red points favoring log.\ reg.}
  \label{fig:vs_linear_pd}
\end{figure}

\begin{figure}[ht!]
  \centering
  \includegraphics[width=1\textwidth,height=0.90\textheight,keepaspectratio]{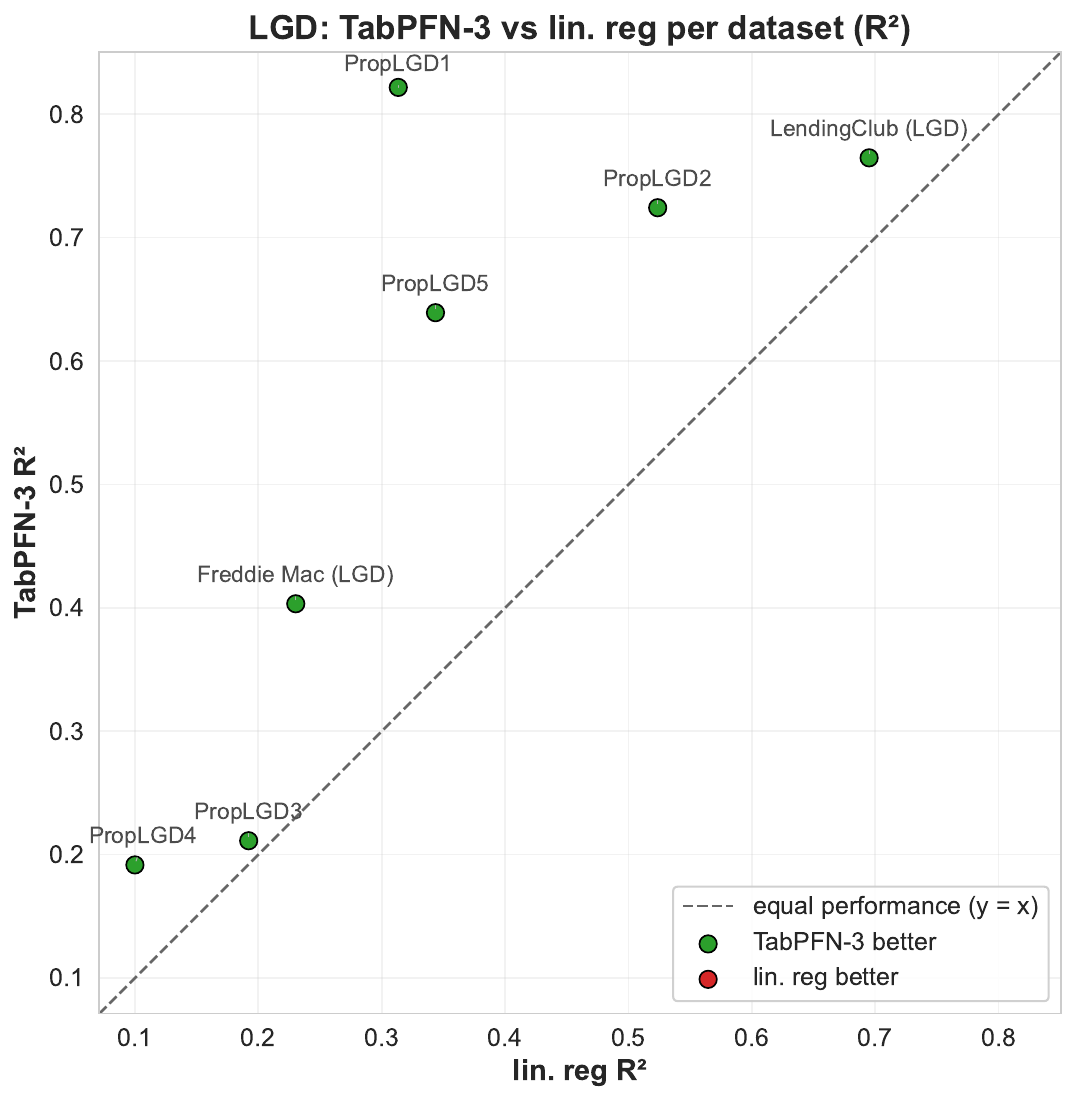}
  \caption{Per-dataset head-to-head of TabPFN-3 against linear regression on \gls{lgd} regression ($R^2$). Each point is a dataset; the diagonal marks equal performance, with green points favoring TabPFN-3 and red points favoring lin.\ reg.}
  \label{fig:vs_linear_lgd}
\end{figure}

Linear models---logistic regression for \gls{pd} classification and linear regression for \gls{lgd} regression---are included here not for their predictive performance but because they remain widely used reference models in credit scoring practice \citep{2015.LESSMANN.benchmarkingstateoftheartclassificationalgorithmscreditscoringupdateresearch,2020.DASTILE.statisticalmachinelearningmodelscreditscoringsystematicliteraturesurvey}. Figures~\ref{fig:vs_linear_pd} and~\ref{fig:vs_linear_lgd} contrast TabPFN-3 with these baselines directly. TabPFN-3 attains a higher mean \gls{auc} than logistic regression on all 14 \gls{pd} datasets and a higher mean $R^2$ than linear regression on all 7 \gls{lgd} datasets, making the descriptive direction of the difference consistent across the datasets in this benchmark.

\clearpage

\Needspace{8\baselineskip}
\subsection{Does the advantage over gradient boosting grow on small data?}
\label{supp:viz_sizetrend}
\begin{figure}[ht!]
  \centering
  \includegraphics[width=1\textwidth,height=0.90\textheight,keepaspectratio]{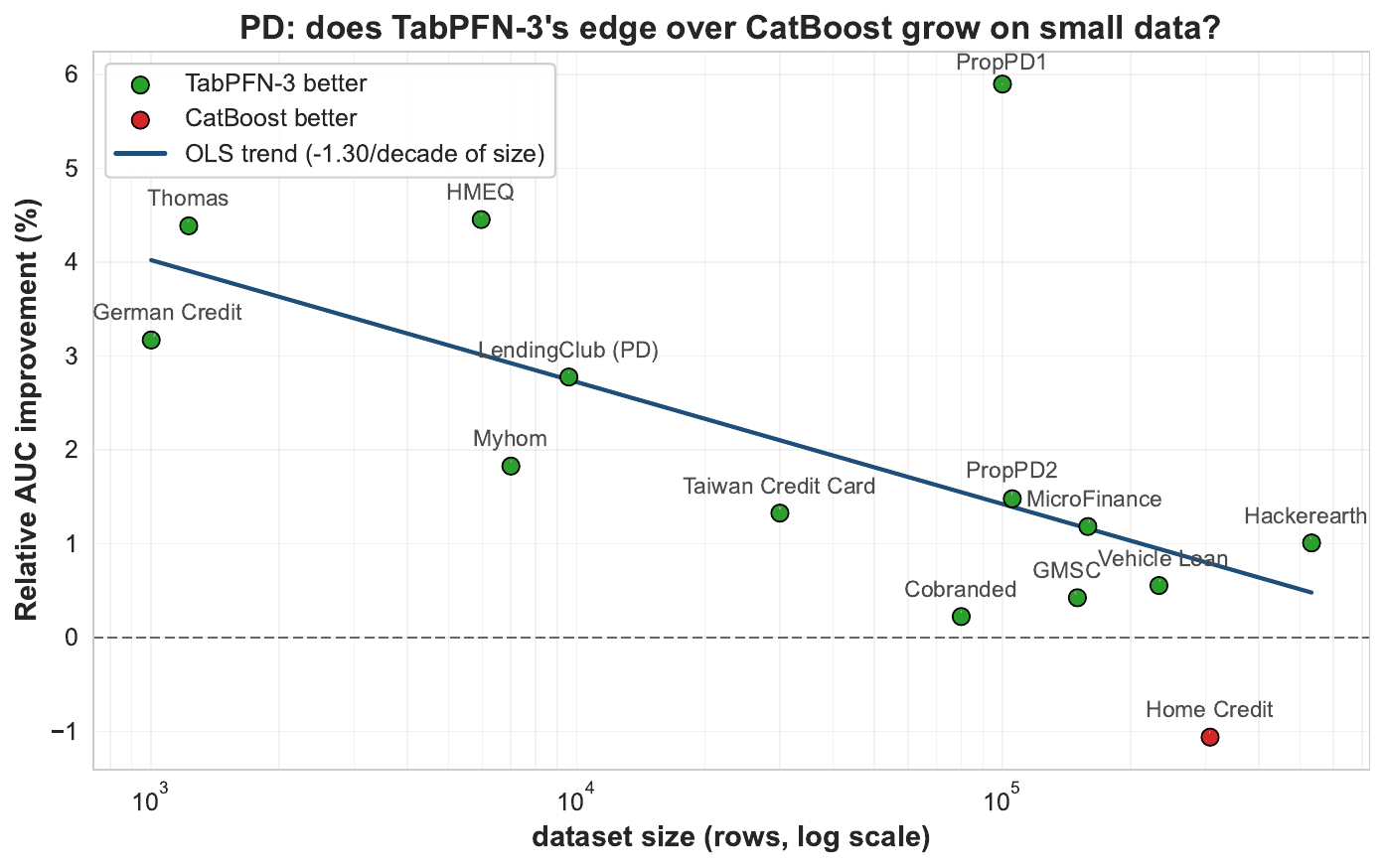}
  \caption{Per-dataset \emph{relative} \gls{auc} gain of TabPFN-3 over CatBoost (the improvement relative to CatBoost's own score, not the absolute difference; vertical axis) against \gls{pd} dataset size in rows (horizontal axis, log scale). Each point is a dataset; green points favor TabPFN-3, red points favor CatBoost, and the line is an \gls{ols} trend over log dataset size.}
  \label{fig:sizetrend_pd}
\end{figure}

\begin{figure}[ht!]
  \centering
  \includegraphics[width=1\textwidth,height=0.90\textheight,keepaspectratio]{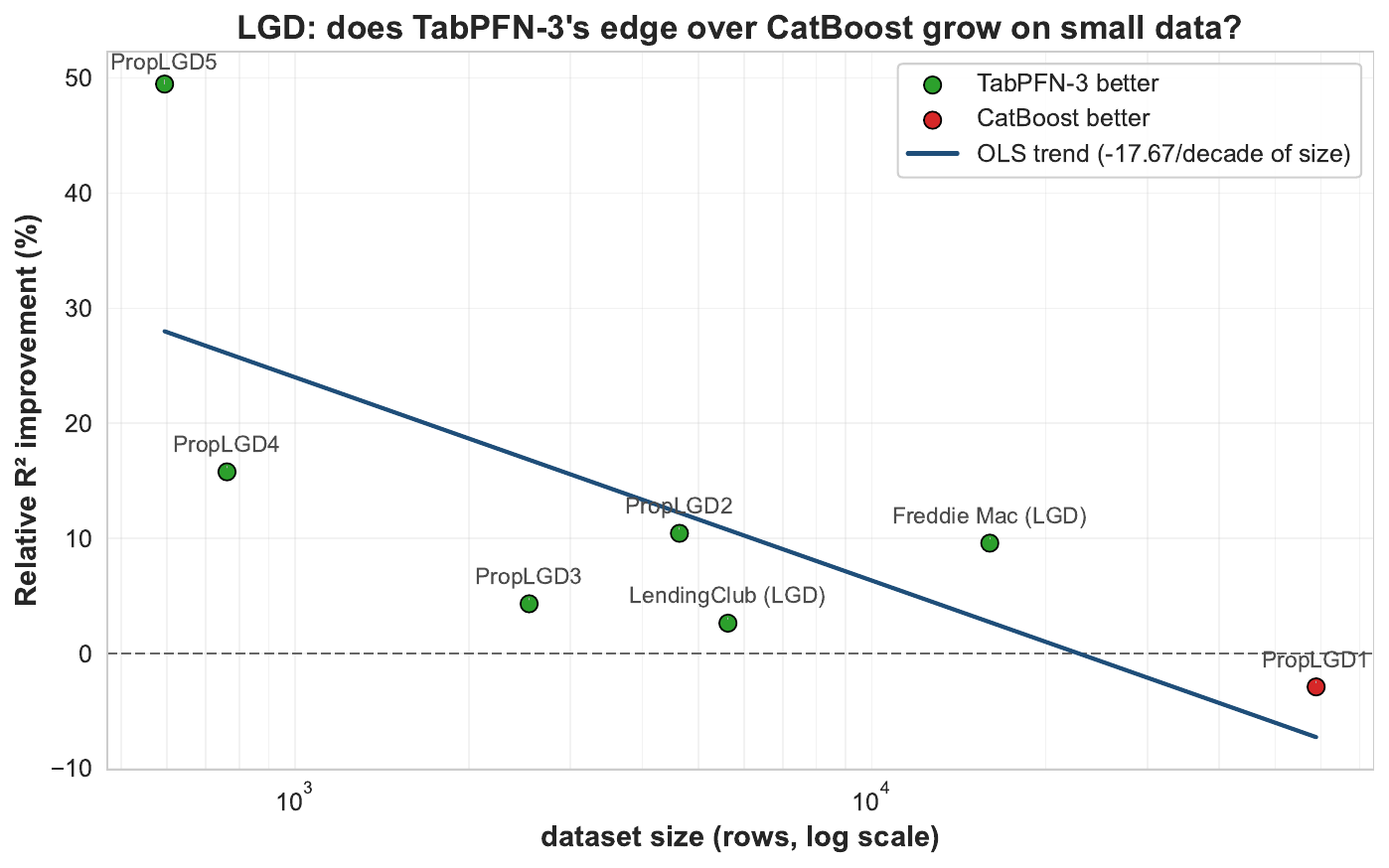}
  \caption{Per-dataset \emph{relative} $R^2$ gain of TabPFN-3 over CatBoost (the improvement relative to CatBoost's own score, not the absolute difference; vertical axis) against \gls{lgd} dataset size in rows (horizontal axis, log scale). Each point is a dataset; green points favor TabPFN-3, red points favor CatBoost, and the line is an \gls{ols} trend over log dataset size.}
  \label{fig:sizetrend_lgd}
\end{figure}

A natural question is whether the advantage of TabPFN-3 over the \gls{gb} champion is concentrated on smaller datasets, as earlier foundation-model evidence might suggest \citep{2025.HOLLMANN.accuratepredictionssmalldatatabularfoundationmodel,2026.ERICKSON.tabarenalivingbenchmarkmachinelearningtabulardata}. Figures~\ref{fig:sizetrend_pd} and~\ref{fig:sizetrend_lgd} plot, for each dataset at its full size, the \emph{relative} performance gain of TabPFN-3 over CatBoost, that is, the improvement expressed relative to CatBoost's own score rather than as an absolute difference, against the number of rows on a logarithmic axis; points above the zero line are datasets on which TabPFN-3 wins. Almost every point lies above zero: TabPFN-3 attains the higher five-fold mean on 13 of the 14 \gls{pd} datasets (mean difference $+0.0148$ \gls{auc}, median $+0.0099$) and on 6 of the 7 \gls{lgd} datasets (mean difference $+0.0493$ in $R^2$, median $+0.0261$). On both tasks the single dataset on which CatBoost wins is a large one: Home Credit for \gls{pd}, with $307{,}511$ observations, and PropLGD1 for \gls{lgd}, which with $58{,}862$ observations is the largest \gls{lgd} dataset in the benchmark.

Both panels show the same descriptive tendency, namely that the relative gain shrinks as datasets grow. The fitted trend declines by $1.30$ percentage points per tenfold increase in the number of rows for \gls{pd} classification and by $17.67$ percentage points for \gls{lgd} regression. For \gls{pd}, the relative gain is $+4.5\%$ on \gls{hmeq} ($5{,}960$ observations) and $+4.4\%$ on Thomas ($1{,}225$ observations), against $+0.2\%$ on Cobranded ($80{,}000$) and $+0.4\%$ on \gls{gmsc} ($150{,}000$). For \gls{lgd}, the same ordering is far steeper, running from $+49.5\%$ on PropLGD5 ($594$ observations) and $+15.8\%$ on PropLGD4 ($762$) down to $+2.6\%$ on LendingClub ($5{,}627$) and $-2.9\%$ on PropLGD1.

Three qualifications are in order. First, with 14 and 7 heterogeneous datasets respectively, these associations are descriptive and should not be interpreted causally or as population-level results; the scatter is substantial in both panels, since the largest \gls{pd} relative gain, $+5.9\%$ on PropPD1, occurs at $100{,}000$ observations, and on the \gls{lgd} side Freddie Mac attains $+9.6\%$ at $16{,}002$ observations, more than two of the five smaller \gls{lgd} datasets. Second, the two slopes are not directly comparable in magnitude, because dividing by CatBoost's own score amplifies differences wherever that score is low, which is common in \gls{lgd} regression. The direction of the \gls{lgd} trend is robust to its most extreme point but its magnitude is not: excluding PropLGD5 leaves the slope at $-7.5$ percentage points per tenfold increase. The \gls{pd} slope is stable in both respects, moving to $-1.5$ when its most influential point, PropPD1, is excluded. Third, dataset size is confounded with the other characteristics that vary across portfolios, including dimensionality and target distribution. Subject to these qualifications, the patterns on both tasks are consistent with, but do not independently confirm, the controlled subsampling experiment in the main text (\gls{rq}~2), in which sample size is varied within datasets while all other characteristics are held fixed.

\clearpage

\section{Per-dataset prediction densities}
\label{sec:supp-prediction-densities}

The dataset-level means in the main text compress each dataset to a single pair of numbers. The figures below show the full out-of-fold prediction distributions underlying those summaries, with one row per dataset and one column per compared method. For \gls{pd}, each panel contains the two class-conditional kernel densities of the predicted default probability. For \gls{lgd}, where the target is continuous, each panel contains a \gls{twod} kernel density of the raw prediction against the realized value.

\begin{figure}[ht!]
    \centering
    \includegraphics[width=1.0\textwidth,height=0.90\textheight,keepaspectratio]{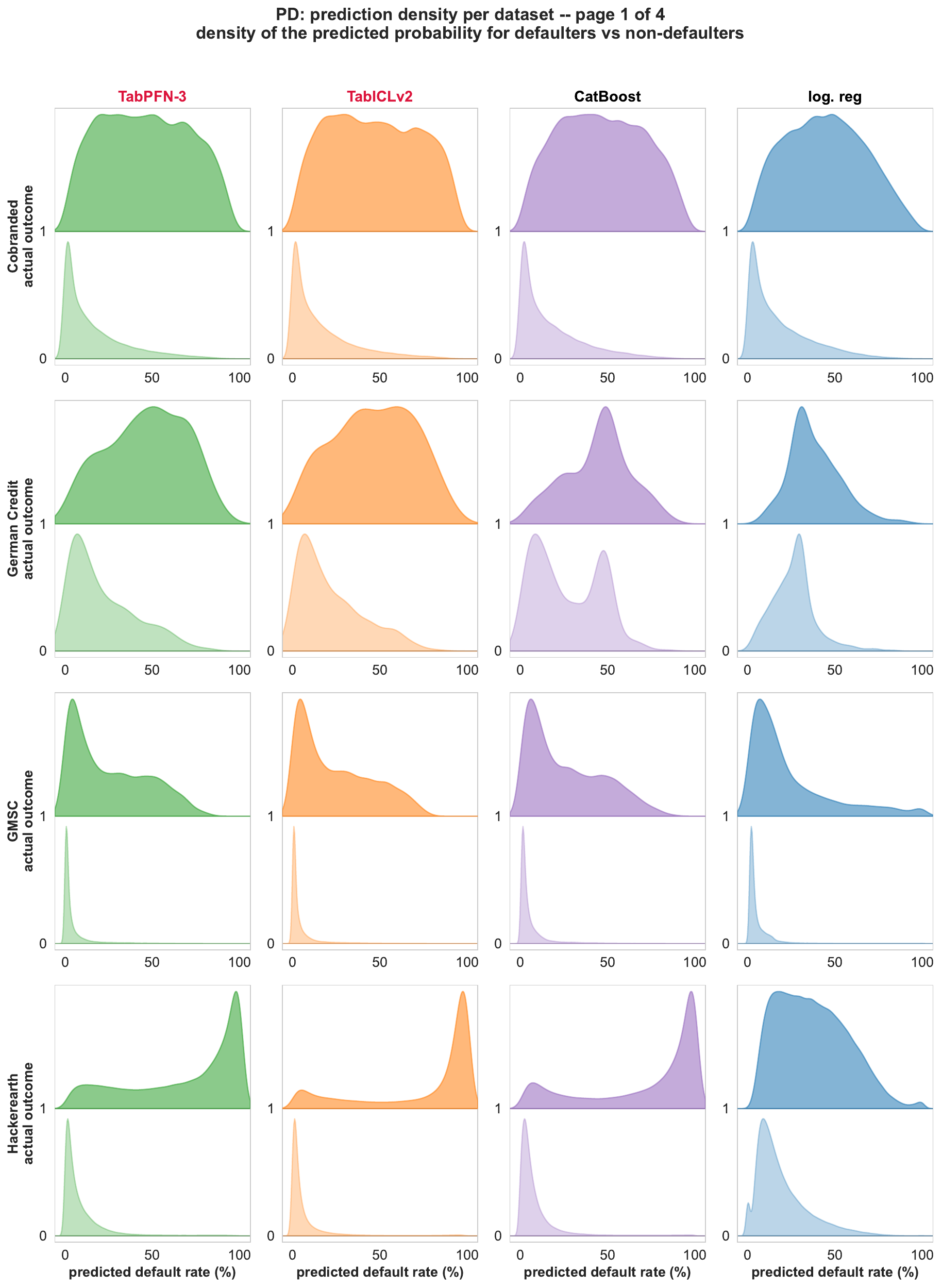}
    \caption{Prediction density per dataset (rows) and compared method (columns), page 1 of 4. Each panel contains the class-conditional kernel densities of the predicted default probability for non-defaulted and defaulted loans.}
    \label{fig:experiment1-pd-pd-per-dataset-prediction-density-p1}
\end{figure}

\begin{figure}[ht!]
    \centering
    \includegraphics[width=1.0\textwidth,height=0.90\textheight,keepaspectratio]{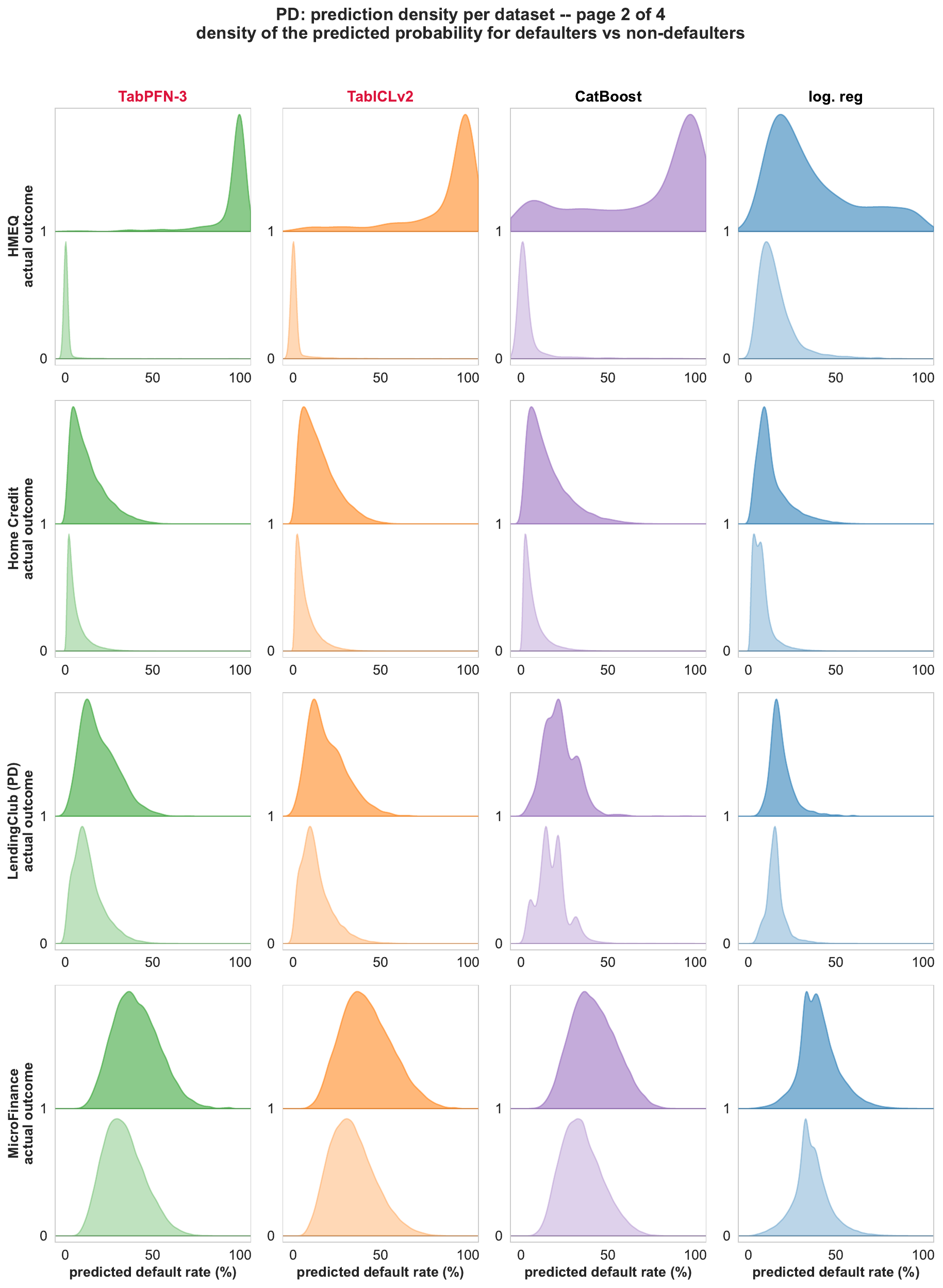}
    \caption{Prediction density per dataset (rows) and compared method (columns), page 2 of 4. Panels are constructed as in Figure~\ref{fig:experiment1-pd-pd-per-dataset-prediction-density-p1}.}
    \label{fig:experiment1-pd-pd-per-dataset-prediction-density-p2}
\end{figure}

\begin{figure}[ht!]
    \centering
    \includegraphics[width=1.0\textwidth,height=0.90\textheight,keepaspectratio]{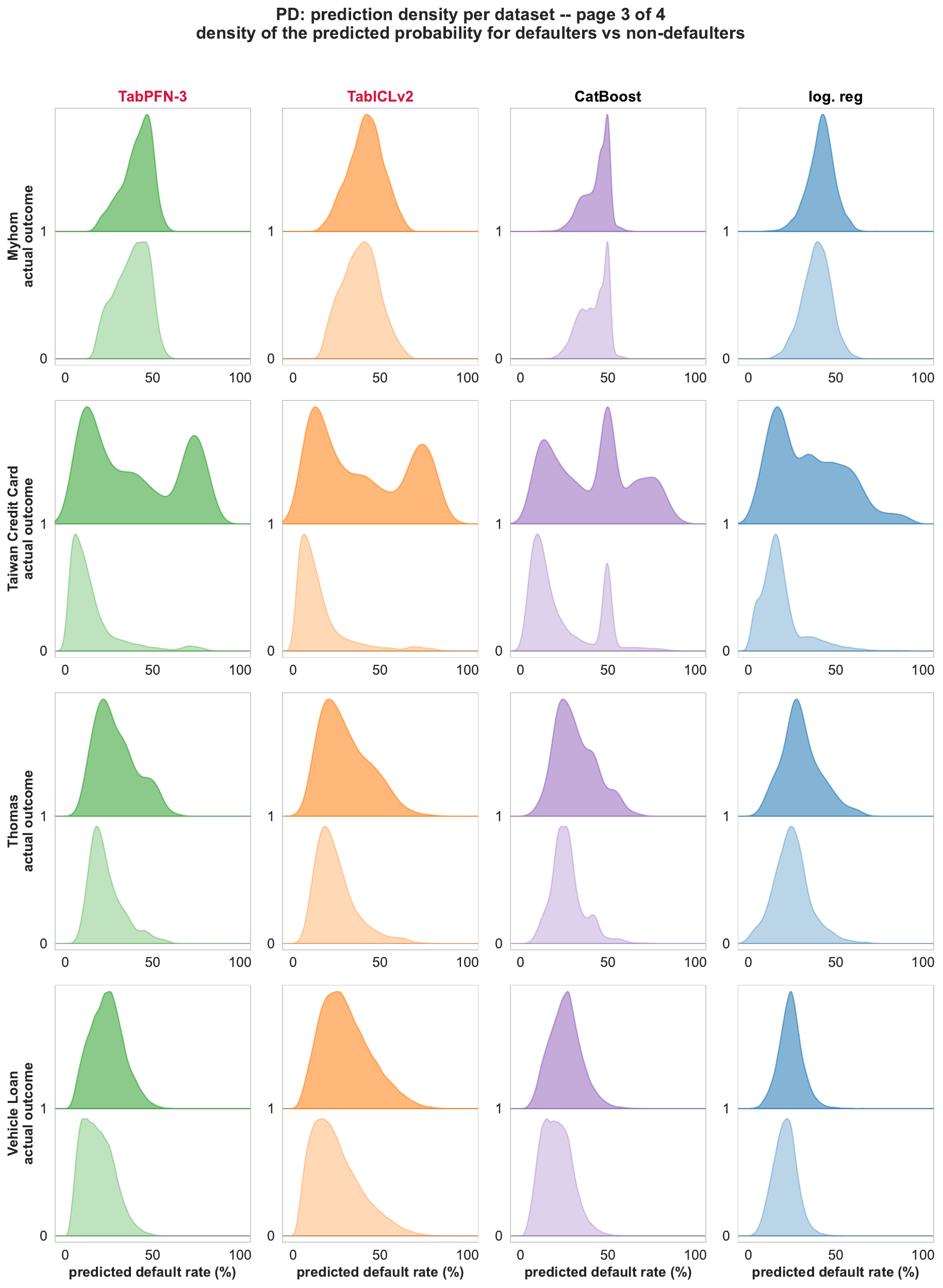}
    \caption{Prediction density per dataset (rows) and compared method (columns), page 3 of 4. Panels are constructed as in Figure~\ref{fig:experiment1-pd-pd-per-dataset-prediction-density-p1}.}
    \label{fig:experiment1-pd-pd-per-dataset-prediction-density-p3}
\end{figure}

\begin{figure}[ht!]
    \centering
    \includegraphics[width=1.0\textwidth,height=0.90\textheight,keepaspectratio]{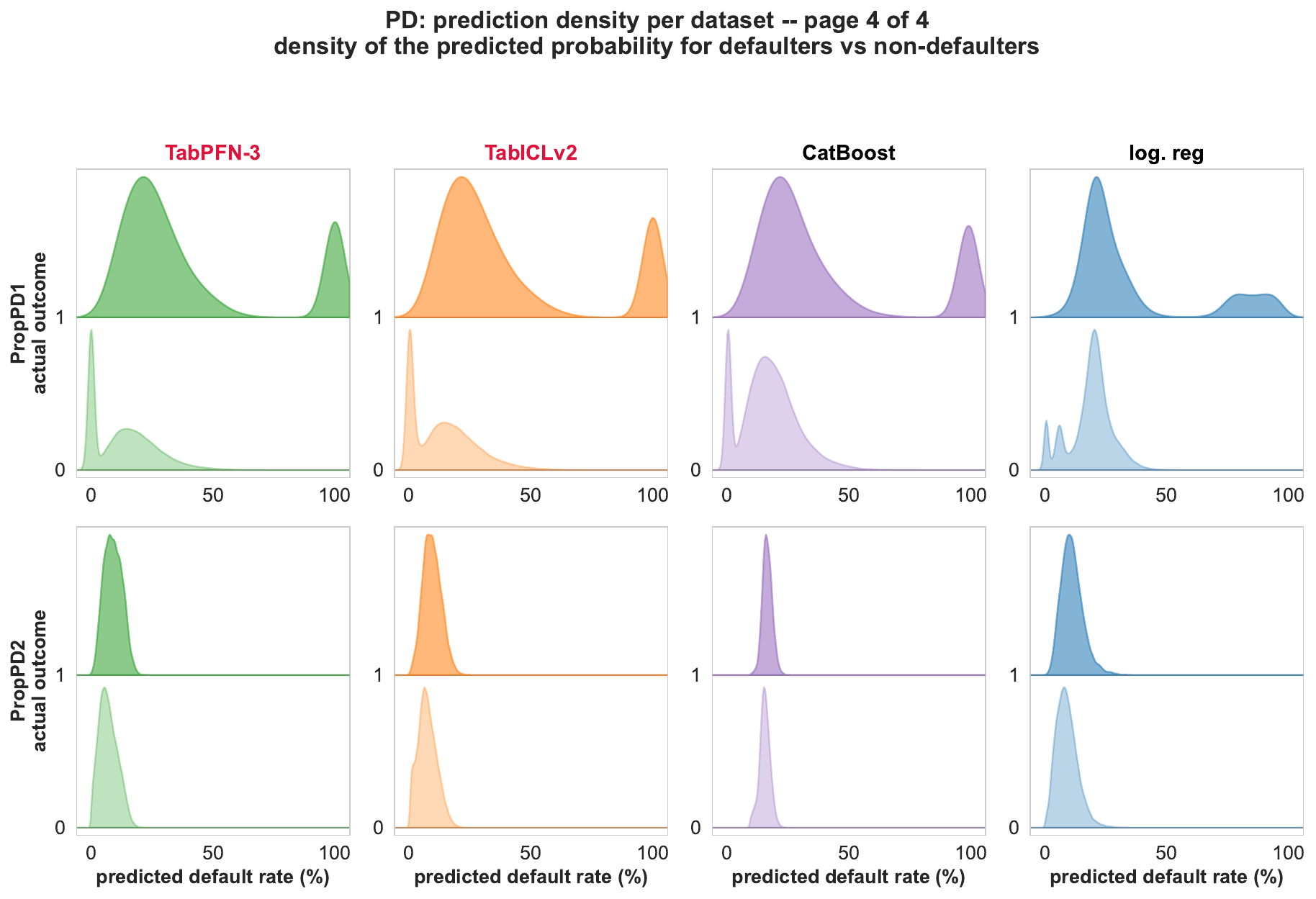}
    \caption{Prediction density per dataset (rows) and compared method (columns), page 4 of 4. Panels are constructed as in Figure~\ref{fig:experiment1-pd-pd-per-dataset-prediction-density-p1}.}
    \label{fig:experiment1-pd-pd-per-dataset-prediction-density-p4}
\end{figure}

\begin{figure}[ht!]
    \centering
    \includegraphics[width=1.0\textwidth,height=0.90\textheight,keepaspectratio]{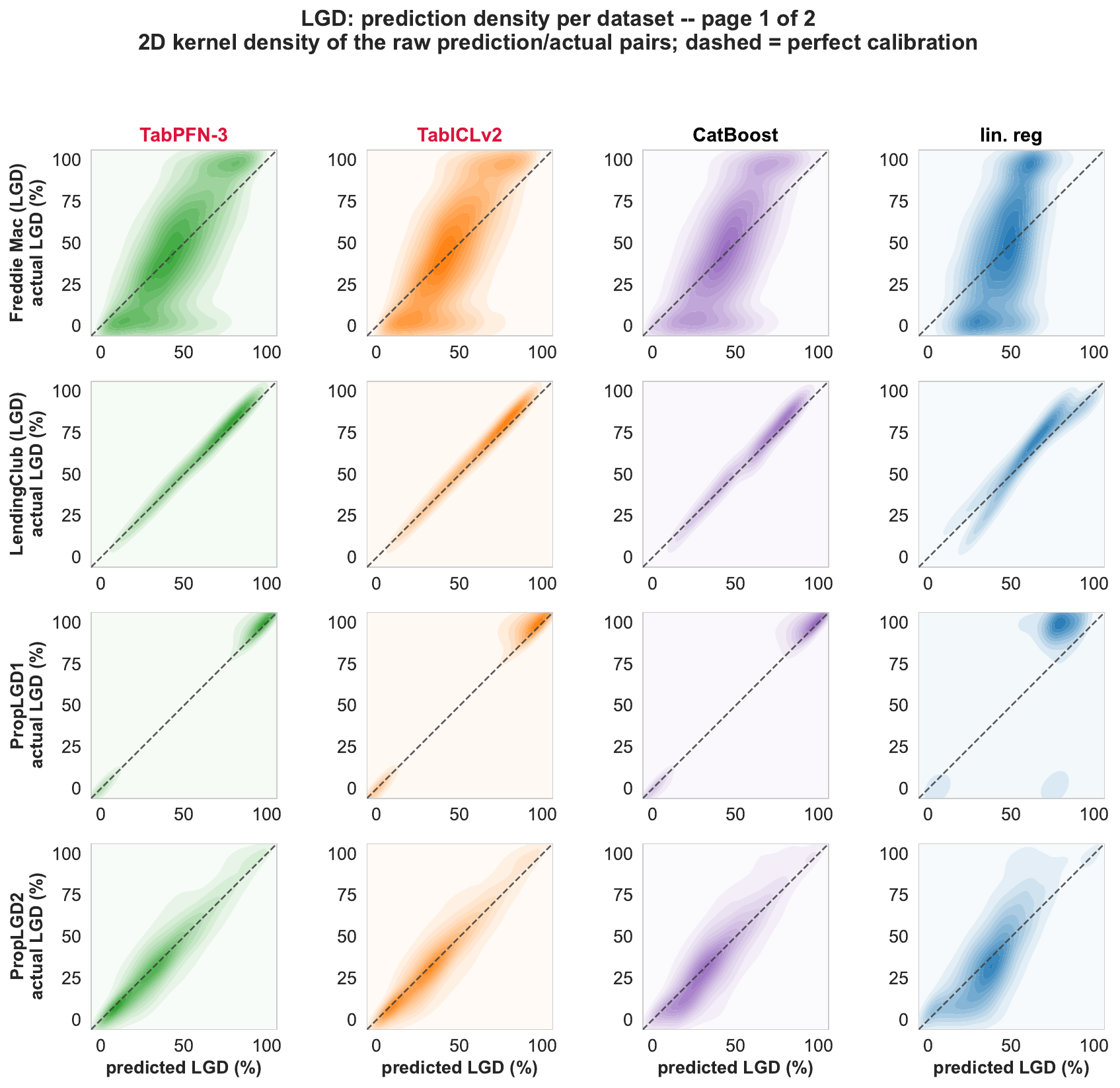}
    \caption{Prediction density per dataset (rows) and compared method (columns), page 1 of 2. Each panel is a \gls{twod} kernel density of raw out-of-fold predictions and realized \gls{lgd} values. Concentration near the dashed diagonal indicates smaller prediction errors, whereas a horizontal concentration indicates predictions that vary little with realized loss.}
    \label{fig:experiment1-lgd-lgd-per-dataset-prediction-density-p1}
\end{figure}

\begin{figure}[ht!]
    \centering
    \includegraphics[width=1.0\textwidth,height=0.90\textheight,keepaspectratio]{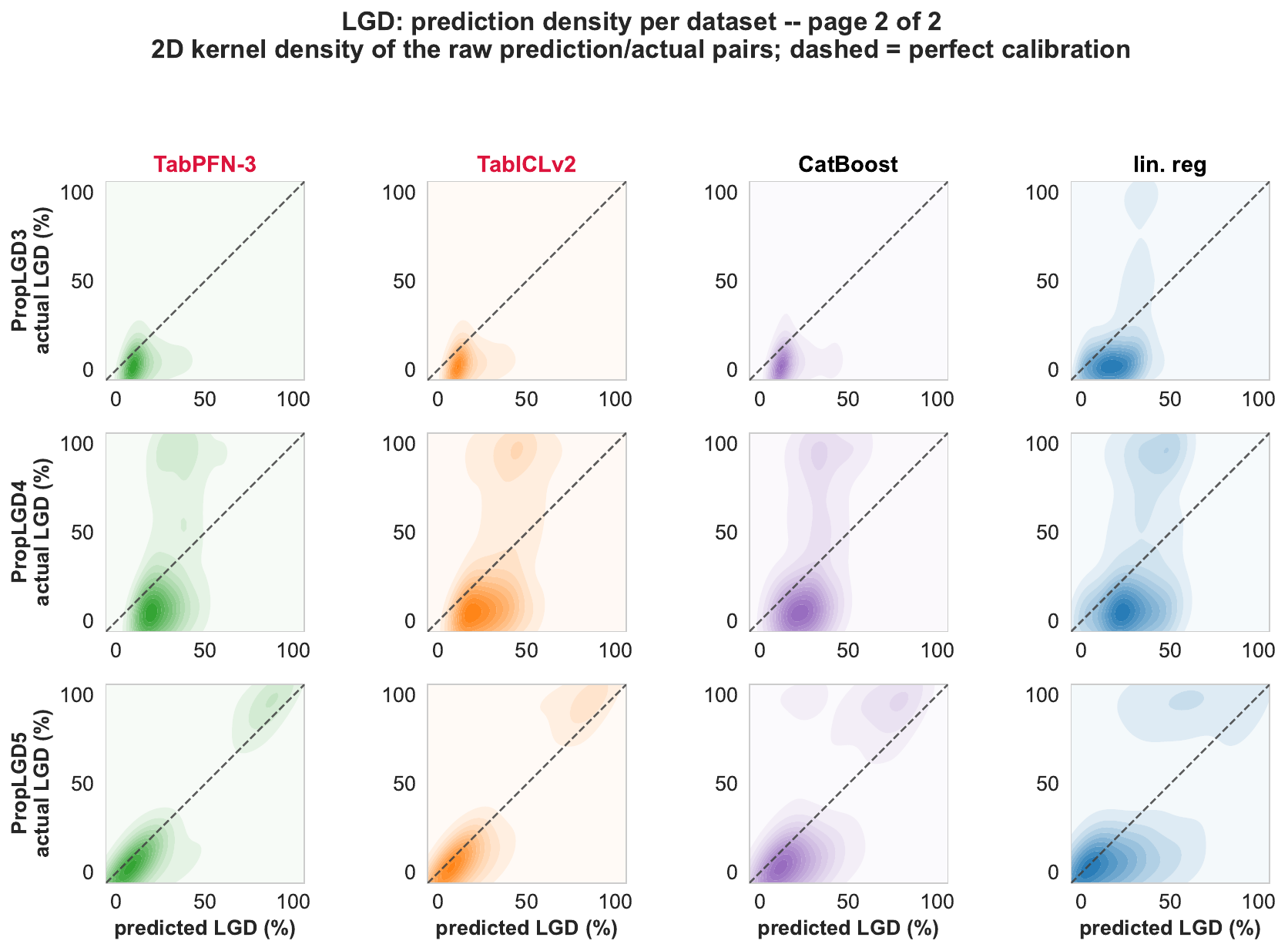}
    \caption{Prediction density per dataset (rows) and compared method (columns), page 2 of 2. Panels are constructed as in Figure~\ref{fig:experiment1-lgd-lgd-per-dataset-prediction-density-p1}.}
    \label{fig:experiment1-lgd-lgd-per-dataset-prediction-density-p2}
\end{figure}

\clearpage

\section{Rank-based results}
\label{supp:ranks}

The main text and Appendix~\ref{supp:metrics} compare methods on the level of the metric, which is informative but sensitive to datasets on which all methods score unusually high or low. A rank-based view removes that sensitivity by scoring each method only against its competitors within a dataset, and it is also the quantity on which the omnibus and post-hoc tests of Appendix~\ref{supp:stats} operate. We therefore report the average rank of every method, together with the per-dataset ranks from which it is computed.

\Needspace{8\baselineskip}
\subsection{PD classification}
\begin{figure}[ht!]
  \centering
  \includegraphics[width=1\textwidth,height=0.90\textheight,keepaspectratio]{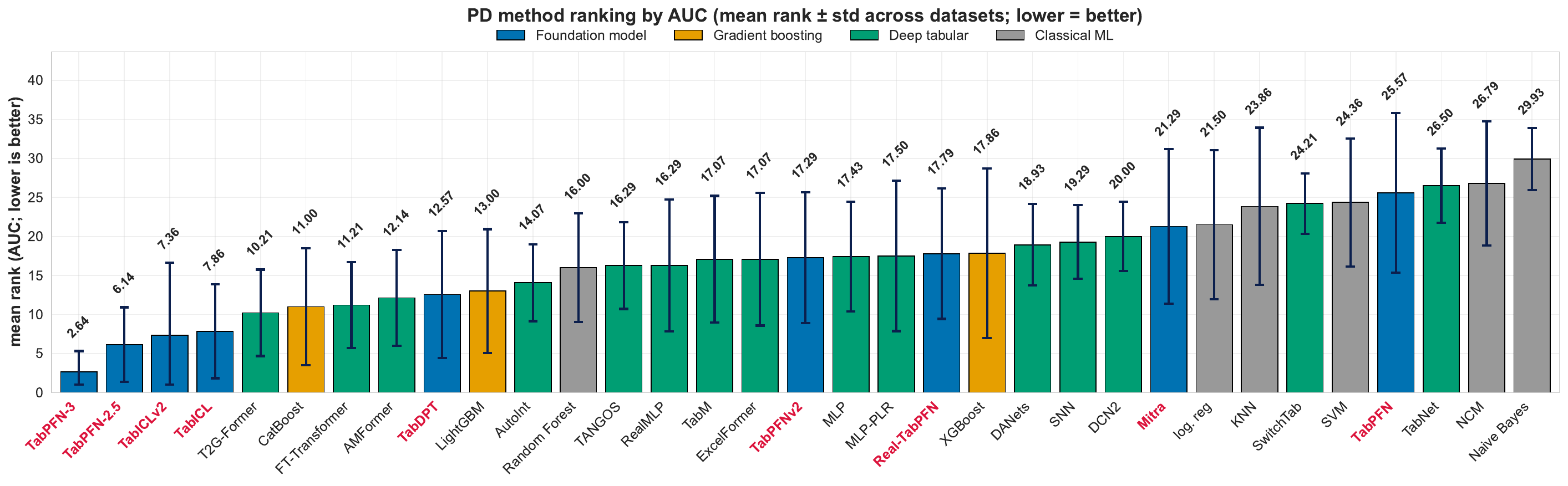}
  \caption{Average rank of each method on \gls{auc} across the 14 \gls{pd} datasets. Lower ranks are better; whiskers show the standard deviation of per-dataset ranks. Foundation models are highlighted in red.}
  \label{fig:pd_ranking}
\end{figure}

\begin{figure}[ht!]
  \centering
  \includegraphics[width=\textwidth,height=0.90\textheight,keepaspectratio]{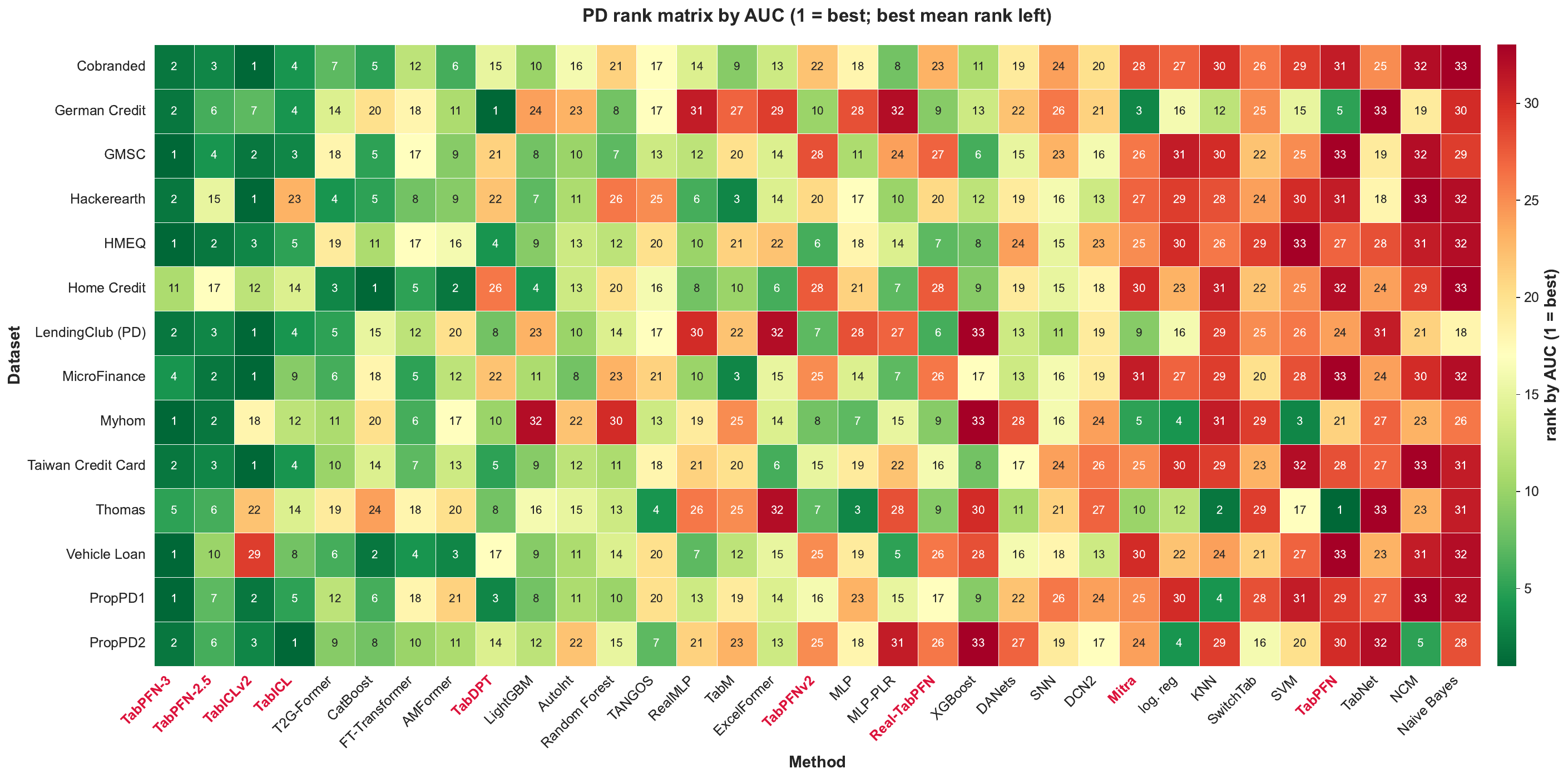}
  \caption{Rank of each method (columns) on every \gls{pd} dataset (rows), computed on the five-fold mean \gls{auc} within each dataset. Lower ranks are better; columns follow average rank across datasets. Foundation models are highlighted in red.}
  \label{fig:pd_rank_matrix}
\end{figure}

Figure~\ref{fig:pd_ranking} reports the average rank of each method on \gls{auc} across the 14 \gls{pd} datasets, and Figure~\ref{fig:pd_rank_matrix} shows the underlying per-dataset ranks. The rank view reproduces the ordering of the main text at the top and sharpens the separation. The four best average ranks all belong to foundation models, TabPFN-3 ($2.64$), TabPFN-2.5 ($6.14$), TabICLv2 ($7.36$), and TabICL ($7.86$), and the best conventional learner, CatBoost ($11.00$), is a clear distance behind, just after the best deep tabular network, T2G-Former ($10.21$). TabPFN-3 is the single best method on 5 of the 14 datasets and never falls below rank 11, which is the narrowest range of any method in the panel.

Two observations are visible in the rank view but not in the average \gls{auc}. First, consistency separates the two runners-up: TabPFN-2.5 overtakes TabICLv2 on average rank despite a lower average \gls{auc}, because TabICLv2 combines five outright wins with one dataset on which it falls to rank 29, giving it a rank standard deviation of $8.93$ against $4.61$ for TabPFN-2.5. Reporting the mean metric alone conceals this difference in cross-dataset rank consistency, which matters for a practitioner choosing a single model to deploy across a portfolio landscape. Second, one of the weakest methods on average, the original TabPFN ($25.57$), attains the top rank on one dataset, confirming that no method in the panel dominates uniformly and that dataset-level heterogeneity is substantial.

\clearpage

\Needspace{8\baselineskip}
\subsection{LGD regression}
\begin{figure}[ht!]
  \centering
  \includegraphics[width=1\textwidth,height=0.90\textheight,keepaspectratio]{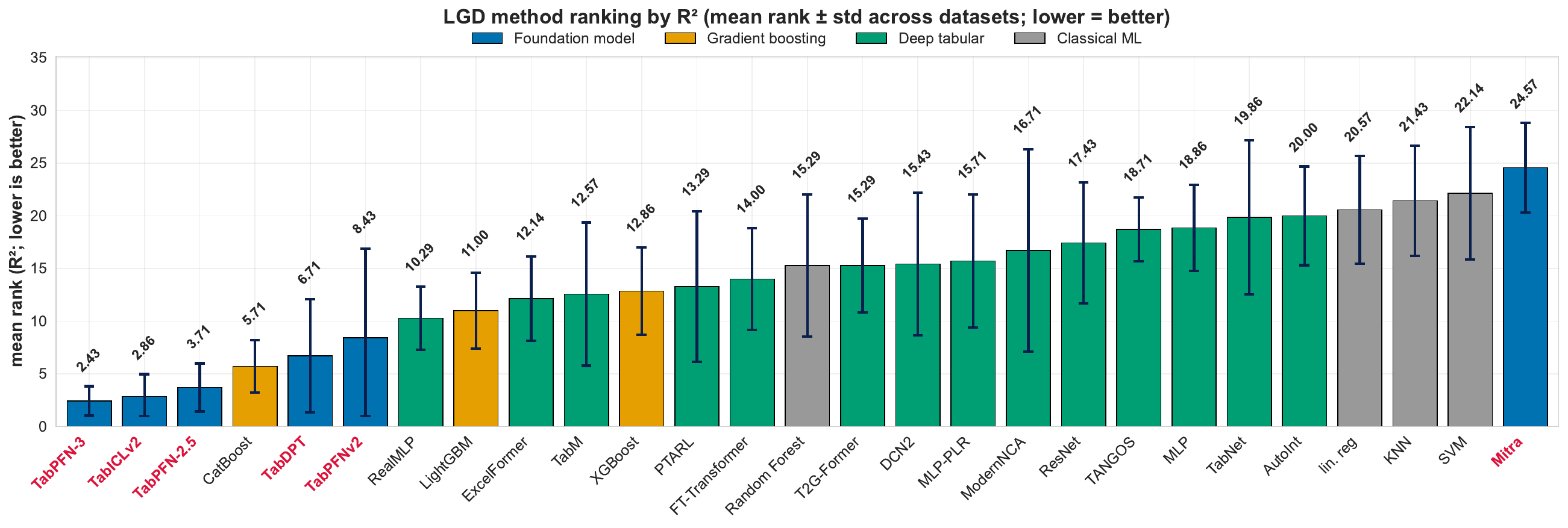}
  \caption{Average rank of each method on $R^2$ across the 7 \gls{lgd} datasets. Lower ranks are better; whiskers show the standard deviation of per-dataset ranks. Foundation models are highlighted in red.}
  \label{fig:lgd_ranking}
\end{figure}

\begin{figure}[ht!]
  \centering
  \includegraphics[width=\textwidth,height=0.90\textheight,keepaspectratio]{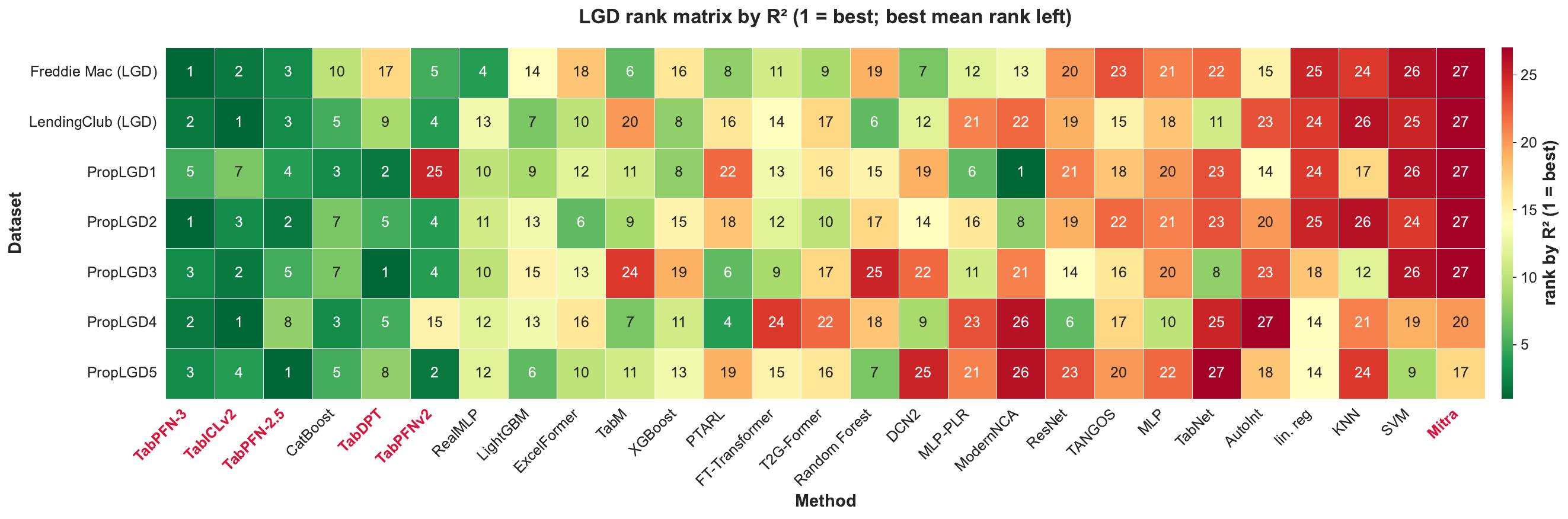}
  \caption{Rank of each method (columns) on every \gls{lgd} dataset (rows), computed on the five-fold mean $R^2$ within each dataset. Lower ranks are better; columns follow average rank across datasets. Foundation models are highlighted in red.}
  \label{fig:lgd_rank_matrix}
\end{figure}

Figure~\ref{fig:lgd_ranking} reports the average rank on $R^2$ across the 7 \gls{lgd} datasets, and Figure~\ref{fig:lgd_rank_matrix} shows the corresponding per-dataset ranks. The separation is sharper than for \gls{pd} classification. The three leading foundation models occupy the top three average ranks, TabPFN-3 ($2.43$), TabICLv2 ($2.86$), and TabPFN-2.5 ($3.71$), and none of them ever falls below rank 8. CatBoost, the best conventional learner, follows at $5.71$ and TabDPT at $6.71$. TabPFN-3 and TabICLv2 each take the top rank on two of the seven datasets, while TabPFN-2.5, TabDPT, and ModernNCA take it once each; TabDPT's rank standard deviation of $4.98$ and worst rank of 17 nonetheless make it far less consistent than the leading three.

The rank view also isolates the one clear failure among the foundation models. Mitra has the worst average rank of the entire 27-method panel ($24.57$), and the per-dataset matrix shows this is uniform rather than driven by one portfolio: its best rank on any \gls{lgd} dataset is 17. The foundation-model family is therefore not homogeneous, and the strong results of its leading members are carried by the recent TabPFN and TabICL generations rather than by the paradigm as such.

\clearpage

\section{Dataset details}
\label{supp:data}

This section documents the benchmark datasets in more detail than Table~\ref{tab:data-summary} allow. Appendix~\ref{supp:data_targets} reports the distribution of the \gls{lgd} target. Appendix~\ref{supp:data_eligibility} records which datasets enter the two auxiliary experiments and why the inclusion thresholds take the values they do. Appendix~\ref{supp:data_leakage} reports an overlap audit between our evaluation datasets and the pretraining corpora of the two benchmarked models that are pretrained on real data, Real-TabPFN and TabDPT.

\Needspace{8\baselineskip}
\subsection{Target distributions}
\label{supp:data_targets}

\begin{figure}[ht!]
  \centering
  \includegraphics[width=\textwidth,height=0.90\textheight,keepaspectratio]{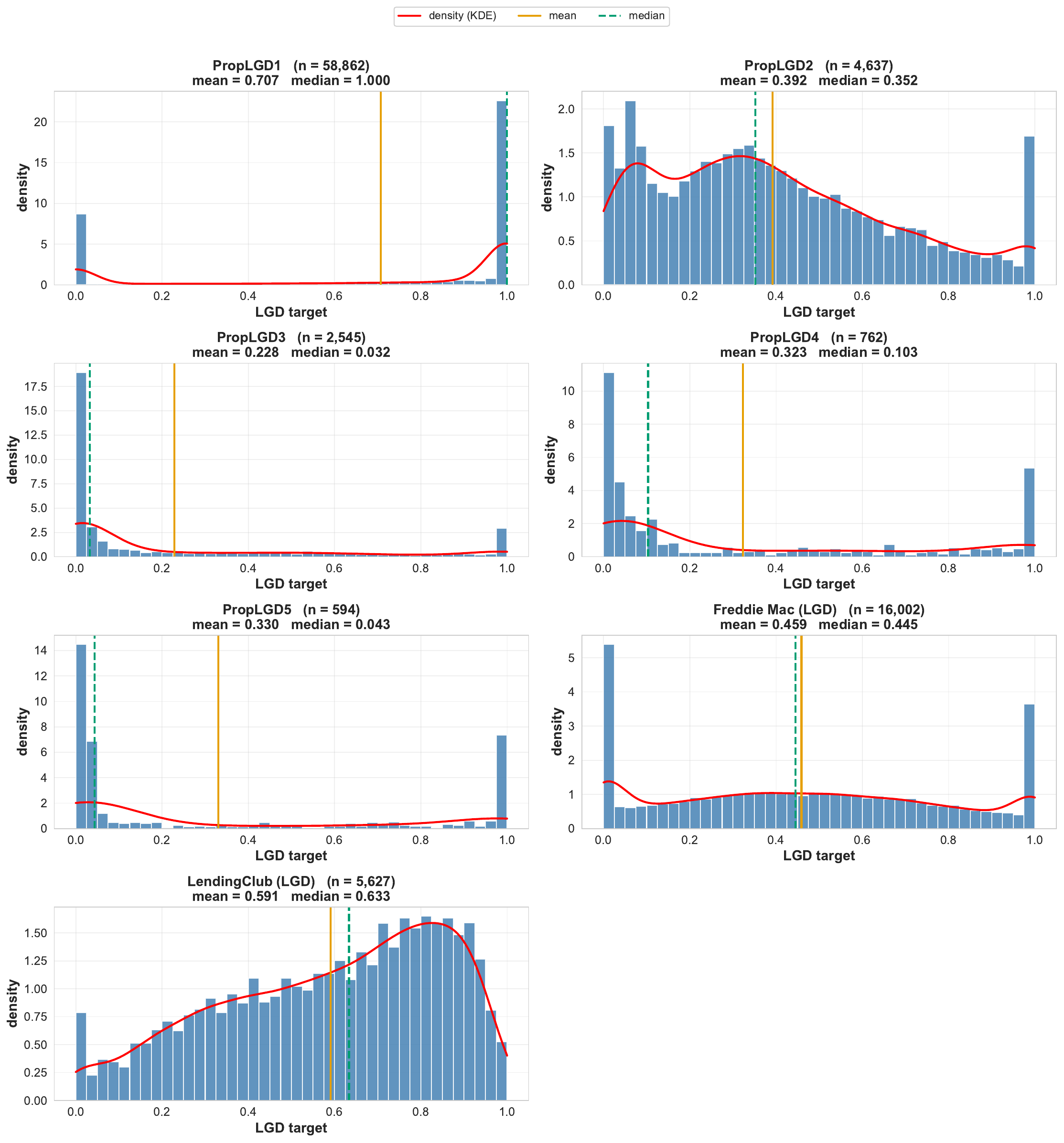}
  \caption{Distribution of the \gls{lgd} target in each of the seven \gls{lgd} datasets, after clipping to the unit interval. Red curves show kernel-density estimates; vertical lines mark the mean and median in each panel.}
  \label{fig:appendix-lgd-target-hists}
\end{figure}

Figure~\ref{fig:appendix-lgd-target-hists} shows the distribution of the target variable for each \gls{lgd} dataset. The bimodal, zero-one-inflated shape described in Section~\ref{subsec:data} is visible in most panels, with substantial probability mass at or near total recovery and at or near total loss, but its severity varies considerably: some portfolios are dominated by a single mode, whereas others place comparable mass at both boundaries.

\clearpage

\Needspace{8\baselineskip}
\subsection{Dataset eligibility for the auxiliary experiments}
\label{supp:data_eligibility}

Both auxiliary experiments described in Section~\ref{subsec:auxiliary_experiments} sweep a data characteristic downward from an upper bound, and that upper bound doubles as an inclusion threshold: a dataset can only be carried through the full sweep if it starts above the bound. The thresholds were therefore not chosen a priori but read off the data, by taking the largest bound that still admits a workable number of datasets. For \gls{pd}, a bound of $30{,}000$ observations is exactly the size of the smallest dataset it retains, Taiwan Credit Card, and admits nine of the fourteen \gls{pd} datasets; raising the bound would drop Taiwan Credit Card without adding anything, since the next smallest eligible dataset has $80{,}000$ observations, and lowering it would shorten every curve for the sake of datasets that are too small to contribute a meaningful sweep. For \gls{lgd}, a bound of $4{,}600$ observations is set just below the size of PropLGD2, the smallest dataset it retains with $4{,}637$ observations, and admits four of the seven \gls{lgd} datasets; the next candidate has only $2{,}545$ observations, so any bound that admitted a fifth dataset would nearly halve the length of the sweep for the other four. In both cases the bound maximizes the number of included datasets subject to keeping the sweep long enough to characterize the small-data regime, which is the object of interest.

\begin{table}[ht!]
\centering
\footnotesize
\setlength{\tabcolsep}{5pt}
\renewcommand{\arraystretch}{0.9}
\caption{Datasets included in the auxiliary experiments. Experiment~2 (sample efficiency) requires at least $30{,}000$ observations for \gls{pd} and $4{,}600$ for \gls{lgd}. Experiment~3 (class imbalance, \gls{pd} only) additionally requires a minority-class proportion above $0.15$.}
\label{tab:data_eligibility}
\begin{tabular}{l l c c}
\toprule
ID & Dataset name & Experiment 2 & Experiment 3 \\
\midrule
PD1  & Cobranded                  & \checkmark & \checkmark \\
PD2  & German Credit              &            &            \\
PD3  & \gls{gmsc} & \checkmark &            \\
PD4  & HackerEarth                & \checkmark & \checkmark \\
PD5  & \gls{hmeq}                       &            &            \\
PD6  & Home Credit                & \checkmark &            \\
PD7  & LendingClub (\gls{pd})           &            &            \\
PD8  & MicroFinance               & \checkmark & \checkmark \\
PD9  & Myhom                      &            &            \\
PD10 & Taiwan Credit Card         & \checkmark & \checkmark \\
PD11 & Thomas                     &            &            \\
PD12 & Vehicle Loan               & \checkmark & \checkmark \\
PD13 & PropPD1                    & \checkmark & \checkmark \\
PD14 & PropPD2                    & \checkmark &            \\
\midrule
LGD1 & Freddie Mac (\gls{lgd})          & \checkmark & n/a \\
LGD2 & LendingClub (\gls{lgd})          & \checkmark & n/a \\
LGD3 & PropLGD1                   & \checkmark & n/a \\
LGD4 & PropLGD2                   & \checkmark & n/a \\
LGD5 & PropLGD3                   &            & n/a \\
LGD6 & PropLGD4                   &            & n/a \\
LGD7 & PropLGD5                   &            & n/a \\
\bottomrule
\end{tabular}
\end{table}

The imbalance experiment applies a second criterion on top of the size threshold. Because the sweep starts at a minority-class proportion of $0.15$ and only removes minority observations, a dataset must already have a positive rate above $0.15$ to enter at the starting point without oversampling. Six of the nine size-eligible \gls{pd} datasets satisfy this; the three that do not are \gls{gmsc}, Home Credit, and PropPD2, whose default rates lie between $6.7\%$ and $9.3\%$ and are therefore already below the starting point of the sweep. Table~\ref{tab:data_eligibility} records the resulting inclusion pattern. One consequence is worth noting for interpretation: because both experiments select on size, they describe the behavior of the methods on subsampled versions of large portfolios rather than on the natively small datasets in the benchmark. Likewise, the imbalance experiment creates low default rates artificially by removing minority observations and therefore does not reproduce all characteristics of naturally low-default portfolios. Appendix~\ref{supp:viz_sizetrend} provides the complementary view at full dataset size.

\Needspace{8\baselineskip}
\subsection{Overlap with pretraining corpora}
\label{supp:data_leakage}

Because \glspl{tfm} are used off the shelf with their pretrained parameters fixed, direct overlap between an evaluation dataset and a model's pretraining corpus could lead to an unfair advantage. For models documented as using synthetic pretraining data only, direct real-dataset overlap is absent by construction. As summarized in Table~\ref{tab:tfm_architectures}, this applies to TabPFN, TabPFNv2, TabPFN-2.5, TabPFN-3, TabICL, TabICLv2, and Mitra. This does not rule out indirect benchmark exposure through model-development or selection decisions. Two of the benchmarked models are pretrained wholly or partly on real data and therefore require an explicit check. Real-TabPFN continues the pretraining of TabPFNv2 on $71$ curated tables drawn from OpenML and Kaggle \citep{2025.GARG.realtabpfnimprovingtabularfoundationmodelscontinuedpretrainingrealworlddata}, and TabDPT is pretrained on $123$ OpenML tables using self-supervised column masking \citep{2025.MA.tabdptscalingtabularfoundationmodelsrealdata}. Both papers document a decontamination procedure, but in each case that procedure was executed against the authors' own evaluation suites, the OpenML AutoML Benchmark for Real-TabPFN and CC18 together with CTR23 for TabDPT, so neither provides a guarantee for a downstream benchmark such as ours.

We therefore compared every dataset in our benchmark against the complete corpus listings published by both models, matching on dataset name, source repository, target variable, and dimensions, and treating a corpus entry as an overlap when it is the same dataset or a subsampled or otherwise reduced version of it. Table~\ref{tab:leakage} reports the outcome. Three overlaps are identified, all of them on the \gls{pd} side. Real-TabPFN's corpus contains the vehicle financing dataset that we use as Vehicle Loan. TabDPT's corpus contains a reduced version of \gls{gmsc}, listed under a generic name but carrying the same target variable, and a subsampled version of Taiwan Credit Card. No other dataset in the benchmark, public or proprietary, matches an entry in either corpus. German Credit, in particular, is absent from the TabDPT corpus because it belongs to CC18, the suite against which TabDPT deduplicated. No \gls{lgd} dataset overlaps with either corpus: Real-TabPFN is a classification-only model whose corpus contains no regression tasks, and none of our \gls{lgd} datasets appears in the TabDPT listing.

\begin{table}[ht!]
\centering
\footnotesize
\setlength{\tabcolsep}{5pt}
\renewcommand{\arraystretch}{0.9}
\caption{Overlap between the benchmark datasets and the pretraining corpora of the two benchmarked models that are pretrained on real data. All other benchmarked foundation models are documented as using synthetic pretraining data only, so direct real-dataset overlap cannot occur under their reported pretraining designs. Only datasets with an identified overlap are listed; no overlap was found for any other dataset in the benchmark.}
\label{tab:leakage}
\begin{tabular}{l l l l}
\toprule
ID & Dataset name & Corpus & Corpus entry \\
\midrule
PD3  & \gls{gmsc} & TabDPT      & OpenML 44089, reduced to 16{,}714 rows \\
PD10 & Taiwan Credit Card         & TabDPT      & OpenML 45020, subsampled to 13{,}272 rows \\
PD12 & Vehicle Loan               & Real-TabPFN & Kaggle, full dataset \\
\bottomrule
\end{tabular}
\end{table}
\clearpage

The available sensitivity checks suggest that these identified overlaps do not drive the aggregate conclusions. First, neither affected model ranks among the top three on either task, while TabPFN-3, TabICLv2, and TabPFN-2.5, which occupy the top three positions for both \gls{pd} and \gls{lgd}, are documented as using synthetic pretraining data only; TabDPT places sixth of the $33$ \gls{pd} methods on mean \gls{auc} and fifth among the nine foundation models. Second, for each affected model, performance on the overlapping datasets can be compared descriptively with performance elsewhere. For TabDPT, the mean rank is $13.0$ over the two overlapping datasets and $12.5$ over the twelve remaining ones, and its \gls{auc} advantage over the panel mean is smaller on the overlapping datasets ($+0.0065$) than on the remaining ones ($+0.0157$). It places fifth among the nine foundation models on both overlapping datasets, matching its median position across the full benchmark. The two overlapping datasets are themselves dissimilar: on \gls{gmsc}, the larger of the two, TabDPT ranks $21$st of $33$ methods and falls below CatBoost, whereas on Taiwan Credit Card it ranks fifth. Removing both datasets from the average moves TabDPT from sixth to seventh place of $33$, a shift that reflects its near-tie with the adjacent methods rather than a loss of advantage. For Real-TabPFN, the comparison with its TabPFNv2 backbone is also informative because continued pretraining includes Vehicle Loan. On Vehicle Loan, Real-TabPFN attains $0.6277$ against the backbone's $0.6279$, a difference of $-0.0003$ that lies near the center of the differences observed on the thirteen remaining datasets (mean $-0.0002$, range $-0.0017$ to $+0.0013$). Its rank on Vehicle Loan, $26$th of $33$, is worse than its median rank of $17$th on the remaining datasets. Third, no \gls{lgd} dataset is identified in either documented real-data pretraining corpus, so the entire \gls{lgd} comparison is free of documented overlap. With two overlapping datasets for one model and one for the other, these comparisons are descriptive rather than inferential: we do not observe an advantage associated with the identified overlaps, but cannot rule out an effect. We report them because reuse of public credit datasets in foundation-model pretraining corpora is likely to become more common as those corpora grow, and because the same consideration applies whenever a pretrained model is evaluated on data that may have entered a public repository.

\end{document}